\documentclass[preprint,12pt]{elsarticle}

\usepackage[dvipsnames]{xcolor} 
\usepackage{array}
\usepackage{algorithmic}
\usepackage{booktabs}
\usepackage{tabularx}
\usepackage{hyperref}
\usepackage{amsfonts} 
\usepackage{breqn}
\usepackage{bm}
\usepackage{multirow}
\usepackage{pdflscape}
\usepackage{float}
\usepackage{caption}
\usepackage{xcolor}
\usepackage{subfigure}
\definecolor{darkred}{rgb}{0.9,0.1,0.1}
\usepackage{longtable}
\usepackage[ruled,vlined]{algorithm2e}
\hypersetup{
 colorlinks,
 citecolor=black,
 filecolor=black,
 linkcolor=black,
 urlcolor=black
}
\usepackage{amssymb}
\usepackage{amsthm}

\usepackage[top=1.2in, bottom=1.4in, left=1.2in, right=1.2in]{geometry}

\usepackage{arydshln}
\journal{arXiv}

\biboptions{sort&compress}

\newtheorem{theorem}{Theorem}

\theoremstyle{definition}
\newtheorem{definition}{Definition}

\newcommand{\bs}[1]{\boldsymbol{#1}}

\begin{document}

\begin{frontmatter}

\title{KKANs: K{$\dot {\rm u}$}rkov{\'a}-Kolmogorov-Arnold Networks and Their Learning Dynamics}

\author[inst1]{Juan Diego Toscano}
\author[inst2]{Li-Lian Wang}
\author[inst1,label2]{George Em Karniadakis}
\affiliation[inst1]{organization={Division of Applied Mathematics, Brown University},
  city={Providence},
  postcode={02912}, 
  state={RI},
  country={USA}}
\affiliation[inst2]{organization={Division of Mathematical Sciences, Nanyang Technological University},
  postcode={637371}, 
  country={Singapore}}

\fntext[label2]{Corresponding author: george\_karniadakis@brown.edu}

\begin{abstract}

Inspired by 
the Kolmogorov-Arnold representation theorem and K{$\dot {\rm u}$}rkov{\'a}'s principle of using  approximate representations, we propose the 
K{$\dot {\rm u}$}rkov{\'a}-Kolmogorov-Arnold Network (KKAN), a new two-block architecture that combines robust multi-layer perceptron (MLP) based inner functions with flexible linear combinations of basis functions as outer functions. We first prove that  KKAN is a universal approximator and then we demonstrate its versatility across scientific machine-learning applications, including function regression, physics-informed machine learning (PIML), and operator learning frameworks. The benchmark results show that KKANs outperform MLPs and the original Kolmogorov-Arnold Networks (KANs) in function approximation and operator learning tasks and achieve performance comparable to fully optimized MLPs for PIML. To better understand the behavior of the new representation models, we analyze their geometric complexity and learning dynamics using information bottleneck theory, identifying three universal learning stages, fitting, transition, and diffusion, across all types of architectures. We find a strong correlation between geometric complexity and signal-to-noise ratio (SNR), with optimal generalization achieved during the diffusion stage. Additionally, we propose self-scaled residual-based attention weights to maintain high SNR dynamically, ensuring uniform convergence and prolonged learning. 

\end{abstract}



\end{frontmatter}

\section{Introduction}

Scientific machine learning (SciML) has emerged as a transformative approach for solving complex scientific problems by integrating machine learning with domain-specific knowledge from physics, biology,  engineering, finance, and beyond \cite{toscano2024pinns}. Within this framework, methods such as Physics-Informed Machine Learning (PIML)~\cite{raissi2019deep} and Operator Learning~\cite{lu2019deeponet} have gained significant attraction due to their ability to merge the predictive power of machine learning with the foundational principles of physics~\cite{toscano2024pinns}. These approaches rely on data-efficient and physics-guided learning to model systems that are otherwise difficult to solve using traditional methods~\cite{raissi2020hidden,cai2021artificial,boster2023artificial,toscano2024inferring,toscano2024inferring_AIV}.

A cornerstone of SciML is the Multilayer Perceptron (MLP) used as the primary model architecture~\cite{karniadakis2021physics}. MLPs are foundational in modern deep learning~\cite{goodfellow2020generative,vaswani2017attention,he2016deep,li2018visualizing,toscano2023teeth,kaelbling1996reinforcement} due to their proven ability to approximate complex functions, as guaranteed by the universal approximation theorem~\cite{Park1991universal}. Despite their widespread success, MLPs face notable challenges, including limited interpretability~\cite{cranmer2023interpretable}, susceptibility to overfitting, and scalability issues~\cite{jagtap2020extended,jagtap2020conservative}. Addressing these limitations has become a central focus of recent advancements in SciML  frameworks~\cite{jagtap2020adaptive,wang2021understanding,mcclenny2023self}.

Kolmogorov-Arnold Networks (KANs)~\cite{liu2024kan,liu2024kan2,wang2024expressiveness} have been proposed as an alternative to MLPs, offering advantages such as enhanced interpretability, high accuracy in function regression, and resilience to catastrophic forgetting and spectral bias~\cite{wang2024expressiveness,vaca2024kolmogorov,samadi2024smooth}. KANs are inspired by the Kolmogorov-Arnold Representation Theorem (KART), which provides a framework for decomposing multivariate functions into sums of univariate inner and outer functions~\cite{sprecher2002space,koppen2002training,schmidhuber1997discovering,lai2021kolmogorov,leni2013kolmogorov,he2023optimal,somvanshi2024survey}. Despite their strengths, the original KANs~\cite{liu2024kan} diverge from the original KART and employ a stacked representation that relies on computationally expensive learnable B-splines as basis functions. Furthermore, their performance rapidly degrades with high parameter counts, limiting their application for SciML and raising questions about their suitability for real-world tasks~\cite{pal2024understanding}.

To address these limitations, several variations of KANs have been introduced~\cite{li2024kolmogorov,bozorgasl2024wavkan,jacobiKANs,ss2024chebyshev,torchkan}. Notable examples include FastKANs~\cite{li2024kolmogorov} and cKANs~\cite{ss2024chebyshev}, which use radial basis functions (RBFs) and Chebyshev polynomials as basis functions, respectively. These methods improve computational efficiency but still exhibit drawbacks. For instance, cKANs, while faster than KANs, are slower than MLPs and exhibit instabilities in single-precision arithmetic. Recursive formulations~\cite{shukla2024comprehensive} can restore cKAN stability, but tuning hyperparameters remains challenging~\cite{toscano2024inferring}, while performance degrades with an increasing number of parameters or higher polynomial orders.

KANs have been successfully extended to PIML~\cite{liu2024kan,shukla2024comprehensive,wang2025physics} and operator learning~\cite{abueidda2024deepokan,shukla2024comprehensive}, with several specialized architectures and applications introduced, such as GRU-KAN~\cite{zhang4957859pkan}, separable physics-informed KANs~\cite{jacob2024spikans}, multifidelity KANs~\cite{howard2024multifidelitykolmogorovarnoldnetworks}, finite basis KANs~\cite{howard2024finite}, and others~\cite{ma2024integrating,fareaa2024learnable,mostajeran2024epi,rigas2024adaptive}. Other frameworks inspired by different versions of the Kolmogorov superposition theorem have also been proposed, such as AcNet~\cite{guilhoto2024deeplearningalternativeskolmogorov}. Some of these approaches have demonstrated advantages over MLPs in PIML, such as robustness to noise~\cite{shukla2024comprehensive} and reduced network size requirements~\cite{toscano2024inferring}. However, due to the maturity of MLPs, achieving comparable performance to a fully optimized MLP often requires extensive hyperparameter tuning and specialized training strategies~\cite{guilhoto2024deeplearningalternativeskolmogorov}, such as learning rate warmup, gradient clipping, and causality-enforcing methods. These challenges highlight the limitations of existing KAN-based models, which often deviate from the original theorem by employing deeply nested architectures.

From the theoretical perspective, KANs' nested formulations~\cite{liu2024kan} and some of their variations~\cite{guilhoto2024deeplearningalternativeskolmogorov} have been shown to be universal approximators. Notably, ~\cite{wang2024expressiveness} compared the approximation capabilities of KANs and MLPs, demonstrating that KANs' representation capabilities are at least as good as those of MLPs, and analyzed KANs' resistance to spectral bias. Additionally, ~\cite{zhang2024generalization} presented a theoretical analysis of KANs by establishing generalization bounds, and  \cite{guilhoto2024deeplearningalternativeskolmogorov} proved that their initialization scheme scales with the size of the network and does not suffer from vanishing derivatives. Furthermore, convergence estimates for one of KAN's predecessors are provided in ~\cite{Igelnik2003}. However, some of these studies relied on assumptions restricted to specific types of basis functions and imposed regularity conditions on the outer mappings, which are not valid for a broader class of functions.

To address these theoretical and computational challenges, we propose the K{$\dot {\rm u}$}rkov{\'a}-Kolmogorov-Arnold Networks (KKANs) inspired by a special variant of the Kolmogorov-Arnold representation theorem by K{$\dot {\rm u}$}rkov{\'a}~\cite{Kourkova1991,Kuurkova1992kolmogorov}. KKANs consist of a new two-block architecture that {\em adheres closely} to the original theorem. KKANs use robust MLPs as inner functions and linear combinations of basis functions as outer functions. This design combines the robustness and adaptability of MLPs with the interpretability and flexibility of basis function representations~\cite{liu2024kan}.
We prove that KKANs are universal approximators regardless of the choice of basis function and for a general class of functions. Then, we extend their applicability to PIML and operator learning frameworks. Additionally, we demonstrate that KKANs can integrate modern enhancements originally developed for MLPs, such as weight normalization~\cite{salimans2016weight}, Fourier feature expansions~\cite{wang2021eigenvector}, and residual connections~\cite{wang2024piratenets}. These adaptations, previously challenging for KANs, unlock KKANs' potential for a wide range of applications. Experimentally, we demonstrate that KKAN performance remains robust with an increasing number of parameters and is stable for a high number of basis functions. Through extensive benchmarking, we observe that KKANs outperform MLPs in function approximation and operator learning tasks, achieve comparable performance, and can outperform several state-of-the-art MLPs in PIML.

In addition to its structural advantages, we analyze the learning dynamics of KKANs, cKANs, and MLPs via the Information Bottleneck (IB) method~\cite{tishby2000information} and identify a strong correlation with the geometric complexity evolution~\cite{dherin2022neural}. We observe that KKANs and cKANs exhibit lower complexity than MLPs at initialization, which could be correlated with enhanced generalization capabilities. Using the IB theory, we identify three distinct learning stages: fitting, transition, and diffusion~\cite{anagnostopoulos2024learning} across all architectures and for all types of problems (i.e., function approximation, PIML and neural operators), with optimal generalization achieved during the diffusion stage when the signal-no-noise ratio (SNR) of the backpropagated gradients is high.

Finally, we propose new optimization techniques for PIML. Specifically, we extend the residual-based attention weights (RBA) introduced in~\cite{anagnostopoulos2024residual,toscano2024inferring,toscano2024inferring_AIV} to self-scaled RBA (ssRBA), a method that dynamically adapts during learning to maintain a high SNR. This ensures uniform convergence across the domain and enables prolonged learning. Additionally, we introduce a scaling weighting strategy that combines the ideas presented in~\cite{wang2021understanding} and~\cite{liu2024config} to automatically balance the loss-specific gradients, which enables achieving an optimal update direction, improving both stability and training efficiency.

In summary, the key contributions of this work are as follows:
\begin{itemize}
    \item We propose KKAN, a novel two-block architecture that combines the robustness of MLPs with the interpretability and flexibility of basis function representations, adhering closely to the Kolmogorov-Arnold theorem.
    \item We prove that KKAN is a universal approximator regardless of the basis function selection and for a general class of functions.
    \item We extend KKAN's applicability to PIML and operator learning.
    \item We analyze KKAN’s learning dynamics using information bottleneck theory, identifying a strong correlation between geometric complexity and SNR and corroborating the diffusion stage as the point of optimal generalization.
    \item We develop self-scaled residual-based attention weights (ssRBA) that dynamically maintain a high SNR, enabling uniform convergence and prolonged training.
    \item Through extensive benchmarking, we demonstrate that KKANs outperform MLPs and cKANs in function approximation and operator learning tasks and achieve comparable performance with fully optimized MLPs for PIML applications.
\end{itemize}

The remainder of this paper is organized as follows. In Section~\ref{Sec2}, we review the Kolmogorov-Arnold representation theorem and its key variants, concluding with a universal approximation theorem for two-block structures, which we use to introduce the KKAN architecture. Section~\ref{sec_methods} describes the methodology, including function approximation, PIML, operator learning frameworks, and evaluation criteria such as relative errors, geometric complexity, and learning dynamics. Additionally, we detail the proposed self-scaled residual-based attention (ssRBA) method. In Section~\ref{Results}, we present computational experiments, including benchmarks for function approximation, PIML, and operator learning. Section~\ref{Sec_learning} analyzes learning dynamics across the proposed models and their connection to the geometric complexity. Finally, in Section~\ref{Sec_summary}, we summarize our findings and discuss potential future research directions.

\section{Kolmogorov-Arnold Representation Theorem (KART)-inspired Architecture}
\label{Sec2}
\subsection{History and representative formulations}

\smallskip 

 In a series of papers~\cite{Kolmogorov1956,Arnold1957a,Arnold1957b,Kolmogorov1957},  Andrey Kolmogorov and Vladimir Arnold studied 
the representation of continuous functions of several variables on a bounded domain by superpositions of continuous functions of a smaller number of variables. Their astonishing discovery was summarized as the Kolmogorov-Arnold Representation Theorem  (KART)  in Kolmogorov~\cite{Kolmogorov1957} (1957): {\em For any integer $d \geq 2,$ there are continuous real functions $\psi_{p, q}(x)$ on the closed unit interval $E^1=[0, 1]$ such that each continuous real function $f(x_1, \ldots, x_d)$ on the $d$-dimensional unit cube $E^d=[0,1]^d$ is representable as}
\begin{equation}\label{KART-0}
f(x_1, \ldots, x_d)=\sum_{q=0}^{2 d} g_q\Big(\sum_{p=1}^d \psi_{p, q}(x_p)\Big),
\end{equation}
{\em where $g_q(y)$ are continuous real functions on $\mathbb R=(-\infty,\infty)$.}
Here, $\psi_{p,q}$ are known as the {\em inner functions} (which are universal and independent of $f$), while $g_q$ are referred to as the {\em outer functions} (which depend on $f$).  
 This theorem and a previous result of Arnold~\cite{Arnold1957b}  can be considered as a refutation of Hilbert's Problem $13$: {\em``There are continuous functions of three variables, not representable as superpositions of continuous functions of two variables''} (see~\cite{Girosi1989}).
 
 Over the past several decades, research following this mathematically elegant representation has advanced mainly along two lines: (i) construction of refined versions to strengthen the connection to neural networks (with a two-fold focus on the reduction of the number of involved one-dimensional functions and on what smoothness conditions could be imposed on them); and (ii)  theory to practical applications in deep learning and general machine learning. 
 
 \begin{center}
 \begin{table}[!h]
\setlength{\tabcolsep}{1pt} 
\renewcommand{\arraystretch}{1.5} 
\small
\caption{KART and some selected variants}\label{Tab:KART} 
\begin{tabular}[!h]{|c|c|c|c|} 
\hline 
Version  & Representation & 
Inner Function&
Outer  Function \\
\hline 
\begin{tabular}{c} 
KART \\
$(1957)$
\end{tabular} & $\displaystyle\sum_{q=0}^{2 d} {g_q}\bigg(\sum_{p=1}^d {\psi_{p, q}}(x_p)\!\bigg)$ & 
\begin{tabular}{c} 
 $ d(2d+1)$ \\[1pt]
\hdashline
${\psi_{p,q}}\in C([0,1])$
\end{tabular}
 &
 \begin{tabular}{c} 
 $2d+1$ \\[1pt]
\hdashline
${g_q}\in  C(\mathbb R)$
\end{tabular} \\
\hline 
\begin{tabular}{c} 
Lorentz \\
$(1962)$
\end{tabular} & 
{$\displaystyle \sum_{q=0}^{2 d} { g}\bigg(\sum_{p=1}^d \lambda_p {\psi_q}(x_p)\!\bigg)$} & 
{\begin{tabular}{c} 
 $2d+1$ \\[1pt]
\hdashline 
${\psi_{q}}\in {\rm Lip}^\alpha; \in [0,1]; \nearrow$
\end{tabular}} & 
{\begin{tabular}{c} 
 $1$ \\[1pt]
\hdashline
${g}\in C([0,d])$
\end{tabular}}  \\
\hline 
\begin{tabular}{c} 
Sprecher \\
$(1965)$
\end{tabular} & $\displaystyle\sum_{q=0}^{m} {g_q}\bigg(\sum_{p=1}^d \lambda_p {\psi}(x_p+q a)\!\bigg)$ & 
\begin{tabular}{c} 
 $1$ \\[1pt]
\hdashline
${{\psi}\in {\rm Lip}^{\log_\gamma \! 2}};\; [0,2]\to [0,2]; \nearrow$
\end{tabular}
& \begin{tabular}{c} 
 $m+1$ \\[1pt]
\hdashline
{${g_q}\in C([0,2\frac{\gamma-1}{\gamma-2}])$}
\end{tabular}\\
\hline 
\begin{tabular}{c} 
Braun  \\
$(2009)$
\end{tabular} & 
$\displaystyle\sum_{q=0}^{2 d} g\Big(\sum_{p=1}^d \lambda_p {\psi}(x_p+q a)+c_q\Big)$ & 
\begin{tabular}{c} 
 $1$ \\[1pt]
\hdashline
${\psi}\in C(\mathbb R); \nearrow$
\end{tabular} & \begin{tabular}{c} 
 $1$ \\[1pt]
\hdashline
${g}\in C(\mathbb R)$
\end{tabular} 
\\
\hline 
 \begin{tabular}{c} 
Schmidt-\\
Hieber  \\
$(2021)$
\end{tabular} & 
\begin{tabular}{c} 
$g\Big( \displaystyle\sum_{p=1}^d 3^{1-p} \psi(x_p)\Big) $\\
\hdashline
$f: \beta\text{-smooth}, \beta\in (0,1]$ 
\end{tabular}
& 
\begin{tabular}{c} 
 $1$ \\[1pt]
\hdashline
$ \psi\in C: {[0,1]\to {\mathcal C}}; \nearrow$
\end{tabular} & \begin{tabular}{c} 
 $1$ \\[1pt]
\hdashline
${g:{\mathcal C}\to \mathbb R}; \alpha\text{-smooth}$
\end{tabular} 
\\
\hline
\begin{tabular}{c} 
 Ismayilova     \\
 \& Ismailov\\
$(2024)$
\end{tabular} & 
\begin{tabular}{c} 
\text{Formula as in Braun} 
$(2009)$\\[1pt]
\hdashline
\text{with given} $a(\gamma),\lambda_p(\gamma)$\\[1pt]
\hdashline
$c_q=(2d+1)q$
\end{tabular}& 
\begin{tabular}{c} 
 $1$ \\[1pt]
\hdashline
$\psi\in {\rm Lip}^{\log_\gamma 2}; [0,2]\to [0,2]; \nearrow$
\end{tabular} & 
\begin{tabular}{c} 
 $f\in C\Rightarrow g\in C $ \\[1pt]
\hdashline
 $f$ \text{dis}-$C$ $\Rightarrow g$ \text{dis}-$C$\\[1pt]
\hdashline
$\|f\|_\infty,\|g\|_\infty=\infty$ \\[1pt]
\end{tabular}
\\
\hline 
\end{tabular}
\end{table}
\end{center}

\vspace*{-22pt} 

 In Table \ref{Tab:KART}, we summarize some representative variants of the original KART.    Lorentz~\cite{Lorentz1962metric}  
 first noticed that the outer functions $g_q$ can be chosen all the same, and Sprecher~\cite{Sprecher1963dissertation} showed that one can take the inner functions  $\psi_{p,q}(x)$ to be 
 $\lambda_p\psi_q(x).$ These interpretations inspired  Lorentz's version  in~\cite[Chapter 11]{Lorentz1966approximation}, where the constants $\lambda_p\in (0,1],$ and the inner functions $\psi_q$ are strictly increasing and are of the $\alpha$-Lipschitz class with $\alpha\in (0,1).$   Kahane~\cite{Kahane1975} (also see~\cite[Chapter 15]{Lorentz1996}) provided an elegant proof of Lorentz's version with the refinement: 
 $\lambda_p>0,$ 
 $\sum_{p=1}^d\lambda_p\le 1,  $ and $\psi_q\in {\rm Lip}_1 1$ (i.e., both the Lipschitz index and constant are $1$).
 Sprecher's version (refined from his earlier versions in~\cite{Sprecher1965structure,Sprecher1996numerical}) was  stated in~\cite{Sprecher1997numerical}, where the parameters $m\ge 2d, \gamma\ge m+2,$ and 
\begin{equation}\label{alambda} 
a=\frac{1}{\gamma(\gamma-1)};\quad   \lambda_1=1, \quad  \lambda_p=\sum_{r=1}^{\infty} \gamma^{-(p-1)(d^r-1) /(d-1)},\;\;\; 2\le p\le d.
\end{equation}
Given these parameters,  the finite domain and range of the outer and inner functions are $[0,2]$ and $[0,\frac{2(\gamma-1)}{\gamma-2}],$ respectively (see~\cite{Montanelli2020error}). 
Like the Lorentz's version,  
the inner function $\psi$ is of 
$\frac{\ln 2} {\ln \gamma}$-Lipschitz class and monotonically increasing. The modified version in Braun~\cite{braun2009application} resulted from the constructive proof of Sprecher, where the parameters could be constructed explicitly as well.  
It is noteworthy that (a) the introduction of an additional dilation $c_q$ was first realized in~\cite{Sprecher1965structure}; and (b) the numerical construction in Sprecher~\cite{Sprecher1996numerical,Sprecher1997numerical} could not guarantee the continuity and monotonicity of the inner function $\psi.$  To remedy the latter, K\"oppen~\cite{Koppen2002} modified  Sprecher's numerical implementation, and 
Braun and Griebel~\cite{Braun2009constructive} provided a theoretical justification. Interested readers are referred to  Sprecher's book~\cite{Sprecher2017algebra} and the references therein for many more developments and insights of KART up to 2017. 
Recently, Schmidt-Hieber~\cite{schmidt2021kolmogorov} constructed a much-simplified version using the notion of space-filling curves (particularly,  the Lebesgue curve on a Cantor set ${\mathcal C},$ see ~\cite[Chapter 7]{Bader2013}). Remarkably, it transfers the smoothness of $f$ to the outer function $g$ in the sense that if $f$ is $\beta$-smooth (i.e., $\exists \beta\in (0,1],$ such that $|f(x)-f(y)|\leq Q\,|x-y|_{\infty}^\beta$ for any $x,y\in E^d$ and some constant $Q>0$), then $g$ is $\alpha$-smooth with $\alpha= \frac{\beta \log 2}{d \log 3}$ on the Cantor set ${\mathcal C}\subsetneq [0,1].$     More recently, Ismayilova~\cite{ismayilova2024kolmogorov}  and Ismailov~\cite{Ismailov2024kart_uar_nn} revisited Sprecher's representation  (with $m=2d,$  the  parameters given in 
\eqref{alambda}, and a dilation $c_q=(2d+1)q$), and proved that 
if $f$ is continuous, 
discontinuous but bounded,  or
unbounded, then the outer function can be constructed to share the same properties (i.e., $g$ is continuous,   discontinuous but bounded, or unbounded, respectively).


Inspired by KART on a bounded domain, Doss~\cite{Doss1977superposition} and
Demko~\cite{Demko1977superposition} were among the earliest to study the superposition of functions on ${\mathbb R}^d$ in the form: $f(\bs x)=\sum_{q=1}^m g_q \circ \phi_q (\bs x),$  and the problem of interest was also extended to $f\in C(X)$ (the set of real continuous functions on a topological space $X$). Hattori~\cite{Hattori1993}, and 
Feng and Gartside~\cite{Feng2011} proved the {\em existence of $2d+1$ continuous functions $\phi_q$}  on every locally compact separable metric space with ${\rm dim}(X) \le d,$ which particularly holds for $X=\mathbb R^d.$ 
 In sum, the number of inner functions in the KART/variants and even more general representations: $f(\bs x)=\sum_{q=1}^m g_q \circ \phi_q (\bs x)$ must be at least $2d+1$   (see ~\cite{Sternfeld1985} and~\cite[Chapter 17]{Lorentz1996}).
Notably, Laczkovich~\cite{Laczkovich2021superposition} showed that $\phi_q$ can be chosen as   
the KART's inner structure: 
$\phi_q(\bs x)=\sum_{p=1}^d  \lambda_p \psi_q(x_p)$. In other words,  the KART is valid on $\mathbb R^d:$  {\em Let $d \geq 2$ and $m>(2+\sqrt{2}) d$ be integers, and let $\{\lambda_1, \ldots, \lambda_d\}$ be distinct positive numbers. Then there are continuous functions $\psi_1, \ldots, \psi_m \in C(\mathbb{R})$ with the following property: for every bounded $f \in C\left(\mathbb{R}^n\right),$ there is a continuous function $g \in C(\mathbb{R})$ such that}
\begin{equation}\label{Laczk-v}
f(x_1,\ldots, x_d)=\sum_{q=1}^m g\Big(\sum_{p=1}^d  \lambda_p \psi_q(x_p)\Big).
\end{equation}
Laczkovich ~\cite{Laczkovich2021superposition} also presented a modified  version with 
$m>(2+\sqrt{2})(2 d-1),$  {\em monotonically increasing inner functions $\psi_q,$} and constructed positive constants $\lambda_{p,q}$ (in place of given $\lambda_p$ in \eqref{Laczk-v}), which shares the same monotonicity with several versions in Table  \ref{Tab:KART}. 

As commonly recognized today,
 the KART/variants can be naturally translated into feed-forward neural networks with two hidden layers.   The groundbreaking transition from theory to practice is attributed to Hecht-Nielsen~\cite{HechtNielsen1987} (1987), who essentially mapped Sprecher's construction~\cite{Sprecher1965structure} to a neural network. 
Hecht-Nielsen's  network
is foundational~\cite[Chapter 5]{Sprecher2017algebra}, but
this short conference paper~\cite{HechtNielsen1987} concluded with some 
neutral (perhaps pessimistic)   views: {\em ``The Kolmogorov's mapping neural network existence theorem for approximations of functions by networks is, at least in theory, sound, but
the direct usefulness of this result  is doubtful.''}  
Girosi and  Poggio~\cite{Girosi1989} further argued that {\em   
``Kolmogorov's theorem: an exact representation is hopeless in representation properties of networks''} for at least two reasons:
\begin{itemize}
\item[(i)] the inner and outer functions lack smoothness, so the Kolmogorov's network may lose  generalization and stability against noise;
\item[(ii)] 
Kolmogorov's network is not the type of parameterized representation with modifiable/learnable parameters.
\end{itemize}
 Indeed, one theoretical evidence for the first point is the existence of differentiable functions of $d\ge 2$ variables that cannot be expressed as a superposition of differentiable functions of fewer than 
$d$ variables~\cite{Vitushkin1954}. In other words, if the KART and its variants are universal (i.e., can represent all multivariate continuous functions), 
we should not expect all the involved one-dimensional functions to be differentiable or more regular.  

On the contrary, 
K{$\dot {\rm u}$}rkov{\'a}~\cite{Kourkova1991,Kuurkova1992kolmogorov} advocated for the relevance of the KART in multilayer neural networks and asserted  that
\begin{itemize}
\item {\em One should sacrifice the exactness of representation by adopting an approximate version instead.}
\end{itemize}
Taking advantage of the fact that any continuous function on a closed interval can be approximated arbitrarily  well by  shallow neural networks with sigmoidal activation function (see e.g.,  Cybenko~\cite{Cybenko1989}),  K{$\dot {\rm u}$}rkov{\'a}~\cite{Kuurkova1992kolmogorov} introduced an approximate representation, where the resulting KART-inspired NN has the universal approximability to $C(E^d)$ (see  Theorem \ref{UAT-KAN-Gen} below). 
 In addition, Brattka~\cite{Brattka2007}
provided some deep insights into 
the computability of the KART, which admits an algorithm that, in principle, allows a Turing machine to evaluate the functions up to any prescribed precision, and where the computable version derived  was based upon Sprecher's construction and Lorentz' contraction mapping.  
Very recently, Freedman~\cite{Freedman2024proof} commented
that KART may illuminate neural network learning and set its foundation.   
In a nutshell, the astonishing representations of the KART/variants present numerous theoretical and practical issues that deserve in-depth investigation. Its mystery may cause misconceptions and misinterpretations at times~\cite{Ismailov2024kart_uar_nn}.

\subsection{Two-block Approximate Kolmogorov's Representation and Its Universality} 

\medskip

Our constructions are largely inspired  by  K{$\dot {\rm u}$}rkov{\'a}'s philosophy~\cite{Kourkova1991,Kuurkova1992kolmogorov}. The main  principle is to construct  suitable {\em approximate  Kolmogorov's representations} with the following features:
\begin{itemize}
    \item[(i)] {\em Adhere to the structure of KART or its variants;}
    \item[(ii)] {\em Establish universal approximability to all continuous functions;}
    \item[(iii)] {\em Allow to configure the neural networks simply by inner and outer blocks.}
\end{itemize}

Our starting point is to generalize the theory in  
K{$\dot {\rm u}$}rkov{\'a} 
~\cite{Kourkova1991,Kuurkova1992kolmogorov} and then construct the approximate representations. In general, we consider Kolmogorov's representation:
\begin{equation}\label{KART-01}
f(x_1, \ldots, x_d)=\sum_{q=0}^{m} g_q\Big(\sum_{p=1}^d \psi_{p, q}(x_p)\Big),
\end{equation}
where $m\ge 2d+1$,  $g_q\in C(I_g)$ and $\psi_{p,q}\in C(I_\psi).$ Note that the inner and outer functions and their domains can be adapted to different versions in Table  \ref{Tab:KART} (where $g_q$ may be the same and $\psi_{p,q}$ may be $\lambda_p\psi_q$ with given 
 $\lambda_p$, etc.). 

\begin{definition}[{\bf Set of Approximators}]\label{Ansatz}
  {\em  Define the subset of $C(E^d)$ generated  by the KART/variants: 
\begin{equation}\label{KappM}
{\mathbb K}_M^{m,d}=\Bigg\{F(\bs x)=\sum_{q=0}^{m}  G_q\Big(\sum_{p=1}^d  \Psi_{p, q}(x_p)\Big)\,:\, \forall\, G_q\in {\mathcal A}_{M_g}(I_g),\; \forall\, \Psi_{p,q}\in   {\mathcal A}_{M_\psi}(I_\psi)  \Bigg\}, 
\end{equation}
where the ansatz or parameterisation spaces ${\mathcal A}_{M_z}(I_z)$ are chosen as dense subsets of $C(I_z)$ for $z=\psi, g.$  It has the cardinality:  
\begin{equation}\label{Card-A}
P:=\# {\mathbb K}_M^{m,d}=(m+1)M_g+d(m+1) M_\psi=(m+1)(M_g+dM_\psi), 
\end{equation}
where we assume the cardinality of ${\mathcal A}_{M_z}(I_z)$ is $M_z.$}
\end{definition}

As a generalisation of the theory in K{$\dot {\rm u}$}rkov{\'a}~\cite{Kuurkova1992kolmogorov}, we can show the universal approximability of the approximate representations (see \ref{UAT-proof} for the proof).
\begin{theorem}[{\bf Universal Approximation Theorem}]\label{UAT-KAN-Gen} 
Let $d\ge 2.$ Assume that  ${\mathcal A}_{M_z}(I_z)$ are dense in $C(I_z)$ for $z=g,\psi.$ Then  
the subset  ${\mathbb K}_M^{m,d}$ defined in \eqref{KappM}  is dense in $C(E^d)$ with $E^d=[0,1]^d$ in the sense that for any $f\in C(E^d)$ and any
$\varepsilon>0,$ there exists $F\in {\mathbb K}_M^{m,d}$ {\rm(}i.e.,  
$\exists\, M_g, M_\psi \in \mathbb{N}$ depending on $\varepsilon${\rm)} such that
\begin{equation}\label{Univ-NNS-gen}
\|f- F\|_\infty=\sup_{\bs x\in E^d}|f(\bs x)-F(\bs x)|<\varepsilon,
\end{equation}
where $\bs x=(x_1,\ldots,x_d).$
\end{theorem}
In K{$\dot {\rm u}$}rkov{\'a}~\cite{Kuurkova1992kolmogorov},  ${\mathcal A}_{M_z}(I_z)$ were both chosen as the set of
staircase-like functions of  a type $\sigma$ (sigmoidal function),  that is, 
\begin{equation}\label{AMz}
{\mathcal A}_{M_z}(I_z):=\Big\{\sum_{i=1}^{M_z} a_i\, \sigma(b_i\, x+c_i)\,:\,  a_i, b_i, c_i\in \mathbb R,\; 
 x\in I_z\Big\},\quad z=g,\psi,
 \end{equation}
 which are universal approximators of $C(I_z)$
 (see e.g.,~\cite{Cybenko1989,PetersenZech2024}). In theory, one can  choose ${\mathcal A}_{M_z}(I_z)$ to be any ansatz classes consisting of orthogonal polynomials/functions (used in spectral methods), splines, wavelets and radial basis functions, among others. The interested readers may refer to~\cite{Lorentz1996} and various other resources for their universal approximability and convergence estimates. On the other hand,  one can also parameterize the one-dimensional inner/outer functions by neural networks such as 
  MLPs.
 
We note that Igelnik and Parikh~\cite{Igelnik2003} proposed the  
Kolmogorov spline network inspired by the Lorentz's version: 
\begin{equation}\label{IP2003}
F_{M}^s(\bs x)=\sum_{q=0}^M s\Big(\sum_{p=1}^d \lambda_p\, s(x_p, \gamma_{p,q}), \gamma_q\Big),
\end{equation}
where $s(\cdot, \gamma_{q})$ and $s(\cdot, \gamma_{p,q})$ are cubic splines with adjustable parameters  
$\gamma_{q}, \gamma_{p,q},$ respectively, and the constants $\lambda_p$ are given in the Lorentz's representation. Remarkably, the so-defined approximate spline network has convergence rate as follows~\cite{Igelnik2003}: {\em For any $f\in C^1(E^d)$ {\rm(}continuously differentiable functions defined on $[0,1]^d$ with bounded gradient{\rm)}, we have
\begin{equation}\label{estIP2003}
\|f-F_{M}^s\|_\infty={\mathcal O}(M^{-1}),
\end{equation}
where the number of net parameters $P={\mathcal O}(M^{3/2}).$ }
Lai and Shen~\cite{lai2021kolmogorov} studied the approximate Lorentz's representation via ReLU NN parameterisation, where a first order convergence was obtained under the assumption that the outer function is Lipschitz continuous (see Kahane~\cite{Kahane1975}). 
Lai and Shen~\cite{Lai2024optimal} attempted to estimate higher order convergence  for the spline networks, but the  differentiability and stronger regularity  must be imposed on the outer functions. Although there are some functions in such  compositions satisfying the regularity assumption (e.g., $x_1x_2\cdots x_d={\rm exp}(\ln x_1+\ln x_2+\cdots+\ln x_d)$ for all $x_i>0$), 
one may not expect such a regularity for general multi-dimensional functions. After all, the highest regularity of the outer functions by construction is Lipschitz  
(see  Table \ref{Tab:KART}). Similar assumptions were made in 
the KANs by Liu et al~\cite{liu2024kan} and some other variants for the stacked KART with an aim to approximate composite functions with more layers (e.g., one representative function in~\cite{liu2024kan}:
$f\left(x_1, x_2, x_3, x_4\right)=\exp \left(\sin \left(x_1^2+x_2^2\right)+\sin \left(x_3^2+x_4^2\right)\right)$). Nevertheless, the theory of the stacked KANs in~\cite[Theorem 2.1]{liu2024kan} holds for a specific class of functions.  


We reiterate that the inner functions in KART are independent of the functions to be represented, while the outer functions are data-dependent. This suggests the use of different ansatz or parameterisation for the inner and outer functions, leading to the two-block architecture below.




\subsection{KKAN Architecture}
\label{KART_model}
\smallskip 
\begin{figure}[t]
    \centering
    \includegraphics[width=1\linewidth]{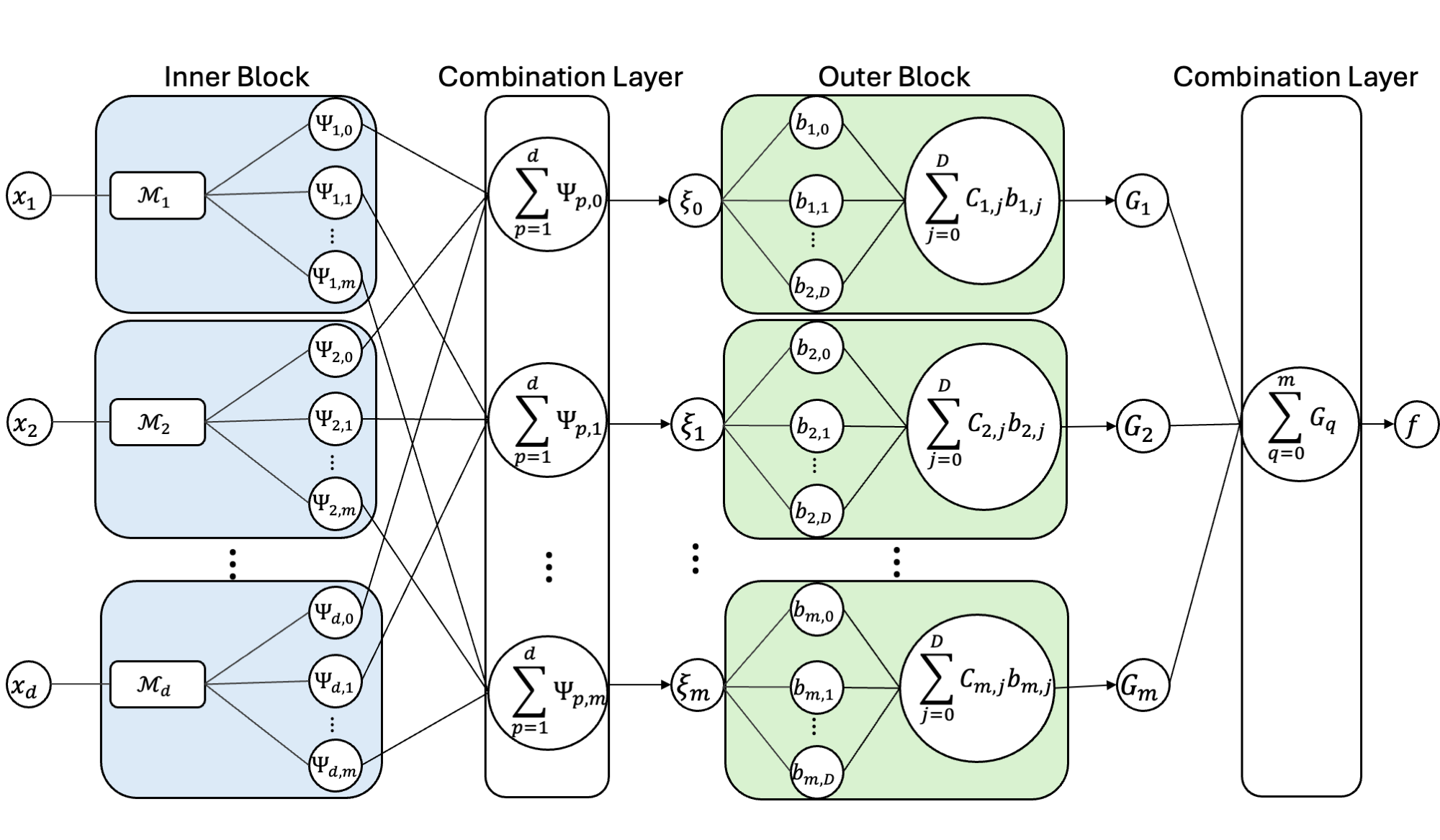}
\caption{KKAN-Inspired architecture. The inner block computes the inner functions by expanding each input dimension into an $m$-dimensional space. The first combination layer sums the inner functions across the input dimensions, i.e., $\xi_q = \sum_{p=1}^d \Psi_{p, q}(x_p)$, to obtain an $m$-dimensional vector $\bm{\xi} = [\xi_0, \ldots, \xi_m]$. The outer block computes the outer functions by transforming each $\xi_q$, and the final combination layer sums all the outer functions $G_q$, enabling the approximation of the target function, closely mimicking the KART.}
    \label{Architecture}
\end{figure}

The proposed architecture aims to closely mimic the KART variations, described as:

\begin{align}
\label{KART_arch}
f(x_1, \ldots, x_d) &= \sum_{q=0}^{m} G_q\Big(\sum_{p=1}^d \Psi_{p, q}(x_p)\Big), \\
f(x_1, \ldots, x_d) &= \sum_{q=0}^{m} G_q(\xi_q),
\end{align}

\noindent where $\Psi_{p,q}(x_p)$ and $G_q(\xi_q)$ are referred to as the \textit{inner} and \textit{outer} functions, respectively. Inspired by this representation, we divide our architecture into an \textbf{inner block}, \textbf{combination layers}, and an \textbf{outer block} (see Figure~\ref{Architecture}).

In this study, we propose defining trainable inner and outer blocks, enabling the approximation of multivariate functions with enhanced flexibility and expressiveness. The inner block computes the inner functions by expanding each input dimension into an $m$-dimensional space. The first combination layer sums the inner functions across the input dimensions, i.e., $\xi_q = \sum_{p=1}^d \Psi_{p, q}(x_p)$, to obtain an $m$-dimensional vector $\bm{\xi} = [\xi_0, \ldots, \xi_m]$. The outer block then computes the outer functions by transforming each $\xi_q$. Finally, the last combination layer sums all the outer functions $G_q$, enabling the approximation of the target function closely mimicking \eqref{KART_arch}.

Notably, the proposed architecture bears a resemblance to Tensor Neural Networks~\cite{wang2022tensor} and Separable PINNs~\cite{cho2024separable}, with the key difference being that our architecture combines dimensions via summation instead of using the tensor product.

\subsubsection{Inner Block}
\smallskip

The role of the inner block is to obtain the inner functions by expanding each input dimension into an $m$-dimensional space. Towards this end, we propose using Multi-Layer Perceptrons (MLPs) due to their flexibility, strong approximation capabilities, and continuous advancements in the deep learning community.

Additionally, we enhance the baseline MLP by drawing inspiration from the ``adaptive basis viewpoint" introduced in~\cite{cyr2020robust}. Under this perspective, an MLP is considered a mesh-free technique that constructs an adaptive basis, where the output is obtained by a linear combination of basis functions in the last linear layer. This viewpoint intuitively justifies the effectiveness of using suitable input transformations~\cite{cai2019multi, Wang2020_Fourier_nets, wang2024piratenets} to improve approximation capabilities in PIML.

\begin{figure}[t]
    \centering
    \includegraphics[width=1\linewidth]{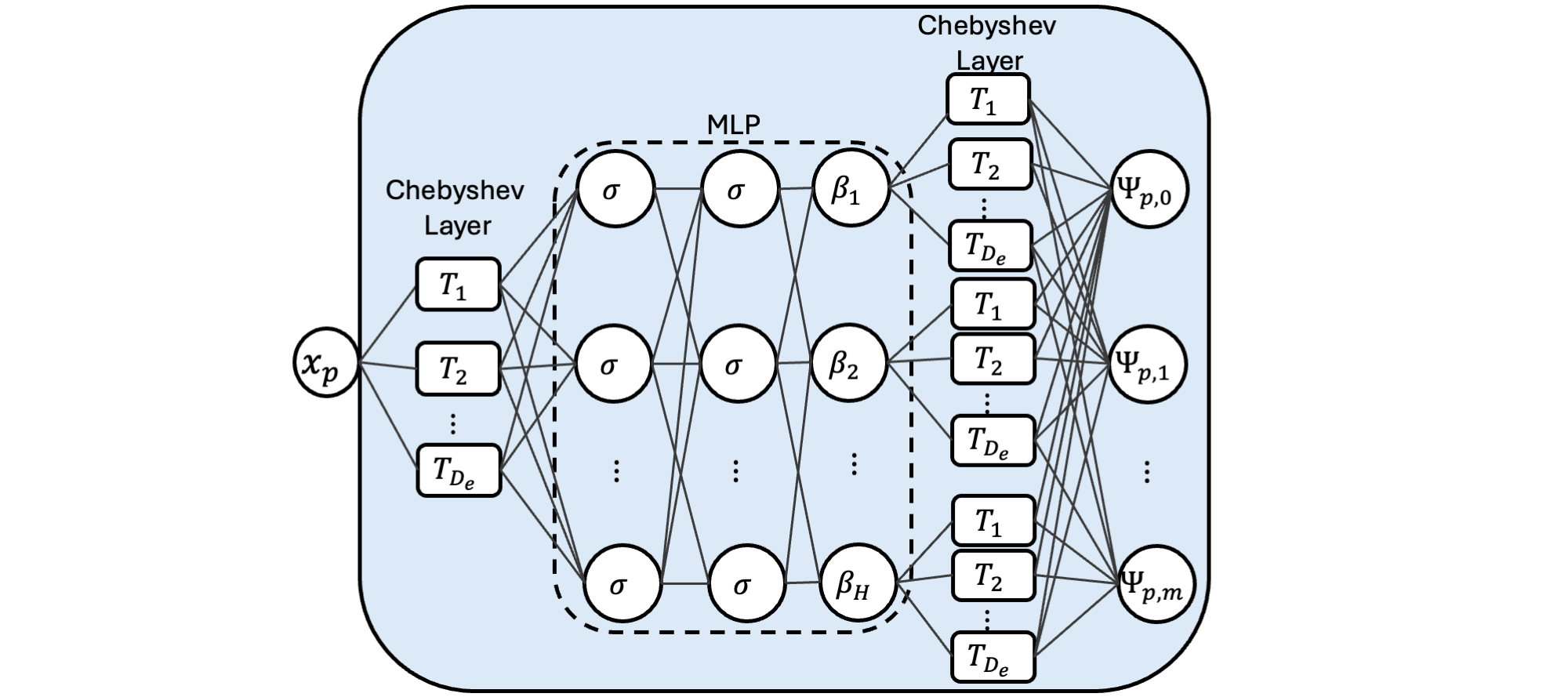}
\caption{Enhanced-basis MLP (ebMLP). Each inner block expands its respective input dimension into an $m$-dimensional space using an enhanced Multi-Layer Perceptron (MLP). The ebMLP incorporates two trainable Chebyshev layers that perform orthogonal expansions of the inputs ($x_p$) and outputs ($\beta_i$), improving the quality of the basis functions and enhancing the network's representation capabilities.}
    \label{ebMLP}
\end{figure}

To further refine the basis construction, we introduce two trainable Chebyshev layers that enable us to obtain orthogonal expansions of the inputs ($x_p$) and outputs ($\beta_i$) of the baseline MLP~\cite{toscano2024pinns}. This architecture, which we denote as the enhanced-basis MLP (ebMLP), is used to construct each component of our inner block (see Figure~\ref{ebMLP}). By incorporating orthogonal basis functions through the Chebyshev layers, we increase the network's representation capabilities and improve its ability to approximate complex functions more effectively.

\subsubsection{Outer Block} 
\smallskip

For the outer block, we follow the approach in~\cite{liu2024kan} and model each outer function as a linear combination of basis functions (see Figure~\ref{Architecture}). Specifically, each outer function is defined as:

\begin{align*}
    g_q(\xi_q) &= \sum_{j=0}^D C_{q,j} \, b_{q,j}(\xi_q),
\end{align*}

\noindent where $b_{q,j}(\xi_q)$ are suitable basis functions, $C_{q,j}$ are the corresponding trainable coefficients, and $D$ is the expansion degree. By selecting appropriate basis functions and making the coefficients trainable, the outer blocks can flexibly model the nonlinear transformations required to accurately approximate the target function.

By combining the strengths of the inner and outer blocks, the architecture effectively captures the underlying structure described by the KART (i.e., equation \ref{KART_arch}). This leads to enhanced flexibility and expressiveness in approximating multivariate functions. A detailed description of the proposed implementation and the basis functions considered in this study is provided in ~\ref{KKAN}.

\section{Methodology}
\label{sec_methods}
This study compares three representation models—KKANs, cKAN (i.e., KANs~\cite{liu2024kan} with Chebyshev basis functions), and MLPs—for function approximation, physics-informed machine learning (PIML), and data-driven Neural Operators (NOs). The trainable parameters for all models, denoted as \(\theta\), are optimized by minimizing a case-specific loss. Detailed formulations for KKANs, MLPs, and cKANs are provided in Section~\ref{KART_model}, ~\ref{MLP_arch} and~\ref{KANs_Arch},~\ref{KKAN}.

\subsection{Function Approximation}
\smallskip

The goal of this test is to evaluate the ability of representation models to fit a specific function. Here, the function \(\hat{u}(x)\) is approximated by a representation model \(u(\theta, x)\), where \(\theta\) are the trainable parameters:
\begin{align}
  \hat{u}(x) \approx u(\theta, x), \quad x\in\Omega.
\end{align}

The training process minimizes the data residuals:
\begin{align}
\label{data_res}
    r_D(x,\theta) &= u(x,\theta) - \hat{u}(x), \quad x \in \Omega_D,
\end{align}
where \(\Omega_D \subset \Omega\) contains all available observations. The corresponding loss function is:
\begin{equation}
    \mathcal{L}=\mathcal{L}_D = \sum_{i=1}^{N} \lambda_{D,i} r_D^2(x_i, \theta), \quad x_i\in \Omega_D,
\label{FA_loss}
\end{equation}
where \(\lambda_{D,i}\) are local weights that balance the contribution of each training point~\cite{anagnostopoulos2024learning}.

\subsection{Physics-Informed Machine Learning and Operator Learning}
\smallskip

SciML is agnostic to specific governing laws, leveraging machine-learning-based models to approximate solutions~\cite{toscano2024pinns}. We consider the general nonlinear ODE/PDE system:
\begin{subequations}\label{eq:problem}
  \begin{align}
    \mathcal{F}_\tau[\hat{u}](x)  &= f(x), \quad x \in \Omega, \\
    \mathcal{B}_\tau[\hat{u}](x)  &= b(x), \quad x \in \Omega_B,
  \end{align}
\end{subequations}
where \(x\) represents the spatial-temporal coordinate, \(\hat{u}\) is the solution, \(\tau\) are the parameters, and \(\mathcal{F}\) and \(\mathcal{B}\) are nonlinear differential and boundary operators, respectively.

In this study, we focus on two SciML approaches: Physics-Informed Machine Learning (PIML) and Neural Operators (NOs). The main difference between PIML and NO is that the former targets solving one specific ODE/PDE, in which the training of the representation model gives an approximated solution that maps point to point, while the latter aims to solve a family of ODEs/PDEs, in which the representation model maps functions to functions~\cite{shukla2024comprehensive}.

\subsubsection{Physics-Informed Machine Learning (PIML)}
\smallskip

PIML approximates the solution \(\hat{u}(x)\) of a PDE/ODE using a representation model \(u(\theta, x)\), ensuring that it satisfies the governing equations and any available data coming from boundary conditions, initial condition or sparse observations inside the domain. The equation and boundary residuals are described as follows:
\begin{align}
\label{PDE_res}
    r_E(x,\theta) &= \mathcal{F}_{\tau}[u](x,\theta) - f(x), \quad x \in \Omega, \\
\label{bcs_res}
    r_B(x,\theta) &= \mathcal{B}_{\tau}[u](x,\theta) - b(x), \quad x \in \Omega_B.
\end{align}

The general total loss function combines residuals for governing equations (\(\mathcal{L}_E\)), boundary conditions (\(\mathcal{L}_B\)) and data (\(\mathcal{L}_D\)) 
\begin{align}
\label{PIML_loss}
    \mathcal{L} &= m_E\mathcal{L}_E + m_B\mathcal{L}_B+m_D\mathcal{L}_D,
\end{align}
\noindent where each term is computed as:
\begin{align}
\mathcal{L}_{\alpha}(\theta) &= \sum_{i=1}^{N_{\alpha}} \lambda_{\alpha,i} f(r_\alpha(x_i, \theta)), \quad x_i \in \Omega_\alpha.
\label{loss_dic}
\end{align}
Here, \(\alpha=\{E,B,D\}\) indexes the loss groups for data, boundary, and equation respectively, \(m_\alpha\) are global weights that balance the contribution of each term, and \(\lambda_{\alpha,i}\) are local weights that balance the contribution for each training point. For inverse problems, equation \eqref{PIML_loss} incorporates the data residuals \(r_D\), while for forward problems, only the boundary (\(r_B\)) and equation (\(r_E\)) residuals are considered~\cite{toscano2024pinns}.
.

\subsubsection{Neural Operators (NOs)}
\smallskip

Neural Operators (NOs) approximate solution operators \( G_\theta \) that map input functions, such as a source term \( f \), to the corresponding solution \( u \)~\cite{lu2021learning, li2020fourier}. In this study, we focus on variations of DeepONet~\cite{lu2019deeponet} and QR-DeepONet~\cite{lee2024training}. A detailed description of these formulations is provided in Appendix~\ref{DeepONet_appendix}. The loss function for the DeepONet models can be described as:
\begin{equation}
\label{NO_loss}
    \mathcal{L}= \mathcal{L}_D = \frac{1}{N} \sum_{i=1}^N\frac{1}{N_u^i}\sum_{j=1}^{N^i_u} \|G_\theta(\mathbf{v}_i)(x_i^j) - u_i^j\|^2,
\end{equation}
where \(\{\mathbf{v}_i, \{x_i^j, u_i^j\}_{j=1}^{N_u^i}\}_{i=1}^N\) are training data. Here, \(N\) is the number of data pairs, \(\mathbf{v}_i\) represents the input functions, and \(u_i^j\) denotes observations of the output function at points \(x_i^j\).

\subsection{Training}
\smallskip

In this study, we learn the model parameters for all our examples by iteratively minimizing their respective loss functions (i.e., \eqref{FA_loss},\eqref{PIML_loss},\eqref{NO_loss}) using gradient-based optimizers:
\begin{align}
\label{General_optimizer}
\theta^{k+1} &= \theta^{k} + \alpha^{k} p^{k}, \\
\label{update_dir}
p^{k} &= -H_k \nabla_{\theta}\mathcal{L}(\theta^k),
\end{align}
where \(\alpha^k\) is the step size and \(H_k\) defines the update direction. Common optimizers include ADAM~\cite{kingma2014adam}, when $H_k=I$, and L-BFGS~\cite{liu1989limited}, with the latter achieving superlinear convergence by approximating the Hessian i.e., $H_k\approx(\nabla^2_{\theta}\mathcal{L})^{-1}$ ~\cite{urban2024unveiling}.

\subsection{Learning Dynamics via the Information Bottleneck Theory}
\label{IB-SNR}
\smallskip

The IB theory provides an information-theoretic framework for analyzing the training and performance of neural networks. It explains how networks form a compressed representation of layer activations, \( \mathcal{T} \), with respect to an input variable \( x \in \mathcal{X} \), retaining only relevant information about the output variable \( y \in \mathcal{Y} \)~\citep{tishby2000information, tishby2015deep}. 

The key concept in IB theory is the mutual information \( I(x, y) \), which quantifies how much information about \( y \) is preserved in the representation \( x \). Optimal models balance this tradeoff by discarding irrelevant information, creating an ``information bottleneck." IB identifies two primary learning stages: fitting and diffusion~\citep{shwartz2017opening, goldfeld2020information, shwartz2022information}, with a third stage, transition, proposed in~\citep{anagnostopoulos2024learning} for PIML. These three stages have been observed across various representation models, including PINNs and physics-informed KANs~\citep{shukla2024comprehensive}.

\paragraph{Signal-to-Noise Ratio (SNR)}

The signal-to-noise ratio (SNR) is a critical metric for understanding training dynamics. It is defined as:
\begin{equation}
\label{SNR_Eq}
\text{SNR} = \frac{\lVert \mu \rVert_{2}}{\lVert \sigma \rVert_{2}} = \frac{\lVert \mathbb{E}[\nabla_{\theta}{\mathcal{L}_{\mathcal{B}}}] \rVert_{2}}{\lVert \text{std}[\nabla_{\theta}{\mathcal{L}_{\mathcal{B}}}] \rVert_{2}},    
\end{equation}
\noindent where \( \theta \) represents the network parameters, \( \mu \) is the batch-wise mean of the gradients, and \( \sigma \) is the batch-wise standard deviation of the gradients of the batch-wise loss \( \mathcal{L}_{\mathcal{B}} \). High SNR indicates that the gradients are signal-dominant, while low SNR corresponds to noise-dominant gradients~\citep{goldfeld2020information, shukla2024comprehensive}. 
This IB-based framework not only provides a lens to analyze convergence but also offers insights into why some models generalize better than others. Models that successfully transition through all three stages, particularly those entering diffusion early, tend to exhibit superior performance~\cite {anagnostopoulos2024learning, shukla2024comprehensive}. Conversely, models trapped in the transition stage generally do not converge~\cite{toscano2024pinns}.

\subsection{Self-scaled Residual-Based Attention (ssRBA)}
\label{ssRBA_section}
\smallskip

One challenge in training neural networks is that point-wise residuals can be neglected in cumulative loss computations~\cite{mcclenny2023self}. To address this issue, various methods have been developed to balance the point-wise contributions using local $\lambda_i$, which enhance the performance in PIML~\cite{mcclenny2023self,hao2024structure,basir2022physics,basir2022investigating,basir2023adaptive,son2023enhanced,anagnostopoulos2024residual,song2024loss,shukla2024comprehensive,ramirez2024residual,chen2024self}. In particular, Residual-Based Attention (RBA)~\cite{anagnostopoulos2024residual} experimentally have been shown to be effective in a wide range of applications and extensions~\cite{toscano2024inferring,toscano2024inferring_AIV,ramireza2024residual,wang2024aspinn,ramirez2025residual,wang2024general,chen2024self,shukla2024comprehensive,anagnostopoulos2024learning,rigas2024adaptive} due to its simplicity, efficiency and accuracy. RBA uses the exponentially weighted moving average of residuals to adaptively prioritize high-error regions during training, significantly improving the model performance with minimal computational overhead.

The RBA weights \( \lambda_{\alpha,i} \), defined for loss term \( \alpha \) and point \( x_i \), are updated iteratively as:
\begin{equation}
 \lambda_{\alpha,i}^{(k+1)} \leftarrow \gamma\lambda_{\alpha,i}^{(k)}+\eta \frac{r_{\alpha,i}^{(k)}}{\lVert \bm{r}_\alpha^{(k)} \rVert_{\infty}}, \quad i \in \{0, 1, \dots, N\},
 \label{Update_RBA}
\end{equation}
where \( r_{\alpha,i} \) is the residual at \( x_i \), \( \eta \) is the learning rate, and \( \gamma \) is a memory term controlling the influence of past residuals, and bounding each multiplier as \( \lambda_{i} \leq \lambda_{\max} = \eta / (1 - \gamma) \). This attention mechanism focuses optimization on regions with high error, enhancing convergence efficiency~\cite{anagnostopoulos2024learning}.

To handle large datasets requiring batch-wise training, Toscano et al.~\cite{toscano2024inferring} proposed Residual-Based Attention with Resampling (RBA-R). Here, RBA weights define a probability density function (PDF) for resampling critical points:
\begin{equation}
   p_{\alpha}^{(k)}(\textbf{x})=\frac{(\bm{\lambda}_{\alpha}^{(k)})^{\nu}}{\mathbb{E}[(\bm{\lambda}_{\alpha}^{(k)})^{\nu}]}+c,
 \label{Update_pdf}
\end{equation}
where \( \bm{\lambda}_{\alpha}^{(k)}=\{\lambda^k_{\alpha,i
}\}_{i=1}^{N_{\alpha}} \) represents RBA weights raised to the power \( \nu \), which controls the standard deviation of the PDF, and \( c > 0 \) which ensures all points are eventually sampled. Unlike prior methods~\cite{lu2021deepxde, wu2023comprehensive}, this PDF, based on cumulative residuals, offers greater stability and computational efficiency, enabling fast sampling with negligible cost.

In this study, we propose extending the RBA method, drawing inspiration from our insights on information bottleneck theory and the signal-to-noise ratio (SNR). As discussed in Section~\ref{IB-SNR}, optimal convergence occurs during the diffusion stage when the SNR is high, indicating an ``agreement" or ``equilibrium" in the gradient flow. However, prior studies~\cite{anagnostopoulos2024learning,shukla2024comprehensive} demonstrate that the diffusion stage may saturate, leading to a decline in the SNR and a plateau in the generalization error. This saturation may arise due to increasing stochasticity in the later diffusion stages and diminishing gradients as the training loss decreases, resulting in machine precision limitations.

To address these challenges, we propose sequentially increasing the magnitude of the memory term $\gamma$. By increasing $\gamma$, the model ``remembers" more information from past iterations, which is crucial for extracting the mean behavior in highly stochastic regimes such as the diffusion stage. To implement this strategy, we split the training process into several stages with $N_{stage}$ iterations per stage, progressively increasing $\gamma$, thereby inducing a higher upper bound $\lambda_{\max}$.

However, for multi-objective loss functions such as PIML (i.e., \eqref{PIML_loss}), modifying $\lambda_{\max}$ could introduce imbalances between individual loss terms (e.g., $\mathcal{L}_B$, $\mathcal{L}_E$). To address this, prior studies have proposed adaptive strategies, such as modifying global weights~\cite{wang2021understanding,jin2021nsfnets,boster2023artificial,cai2021artificial,xiang2022self,liu2021dual,basir2022investigating,wang2022respecting,chen2024self} or refining update directions using cosine similarities and related methods~\cite{liu2024config,zhou2023generic,yao2023multiadam}. Building on these approaches, we propose a combined method that leverages global weights to scale loss-specific gradients based on their magnitudes, resulting in an improved update direction.

Notice that for first-order optimizers such as ADAM, the update direction (i.e., equation \eqref{update_dir}) for PIML (i.e., equation \eqref{PIML_loss}) is given by:
\begin{align}
    p^k &= -m_E\nabla_{\theta}\mathcal{L}_E(\theta^{k}) - m_B\nabla_{\theta}\mathcal{L}_B(\theta^{k}) - m_D\nabla_{\theta}\mathcal{L}_D(\theta^{k}),
\end{align}
where $\nabla_{\theta}\mathcal{L}_{E}$, $\nabla_{\theta}\mathcal{L}_{B}$, and $\nabla_{\theta}\mathcal{L}_{D}$ are the loss gradients which can be represented as high-dimensional vectors defining directions to minimize their respective loss terms. Notice that if the gradient magnitudes are imbalanced, one direction will dominate, which may lead to poor convergence (See Figure~\ref{Grads_dirs}(a)).  To address this challenge, we propose modifying the magnitude of the individual directions by scaling their respective global weights. In particular, we fix $m_E$ and update the remaining global weights using the rule:
\begin{align*}
    m_B^k &= \alpha m_B^{k-1} + (1-\alpha) \frac{\|\nabla_{\theta}\mathcal{L}_E\|}{\|\nabla_{\theta}\mathcal{L}_B\|}, \\
    m_D^k &= \alpha m_D^{k-1} + (1-\alpha) \frac{\|\nabla_{\theta}\mathcal{L}_E\|}{\|\nabla_{\theta}\mathcal{L}_D\|},
\end{align*}
where $\alpha \in [0,1]$ is a stabilization parameter~\cite{wang2021understanding}. This formulation computes the iteration-wise average ratio between gradients, enabling normalized scaling, which, on average, allows us to define a balanced update direction $\hat{p}^k$ (See Figure~\ref{Grads_dirs}(b)):
\begin{align}
    \hat{p}^k &\approx -m_E\|\nabla_{\theta}\mathcal{L}_E\| \left[ \nabla_{\theta}\mathcal{L}_E(\theta^{k}) - \frac{\nabla_{\theta}\mathcal{L}_B(\theta^{k})}{\|\nabla_{\theta}\mathcal{L}_B\|} - \frac{\nabla_{\theta}\mathcal{L}_D(\theta^{k})}{\|\nabla_{\theta}\mathcal{L}_D\|} \right].
\end{align}

Under this approach, all loss components have balanced magnitudes, allowing each optimization step to minimize all terms effectively.

A generalized training procedure for optimizing representation models using gradient descent and self-scaled residual-based attention is summarized in Algorithm~\ref{ssRBA_alg}. 

\subsection{Evaluation Metrics}
\smallskip

In the current study, we analyze our models (i.e., MLP, cKAN, and KKAN) under three criteria, namely (1) the relative \( L^2 \) error on the ground truth data, (2) their geometric complexity, which has been linked to generalization capabilities, and (3) the learning dynamics using the IB theory.

\subsubsection{Relative \( L^2 \) Error}
\smallskip

The relative \( L^2 \) error is used to benchmark the accuracy of the final predictions from different representation models. It is defined as:
\begin{equation}
\text{Relative } L^2 = \frac{\lVert \text{Reference} - \text{Predicted} \rVert_{2}}{\lVert \text{Reference} \rVert_{2}},
\end{equation}
where \( \text{Reference} \) denotes the ground truth, obtained either analytically (e.g., for function approximation tasks) or from high-accuracy numerical solvers (e.g., for SciML problems). This metric quantifies the deviation of the model's predictions from the expected solution, providing a clear measure of model performance.

\subsubsection{Geometric Complexity}
\smallskip

Simpler models are generally preferred over more complex ones, and controlling model complexity has been a long-standing goal in machine learning through techniques like regularization, hyperparameter tuning, and architecture design. In this study, we evaluate model complexity using the notion of geometric complexity, as defined in~\cite{dherin2022neural}. Geometric complexity measures the variability of a model’s function using the discrete Dirichlet energy:
\begin{align}
\label{geometric_complexity_eq}
    \langle f_{\theta}, D \rangle = \frac{1}{|D|} \sum_{x \in D} \|\nabla_x f_{\theta}(x)\|_F^2,
\end{align}
where \( f_{\theta} \) is the model's output, \( D = \{x_i\}_{i=1}^N \) is the training dataset, and \( \|\nabla_x f_{\theta}(x)\|_F \) is the Frobenius norm of the Jacobian of the model’s output with respect to its input.

This metric has been connected to various standard training heuristics, such as parameter normalization, spectral norm constraints, and noise regularization. Furthermore, it has been used to study initialization strategies and phenomena like double descent~\cite{dherin2022neural}. By analyzing the geometric complexity, we aim to understand its relationship with the generalization capabilities of different representation models.

\section{Results}
\label{Results}
\subsection{Function Approximation}
\smallskip

These models were trained using the loss function described in \eqref{FA_loss}. To ensure a fair comparison, we aimed to match the total number of parameters across all models. However, it was observed that cKANs could not handle a significantly higher number of parameters without degrading performance, thereby limiting their scalability. Consequently, smaller networks were used for all models. 

For all cases, the model's performance was evaluated on a $256 \times 256$ uniform grid, and training was performed using 10,000 points sampled with Latin hypercube sampling to ensure uniform coverage of the domain. Additional implementation details and hyperparameters are provided in Section~\ref{Imp_Details}.
\subsubsection{Discontinuous}

\begin{figure}[H]
    \centering
    \includegraphics[width=1\linewidth]{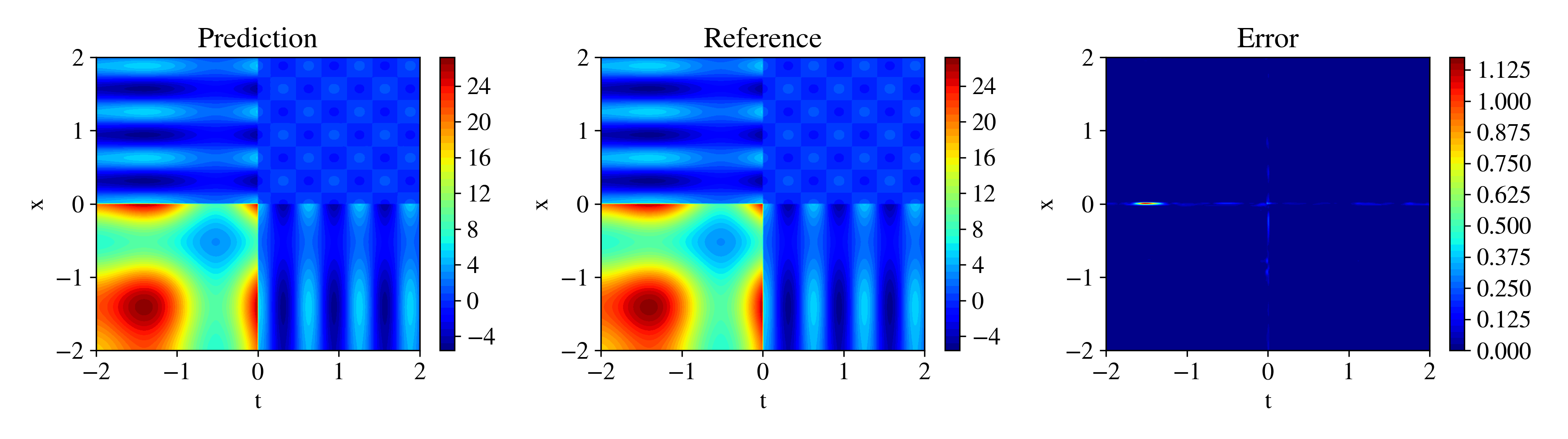}
\caption{Performance of KKANs for discontinuous function approximation. Columns show predictions, ground truth references, and absolute errors, respectively. This function is particularly challenging to learn due to two discontinuities at $x_1=0.0$ and $x_2=0.0$, along with smooth regions containing relatively high frequencies. Additionally, the function exhibits a wide range of magnitudes, with outputs spanning from $-5$ to $25$. The KKAN model achieves a relative $L^2$ error of $5.86 \times 10^{-3}$.}
    \label{fig:HD_Results}
\end{figure}

To evaluate the robustness of the analyzed models, we tested their performance on a highly discontinuous function inspired by~\cite{shukla2024comprehensive}. Originally introduced as a challenging 1D example, we extended it to 2D using a tensor product formulation:
\[
f(x_1, x_2) = \prod_{i=1}^2 h(x_i),
\]
\noindent where:
\[
h(x_i) = 
\begin{cases}
5 + \sum_{k=1}^4 \sin(k x_i), & x_i < 0.5, \\
\cos(10 x_i), & x_i \geq 0.5.
\end{cases}
\]

This problem introduces abrupt changes in the function (two discontinuities along $x_1=0$ and $x_2=0$), making it particularly challenging for models trained using gradient-based methods. For the KKAN model, we used the sin-series basis function introduced in~\cite{guilhoto2024deeplearningalternativeskolmogorov}. The relative $L^2$ error convergence and geometric complexity for other basis functions are shown in Figure~\ref{fig:Disc_Other_basis}, highlighting the robustness of our approach. While the sin-series basis performs best, the model converges effectively with all the tested cases.

\begin{figure}
    \centering
    \includegraphics[width=1\linewidth]{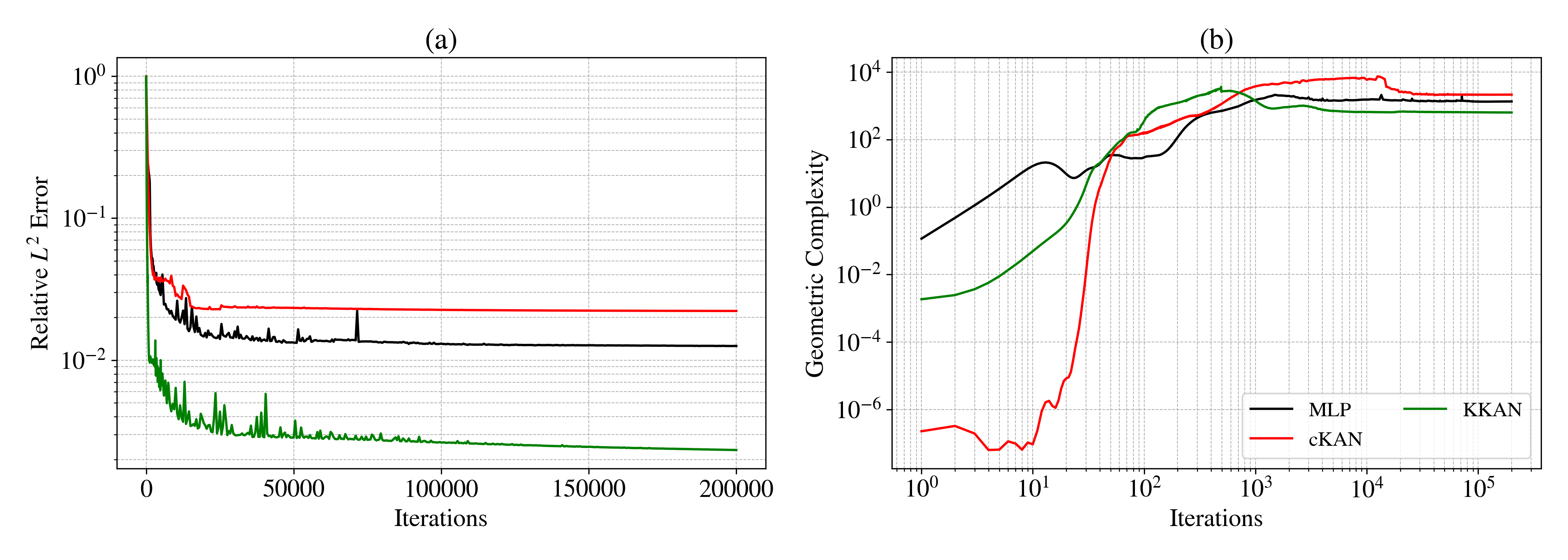}
\caption{Results for discontinuous function approximation. (a) Relative $L^2$ error convergence on the testing dataset, evaluated on a uniform $256\times256$ mesh. KKANs converge significantly faster than MLPs, achieving a relative $L^2$ error of $5.86\times10^{-3}$ after 200,000 ADAM iterations. cKANs converge slightly faster than MLPs initially but start to overfit after several iterations, as indicated by a sudden increase in the test error. (b) Geometric complexity evolution during training. Geometric complexity, represented by the discrete Dirichlet energy, reflects the gradient of the function with respect to its inputs. For this case, the geometric complexity is significantly higher for all models due to the two discontinuities in the function, which introduce sharp changes and amplify gradient variations. Initially, cKANs exhibit lower complexity than the other methods. However, their final complexity is significantly higher, indicating overfitting. In contrast, KKANs maintain the lowest complexity throughout training, contributing to their superior generalization and performance.}
    \label{fig:HD_RL2}
\end{figure}

We present the results for this function in Table~\ref{tab:model_comparison_discontinuous} and Figure~\ref{fig:HD_RL2}. KKANs demonstrated significantly better performance on the testing dataset, achieving a relative $L^2$ error of $5.86 \times 10^{-3}$, outperforming both MLPs and cKANs. As shown in Figure~\ref{fig:HD_RL2}(a), KKANs converge significantly faster than MLP and cKAN.

As shown in Table~\ref{tab:model_comparison_discontinuous}, all models exhibited comparable speeds during training, with MLPs being slightly faster, averaging 2.36 ms per iteration compared to 2.64 ms for cKANs and 2.77 ms for KKANs. Figure~\ref{fig:HD_RL2}(b) highlights the evolution of geometric complexity during training. Initially, cKANs exhibit lower complexity than other methods, but their final complexity is slightly higher, potentially indicating overfitting, as lower convergence has been associated with better generalization capabilities~\cite{dherin2022neural}. In contrast, KKANs display the lowest complexity at the end of training, which aligns with better generalization and contributes to their robust performance. These results underscore KKANs' ability to balance efficiency, accuracy, and robustness in approximating highly discontinuous functions.

\begin{table}[H]
    \centering
    \begin{tabular}{lccc}
        \toprule
        \textbf{Model} & \textbf{N. Params} & \textbf{Time (ms/it)} & \textbf{Rel. $L^2$ Error} \\
        \midrule
        MLP       & 40801 & 2.36 & $1.26 \times 10^{-2}$ \\
        cKAN      & 39360 & 2.64 & $2.22 \times 10^{-2}$ \\
        \textbf{KKAN} & 40302 & 2.77 & $\mathbf{5.86 \times 10^{-3}}$ \\
        \bottomrule
    \end{tabular}
\caption{Comparison of models for function approximation of a discontinuous function. KKANs significantly outperform both MLPs and cKANs. The training times are comparable across models. The challenging nature of the discontinuous function, characterized by abrupt changes, amplifies the difficulty for all models, highlighting KKANs’ robustness and ability to generalize better under such conditions.}
    \label{tab:model_comparison_discontinuous}
\end{table}

\subsubsection{Smooth}
\begin{figure}[H]
    \centering
    \includegraphics[width=1\linewidth]{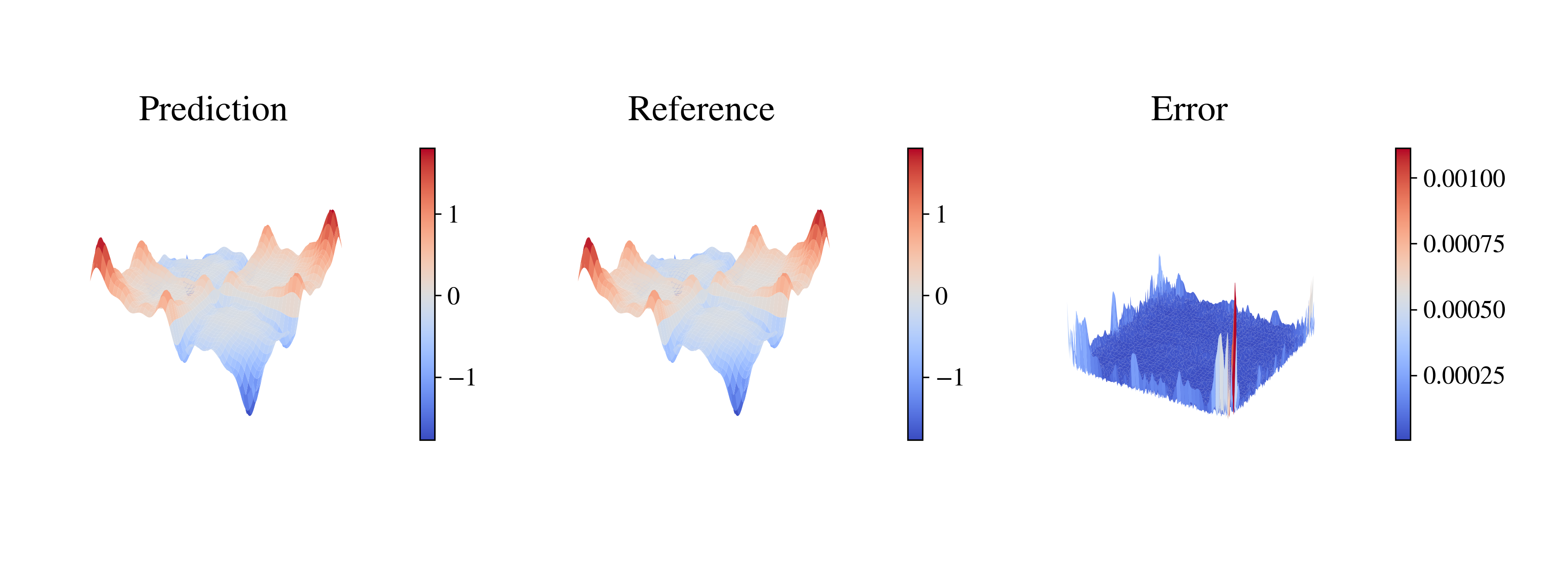}
\caption{Performance of KKANs+ssRBA for smooth function approximation. Columns show predictions, ground truth references, and absolute errors, respectively. This smooth function is challenging due to its rapidly varying gradients. The inclusion of ssRBA enhances convergence and accuracy, enabling the KKAN model to achieve a relative $L^2$ error of $1.75 \times 10^{-4}$.}
    \label{fig:smooth_Results}
\end{figure}

In the second example, we present results for a smooth oscillatory function. This example was introduced in~\cite{ainsworth2021galerkin} as a challenging benchmark for MLP due to the combination of frequencies. Similar to the previous case, we extended it to 2D using a tensor product formulation:
\[
f(x_1, x_2) = \prod_{i=1}^2 h(x_i),
\]
\noindent where
\[
h(x_i) = \sin(x_i)+\frac{1}{3}\sin(3\pi x_i)+\frac{1}{5}\sin(5\pi x_i)+\frac{1}{7}\sin(7\pi x_i).
\]

 For the KKAN model, we use a Chebyshev-grid basis function for the outer blocks. In this example, a polynomial degree of $D=15$ is used to demonstrate KKANs' ability to handle high polynomial orders effectively. In contrast, cKANs exhibit performance degradation for $D>7$, as noted in previous studies~\cite{shukla2024comprehensive}. Specific details about this basis function, along with others, are provided in Section~\ref{Basis}. The relative $L^2$ error convergence and geometric complexity for other basis functions are shown in Figure~\ref{fig:Smooth_Other_basis}, further demonstrating KKANs' robustness across different configurations.

\begin{figure}
    \centering
    \includegraphics[width=1\linewidth]{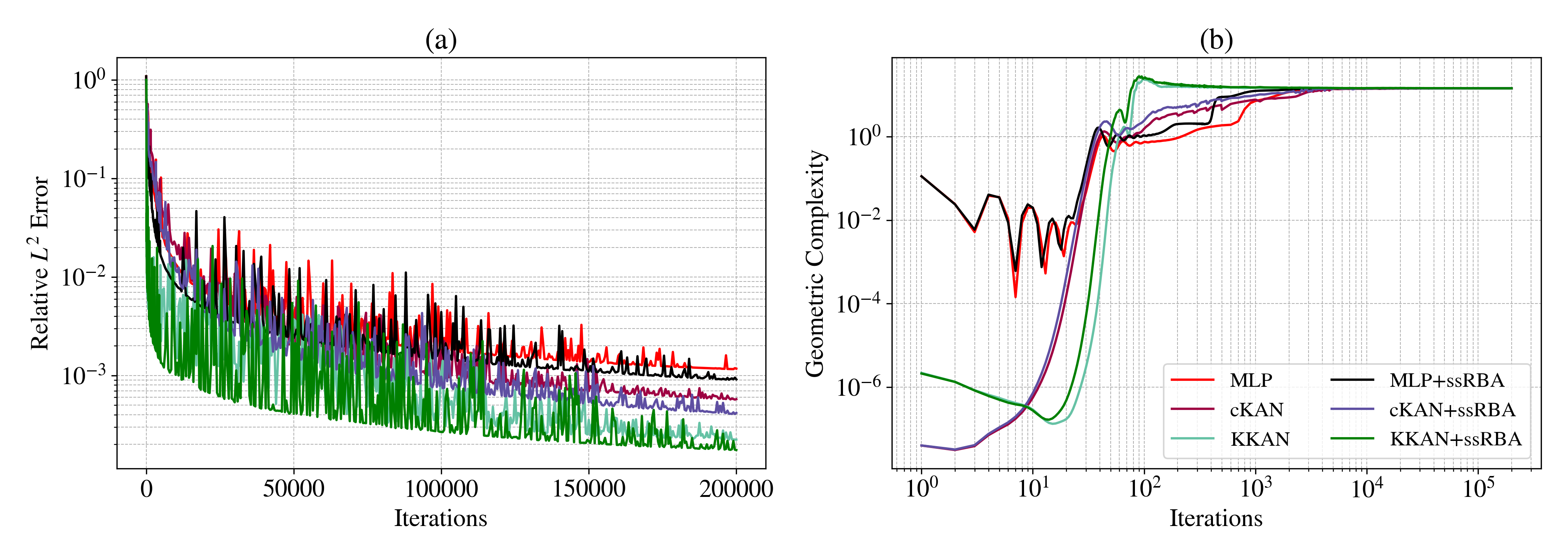}
\caption{Results for smooth function approximation. (a) Relative $L^2$ error convergence on the testing dataset, evaluated on a uniform $256\times256$ mesh. KKANs converge significantly faster than both MLPs and cKANs. The proposed ssRBA improves the performance of all representation models by enhancing accuracy and accelerating convergence. The best-performing model, KKANs+ssRBA, achieves a relative $L^2$ error of $1.75\times10^{-4}$. (b) Geometric complexity evolution during training. Both cKANs and KKANs exhibit lower geometric complexity at the start of training, reflecting their simplicity in the early stages. However, by the end of training, all models converge to a similar geometric complexity, indicating a shared characteristic among all representation models.}
    \label{fig:Smooth_RL2}
\end{figure}

Figure~\ref{fig:smooth_Results} illustrates the performance of KKANs+ssRBA for smooth function approximation, showing predictions, ground truth references, and absolute errors. The smooth function, characterized by rapidly varying gradients, poses a challenge for accurate approximation. The inclusion of ssRBA significantly improves convergence and accuracy, enabling KKANs to achieve a relative $L^2$ error of $1.75 \times 10^{-4}$. The relative $L^2$ error convergence and geometric complexity evolution during training are presented in Figure~\ref{fig:Smooth_RL2}. KKANs converge significantly faster than both MLPs and cKANs, with the ssRBA mechanism enhancing the performance of all models by accelerating convergence and improving accuracy. Both cKANs and KKANs start with lower geometric complexity, reflecting their simplicity in early training. However, by the end of training, all models converge to a similar geometric complexity, indicating a shared characteristic among the representations. Table~\ref{tab:model_comparison_smooth} provides a detailed comparison, showing that KKANs consistently outperform cKANs and MLPs in accuracy, both with and without ssRBA, while maintaining comparable training times across all models.

\begin{table}[H]
    \centering
    \begin{tabular}{lccc}
        \toprule
        \textbf{Model} & \textbf{N. Params} & \textbf{Time (ms/it)} & \textbf{Rel. $L^2$ Error} \\
        \midrule
        MLP       & 40801 & 2.38 & $1.17 \times 10^{-3}$ \\
        cKAN      & 45900 & 3.23 & $5.75 \times 10^{-4}$ \\
        KKAN & 40842 & 2.88 & $2.234 \times 10^{-4}$ \\
        \midrule
        MLP +ssRBA      & 50049 & 2.32 & $9.14 \times 10^{-4}$ \\
        cKAN +ssRBA     & 45900 & 3.26 & $4.17 \times 10^{-4}$ \\
        \textbf{KKAN+ssRBA} & 40842 & 3.09 & $\mathbf{1.74 \times 10^{-4}}$ \\
        \bottomrule
    \end{tabular}
\caption{Comparison of models for smooth function approximation. The performance of all models improves with the addition of ssRBA. The training times are comparable across models. KKANs consistently outperform cKANs and MLPs in accuracy, both with and without ssRBA, demonstrating their robustness and efficiency for smooth function approximation.}
    \label{tab:model_comparison_smooth}
\end{table}

\subsection{Physics-Informed Machine Learning}
\smallskip

In this study, we compare the performance of KKANs, cKANs, and MLPs for PIML tasks. For KKANs, radial basis functions (RBFs) are used as the outer blocks. The PDE is trained using 25,600 collocation points, with initial conditions imposed on 201 points. Periodic boundary conditions are enforced as hard constraints through architecture modifications. The models are evaluated on a fine grid of $501\times201$, and the exact solution is obtained from established benchmark studies~\cite{mcclenny2023self}. Additional implementation details are provided in~\ref{Imp_Details}.
\subsubsection{Allen-Cahn Equation}
\begin{figure}[H] \centering \includegraphics[width=1\linewidth]{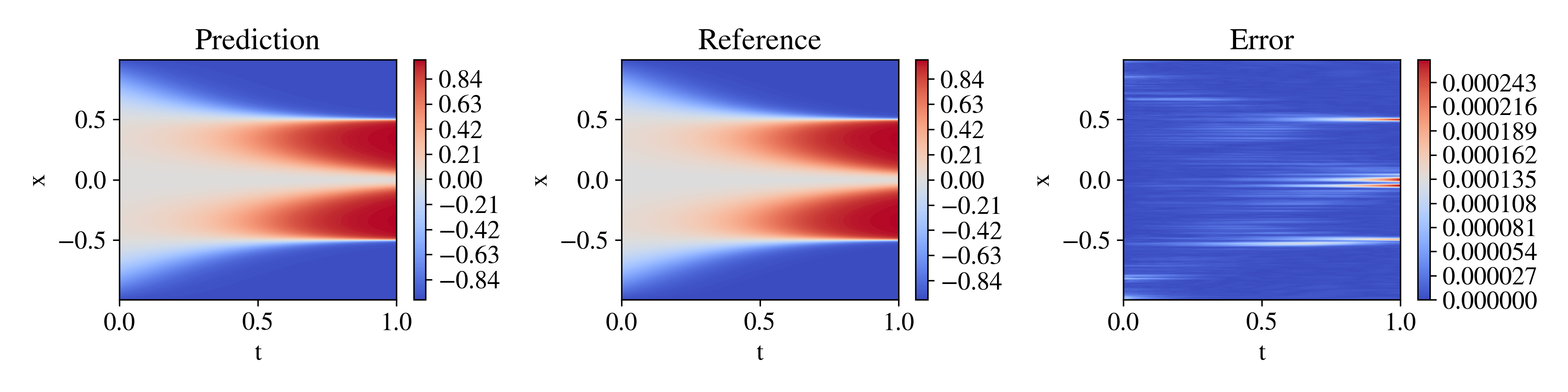} \caption{Performance of KKAN+ssRBA for solving the Allen-Cahn equation. The columns display predictions, ground truth references, and absolute errors, respectively. The model achieves a relative $L^2$ error of $2.28 \times 10^{-5}$ after 500,000 Adam iterations.} \label{AC_Results} \end{figure}

The Allen-Cahn equation is a widely recognized benchmark in Physics-Informed Machine Learning (PIML) due to its challenging characteristics~\cite{raissi2019physics,mcclenny2023self,anagnostopoulos2024residual,wang2024piratenets}. This complexity arises from the equation's tendency to generate solutions with sharp transitions in both spatial and temporal dimensions, which makes accurate approximation and prediction particularly difficult. The 1D Allen-Cahn partial differential equation (PDE) is defined as:

\begin{equation} \frac{\partial u}{\partial t} = k\frac{\partial^2 u}{\partial x^2} - 5u(u^{2}-1), \end{equation}

\noindent where $k = 10^{-4}$. The problem is further defined by the following initial and periodic boundary conditions:

\begin{equation} u(0, x) = x^2 \cos(\pi x), \quad \forall x \in [-1, 1], \end{equation}

\begin{equation} u(t, x + 1) = u(t, x - 1), \quad \forall t \geq 0 \quad \text{and} \quad x \in [-1, 1]. \end{equation}

The Allen-Cahn equation’s complexity has motivated the development of numerous techniques and methods specifically tailored for MLP-based PIML to address its inherent difficulties. Consequently, directly comparing cKANs and KKANs with MLPs presents a challenge: excluding enhancements designed for MLPs may unfairly disadvantage them while including these techniques could potentially undermine the unique capabilities of cKANs and KKANs. Therefore, we divide our analysis into three parts.

\begin{figure}[t]
    \centering
    \includegraphics[width=1\linewidth]{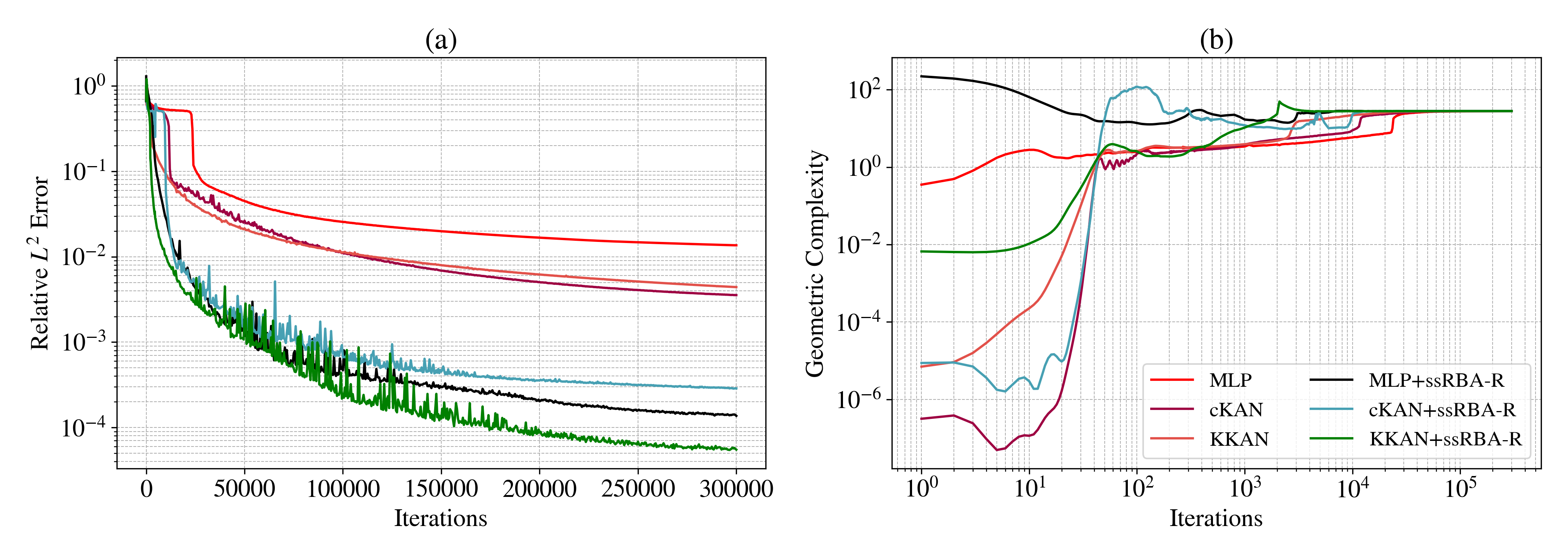}
\caption{Results for solving the Allen-Cahn Equation. (a) Relative $L^2$ error convergence for the analyzed models. Vanilla KKAN and cKAN models converge significantly faster than MLP, which begins converging after $20,000$ Adam iterations. The inclusion of enhancements such as ssRBA and Fourier Feature embeddings~\cite{wang2021eigenvector} accelerates convergence for all models, with KKAN+ssRBA achieving a relative $L^2$ error of $5.5 \times 10^{-5}$. While larger networks could achieve better performance, the proposed approach is approximately four times faster than alternative methods. (b) Geometric complexity evolution during training. Initially, cKANs and KKANs exhibit lower complexity compared to MLPs. The Fourier Feature embeddings increase the initial complexity of the vanilla models, but as observed in previous cases, all models eventually converge to a uniform or optimal geometric complexity.}
    \label{AC_RL2_plot}
\end{figure}

First, to ensure a fair comparison with cKANs, which tend to degrade in performance with a higher number of parameters, we evaluate models with a reduced parameter count. Specifically, we compare the baseline versions of cKANs, MLPs, and KKANs without incorporating any additional enhancements. All models are trained using the loss function described in \eqref{PIML_loss}, with $\lambda_{\alpha,i} = 1$ and $m_B = 100$, as proposed in~\cite{mcclenny2023self}, using a batch size of 10,000 and train it for $300000$ ADAM~\cite{kingma2014adam} iterations. For KKANs, the inner block starts with a polynomial embedding layer, which enforces periodicity exactly by using odd polynomial degrees. To maintain consistency and fairness, this embedding layer is also included in the MLP and cKAN models.

The results, presented in Table~\ref{AC_PIML}(a), show that both KKANs and cKANs outperform MLPs in this configuration. MLP achieves a relative $L^2$ error of $1.36 \times 10^{-2}$, which is significantly worse compared to cKANs ($3.55 \times 10^{-3}$) and KKANs ($4.41 \times 10^{-3}$). Furthermore, as shown in Figure~\ref{AC_RL2_plot}(a), KKANs and cKANs converge faster than MLPs, demonstrating their effectiveness in capturing the sharp transitions characteristic of the Allen-Cahn equation.

Figure~\ref{AC_RL2_plot}(b) shows the evolution of geometric complexity during training. At the beginning of training, MLPs exhibit the highest geometric complexity, while cKANs show the lowest, reflecting their initial simplicity. By the end of training, all models converge to similar geometric complexity values, suggesting that their representational capacities become comparable after sufficient training iterations. Despite this, KKANs and cKANs maintain superior accuracy, as indicated by their significantly lower $L^2$ errors compared to MLPs.

\begin{table}[t]
    \centering
    \begin{tabular}{clccc}
        \toprule
        &\textbf{Model} & \textbf{N. Params} & \textbf{Time (ms/it)} & \textbf{Rel. $L^2$ Error} \\
        \midrule
        a&MLP       & 21318 & 6.75 & $1.36 \times 10^{-2}$ \\
        &cKAN      & 20361   & 6.37  & $3.55 \times 10^{-3}$ \\
        &KKAN      & 19572 & 6.75 & $4.41 \times 10^{-3}$ \\
        \midrule
        b&MLP+ssRBA-R     & 25281 & 6.29 & $1.37 \times 10^{-4}$ \\
        &cKAN+ssRBA-R              & 22848   & 5.15  & $2.87 \times 10^{-4}$ \\
        &KKAN+ssRBA-R    & 22160 & 6.80 & $5.50 \times 10^{-5}$ \\
        \midrule
        c&MLP+ssRBA            & 91521 & 26.85 & $3.52\times 10^{-5}$ \\
        &cKAN+ssRBA            & 107136 & 31.71 & $2.91\times 10^{-4}$ \\

        &\textbf{KKAN+ssRBA}            & 118491 & 21.36 & $\mathbf{3.07 \times 10^{-5}}$ \\
        \bottomrule
    \end{tabular}
    \caption{Comparison of Models for Physics-Informed Machine Learning for the Allen-Cahn Equation Based on Relative $L_2$ Errors.}
    \label{AC_PIML}
\end{table}

For the second part, we enhance the models by incorporating Fourier feature embeddings to encode periodicity across all architectures. For MLPs, we combine weight normalization~\cite{salimans2016weight} with the modified MLP (mMLP)~\cite{wang2021understanding}, resulting in an enhanced architecture referred to as weight-normalized mMLP (WNmMLP) (see ~\ref{WN_MMLP}). Leveraging the inherent flexibility of KKANs, WNmMLP is seamlessly integrated into their inner block. In contrast, cKANs are not compatible with enhancements like mMLP or weight normalization, as current methods designed for MLPs do not directly translate to cKANs. Consequently, cKANs are implemented without these modifications, which partially explains their slightly faster training times.

\begin{table}[t]
    \centering
    \begin{tabular}{lccc}
        \toprule
        \textbf{Model} & \textbf{Enhancements} & \textbf{Boundary Conditions} & \textbf{Rel. $L^2$ Error} \\
        \midrule
        MLP & SA~\cite{mcclenny2023self}           & Periodic  & $(1.51 \pm 2.76) \times 10^{-4}$ \\
        MLP & RBA~\cite{anagnostopoulos2024residual}          & Periodic  & $(5.80 \pm 0.74) \times 10^{-5}$ \\
        \textbf{MLP} & \textbf{BRDR+}~\cite{chen2024self} & Periodic  & $\mathbf{(1.45 \pm 0.46) \times 10^{-5}}$ \\
        MLP & PirateNets~\cite{wang2024piratenets}   & Periodic  & $2.24 \times 10^{-5}$ \\
        \midrule
        KAN & AcNet~\cite{guilhoto2024deeplearningalternativeskolmogorov}         & Dirichlet & $6.80 \times 10^{-5}$ \\
        KAN~\cite{liu2024kan} & ~\cite{guilhoto2024deeplearningalternativeskolmogorov} implementation & Dirichlet & $5.30 \times 10^{-4}$ \\
        KAN~\cite{liu2024kan} & ~\cite{shukla2024comprehensive} implementation & Dirichlet & $5.65 \times 10^{-2}$ \\

        \midrule
        KKAN & ssRBA (Ours) & Periodic  & $(2.56 \pm 0.17) \times 10^{-5}$ \\
        \bottomrule
    \end{tabular}
\caption{Comparison of state-of-the-art methods for solving the Allen-Cahn equation using various representation models and enhancements. The table includes relative $L^2$ errors for models trained with periodic and Dirichlet boundary conditions. MLP-based methods demonstrate strong performance, with BRDR+ achieving the best result among MLPs. The proposed KKAN+ssRBA formulation outperforms all KAN-based formulations and demonstrates highly competitive accuracy compared to state-of-the-art MLPs.}
    \label{tab:state_of_art_periodic}
\end{table}

For all models, we employ ssRBA-R with a batch size of 10,000. This combination of resampling and reduced parameter counts provides a speed-up of more than four times compared to standard setups. KKAN+ssRBA-R outperforms all other models, achieving a relative $L^2$ error of $5.5 \times 10^{-5}$, which is comparable to state-of-the-art results (see Figure~\ref{AC_RL2_plot}(a)). As shown in Figures~\ref{fig:RBAR-KART} and~\ref{fig:RBAR-MLP}, the proposed formulation effectively resamples high-error regions, enabling a double-attention mechanism via multipliers and sampling. Consequently, as depicted in Figure~\ref{AC_RL2_plot}(a), ssRBA-R accelerates convergence for all models, particularly for MLPs. Without ssRBA-R, MLPs require approximately 20,000 iterations to converge, but with this method, convergence is achieved in around 3,000 iterations.

Figure~\ref{AC_RL2_plot}(b) shows the geometric complexity evolution during training. While Fourier feature embeddings improve model performance, they also increase the geometric complexity for all architectures. cKANs and KKANs begin with lower geometric complexity than MLPs, demonstrating their simpler initial representations. However, all models converge to the same final complexity value, highlighting a shared representational capacity at the end of training.

For the final part of the analysis, we employ an architecture tailored to benefit MLPs. Specifically, we use larger networks and full-batch training, following the original RBA implementation~\cite{anagnostopoulos2024residual}. For MLPs, we adopt the WNmMLP architecture, while for KKANs, we introduce a weight-normalized adaptive ResNet (\ref{AdResNet}), a KKAN-specific design inspired by~\cite{wang2024piratenets}. To ensure a fair comparison, we scale the number of parameters in cKANs to approximately match those of the other architectures.

As shown in Table~\ref{AC_PIML}(c), with larger networks and full-batch training, even the base cKAN architecture becomes significantly slower than both MLPs and KKANs. Despite this, cKANs still demonstrate competitive accuracy. MLP+ssRBA achieves a relative $L^2$ error of $3.52 \times 10^{-5}$, highlighting the effectiveness of ssRBA compared to the original RBA~\cite{anagnostopoulos2024residual}, where the authors' best-performing model achieved $4.57 \times 10^{-5}$. Notably, KKAN+ssRBA outperforms all other models, achieving a relative $L^2$ error of $3.07 \times 10^{-5}$ while also being faster than MLP+ssRBA, demonstrating its efficiency and superior accuracy. The convergence history for these three models is shown in Figure~\ref{fig:AC-other}(a).

\begin{figure}[H]
    \centering
    \includegraphics[width=1\linewidth]{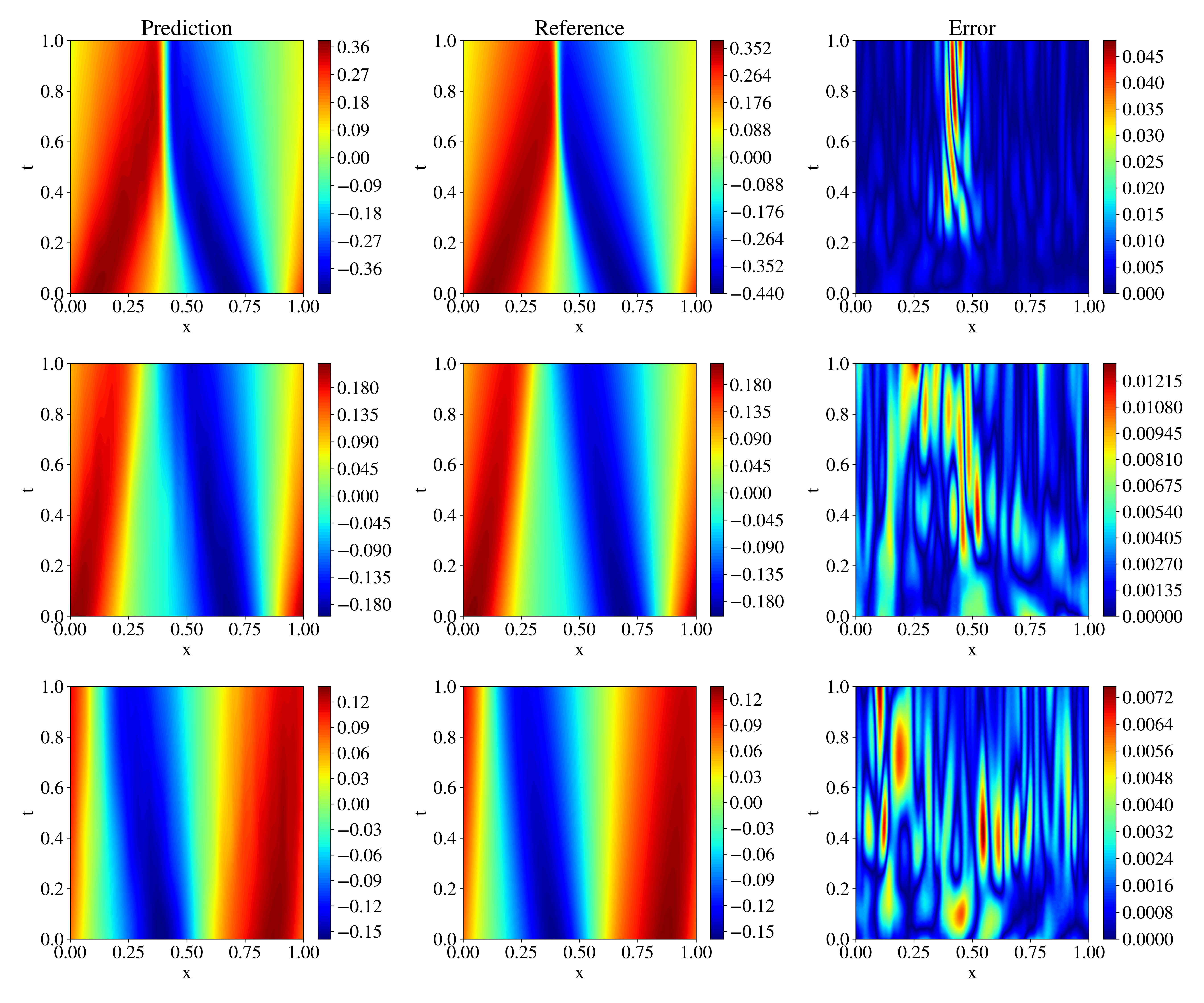}
\caption{QR-DeepOKKAN predictions for three different initial conditions from the testing dataset. The corresponding relative $L^2$ errors are: (top row) $2.30 \times 10^{-2}$, (middle row) $2.03 \times 10^{-2}$, and (bottom row) $1.52 \times 10^{-2}$.}
    \label{Burgers_results}
\end{figure}

Additionally, since ssRBA allows our model to train for longer periods by maintaining a higher signal-to-noise ratio (SNR), we further trained the model for 500,000 Adam iterations, resulting in a best-performing model with a relative $L^2$ error of $2.28 \times 10^{-5}$. The predictions, references, and pointwise errors are shown in Figure~\ref{AC_Results}. Furthermore, we analyzed the model's performance across five different initializations (see Figure~\ref{fig:AC-other}(b)). Additionally, the global weight, relative $L^2$ error, and SNR evolution for KKANs are presented in Figure~\ref{fig:ssRBA-KKAN}. 

Table~\ref{tab:state_of_art_periodic} shows that the proposed model achieves an average relative error of $2.56 \times 10^{-5}$ with a standard deviation of $0.17 \times 10^{-5}$. This table compares our results with current state-of-the-art methods, demonstrating that the proposed KKAN formulation outperforms all other KAN variants and achieves performance highly comparable to fully optimized MLPs.

\subsection{Neural Operators}
\smallskip

Finally, we evaluate the performance of MLPs, cKANs, and KKANs within operator learning frameworks. To this end, we extend the DeepONet framework~\cite{lu2019deeponet} to KKANs, resulting in the models DeepOKKAN. These models are trained using the loss function described in Section~\ref{NO_loss}. For KKAN-based models, we use radial basis functions (RBFs) as outer blocks. It is worth noting that the DeepONet framework has previously been extended to cKANs and KANs in~\cite{shukla2024comprehensive} and~\cite{abueidda2024deepokan}, respectively. 

Additionally, we enhance our models by incorporating the QR-DeepONet framework introduced in~\cite{lee2024training}. QR-DeepONet improves the standard DeepONet architecture by reparameterizing the trunk network and leveraging a QR decomposition to enhance training stability and accuracy. A detailed description of DeepONets, QR-DeepONets, and their corresponding objective functions (i.e., \eqref{train_trunk} and \eqref{train_branch}) is provided in Appendix~\ref{DeepONet_appendix}.

All models were trained on a dataset of 3,500 functions and tested on 1,500 functions. The QR-based models utilized a two-stage training process, with Stage 1 running for 200,000 iterations and Stage 2 for 400,000 iterations. To ensure a fair comparison, the model architectures were chosen so that the number of parameters was approximately matched across all models. Further details regarding the specific implementation and architectural choices are provided in Section~\ref{Imp_Details}.

\begin{figure}[ht]
    \centering
    \includegraphics[width=0.75\linewidth]{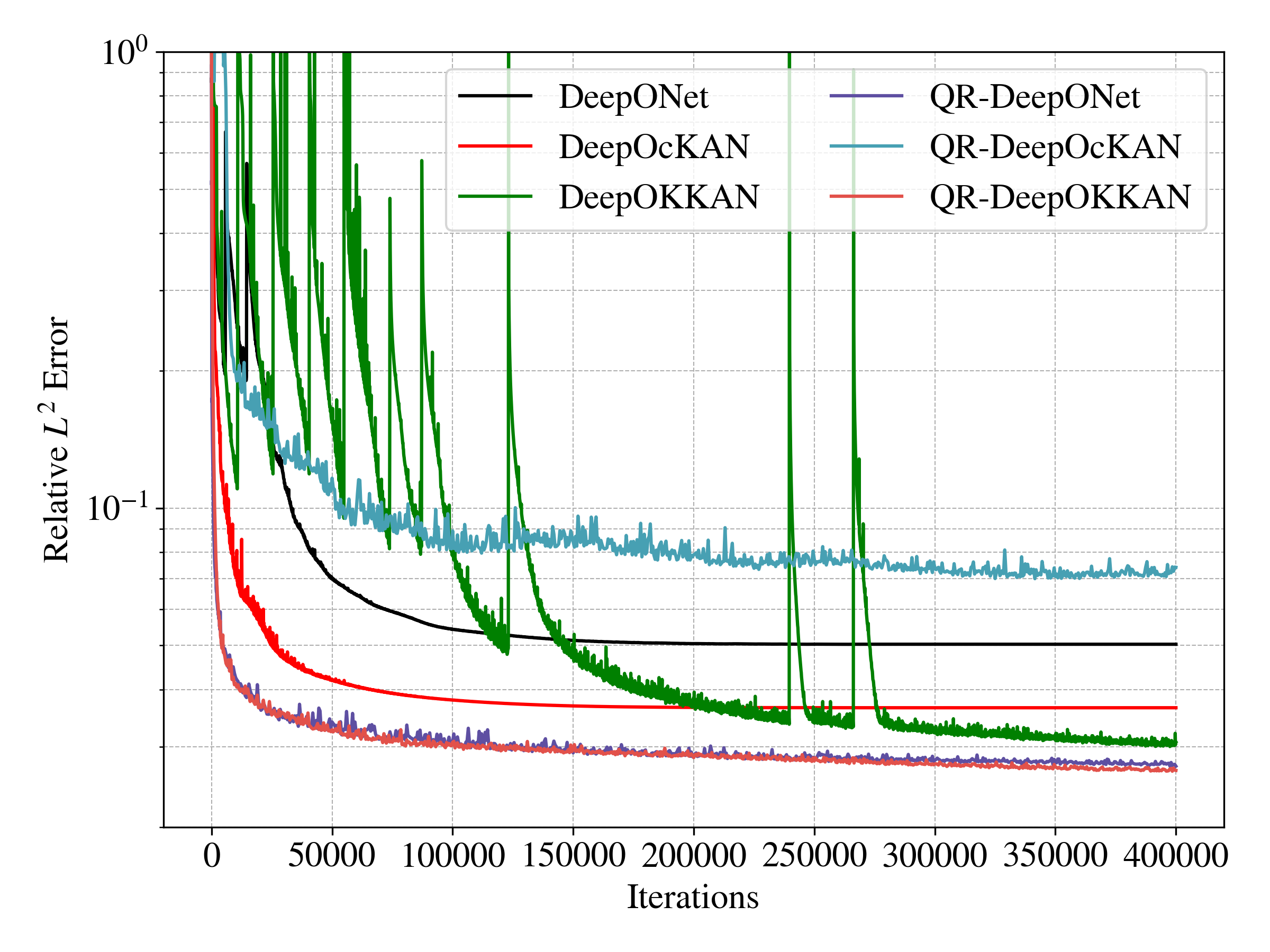}
\caption{Relative $L^2$ error convergence for operator learning the testing dataset. KKANs outperform cKANs and MLPs, while cKANs demonstrate faster initial convergence compared to the other models. The QR formulation enhances the performance of MLPs and KKANs but negatively impacts cKANs, indicating their sensitivity to optimization techniques. This highlights the robustness and adaptability of KKANs compared to cKANs.}
    \label{Errors_deepONet}
\end{figure}

\subsubsection{Burgers Equation}

 For this case, we consider the Burgers equation, which is known for producing sharp transitions at low viscosity values, $\nu$. Specifically, we focus on the 1D Burgers equation benchmark investigated in~\cite{li2020fourier,wang2021learning}:
\begin{equation} u_t + uu_x = \nu u_{xx}, \label{eq:Burgers} \end{equation}
\noindent where $u$ represents the velocity field, and $\nu$ denotes the dynamic viscosity. The equation is subject to periodic boundary conditions:
\begin{align} u(t, 0) &= u(t, 1), \\ u_x(t, 0) &= u_x(t, 1), \label{Burgers_BC}
\end{align}
\noindent defined over the domain $\Omega = (0,1) \times (0,1)$. The goal is to learn the operator map from the initial condition $u(0,x)$ to the solution $u(t,x)$ for all $t \in (0,1)$. The initial condition, $u(0,x)$, is sampled from a Gaussian random field (GRF) using the code provided in~\cite{wang2021learning}. Previous studies~\cite{li2020fourier} analyzed these results only for the final state $u(1,x)$ and used a lower viscosity $\nu=0.01$. In this study, we challenge our models' representation capabilities and predict the full solution history with a much lower viscosity $\nu=1/(100\pi)$.

Figure~\ref{Burgers_results} shows the predictions of QR-DeepOKKAN for three representative initial conditions sampled from the testing dataset. Figure~\ref{Errors_deepONet} provides a comparison of relative $L^2$ error convergence on the testing dataset. KKAN-based models consistently outperform cKANs and MLPs, showcasing their superior representational capability and flexibility. While cKANs converge faster initially, their performance plateaus. The QR formulation significantly enhances the performance of MLPs and KKANs, reducing errors and improving convergence speed. However, the QR reparameterization adversely affects cKANs, highlighting their architectural rigidity compared to the adaptability and robustness of KKANs.

\begin{table}[H]
    \centering
    \begin{tabular}{lcccc}
        \toprule
        \textbf{Model} & \textbf{N. Params} & \textbf{ S1 (ms/it)} & \textbf{S2 (ms/it)} & \textbf{Rel. $L_2$ Error} \\
        \midrule
        DeepONet            & 131700     & 8.14&  & $5.03 \times 10^{-2}$ \\
        QR-DeepONet         & 481700     & 3.42&5.84    & $2.72 \times 10^{-2}$ \\
        \midrule
        DeepOcKAN           & 107328        & 7.76&     & $3.65 \times 10^{-2}$ \\
        QR-DeepOcKAN        & 461212        & 4.69&5.43    & $7.41 \times 10^{-2}$ \\
         \midrule
        DeepOKKAN           & 155920     & 12.69&   & $3.07 \times 10^{-2}$ \\
        \textbf{QR-DeepOKKAN} & 496358    & 3.37&5.14      & $\mathbf{2.66 \times 10^{-2}}$ \\
        \bottomrule
    \end{tabular}
\caption{Comparison of operator learning variants for the Burgers equation with $\nu = 1/(100\pi)$. The table reports the number of parameters, training time per iteration for Stage 1 (S1) and Stage 2 (S2) of the QR variants, and the relative $L^2$ errors. The QR-based models utilize a two-stage training process, with S2 involving the QR reparameterization. For reference, under the physics-informed framework,~\cite{wang2021learning} reported a relative $L^2$ error of $3.3 \times 10^{-2}$ for $\nu = 1/100$, while~\cite{chen2024self} achieved $3.4 \times 10^{-2}$ for $\nu = 10^{-3}$. In a comparable data-driven setup with $\nu = 1/(100\pi)$, ~\cite{shukla2024comprehensive} reported relative $L^2$ errors of $5.83 \times 10^{-2}$ for cDeepOKAN and $3.02 \times 10^{-2}$ for DeepONet when predicting only the final time step $u(x,1)$. In contrast, our models predict the full solution $u(x,t)$ for $t \in [0,1]$, with the QR-DeepOKKAN achieving the best accuracy, a relative $L^2$ error of $2.66 \times 10^{-2}$.}
    \label{tab:deep_models_comparison}
\end{table}

Table~\ref{tab:deep_models_comparison} provides a detailed comparison of operator learning variants for the Burgers equation with $\nu = 1/(100\pi)$. The table reports the number of parameters, training time per iteration for Stage 1 (S1) and Stage 2 (S2) of the QR-based models, and the relative $L^2$ errors. QR-DeepONet achieves a significant reduction in error ($2.72 \times 10^{-2}$) compared to the standard DeepONet ($5.03 \times 10^{-2}$), demonstrating the effectiveness of QR reparameterization for enhancing stability and accuracy. However, for cKANs, the QR formulation results in performance degradation, as seen with the QR-DeepOcKAN error increasing to $7.41 \times 10^{-2}$ compared to $3.65 \times 10^{-2}$ for the standard DeepOcKAN.

In contrast, QR-DeepOKKAN achieves the best overall performance with a relative $L^2$ error of $2.66 \times 10^{-2}$ while maintaining competitive training times. This result highlights the ability of KKANs to effectively integrate with advanced frameworks like QR-DeepONet, benefiting from improved stability and accuracy without sacrificing computational efficiency. Unlike earlier studies~\cite{shukla2024comprehensive} that predict only the final time step $u(x,1)$, our models successfully predict the entire solution $u(x,t)$ for $t \in [0,1]$, further underscoring the robustness and generalization capabilities of QR-DeepOKKAN.

\newpage
\section{Learning Dynamics via Information Bottleneck Theory}
\label{Sec_learning}
\begin{figure}[H]
    \centering
    \includegraphics[width=1\linewidth]{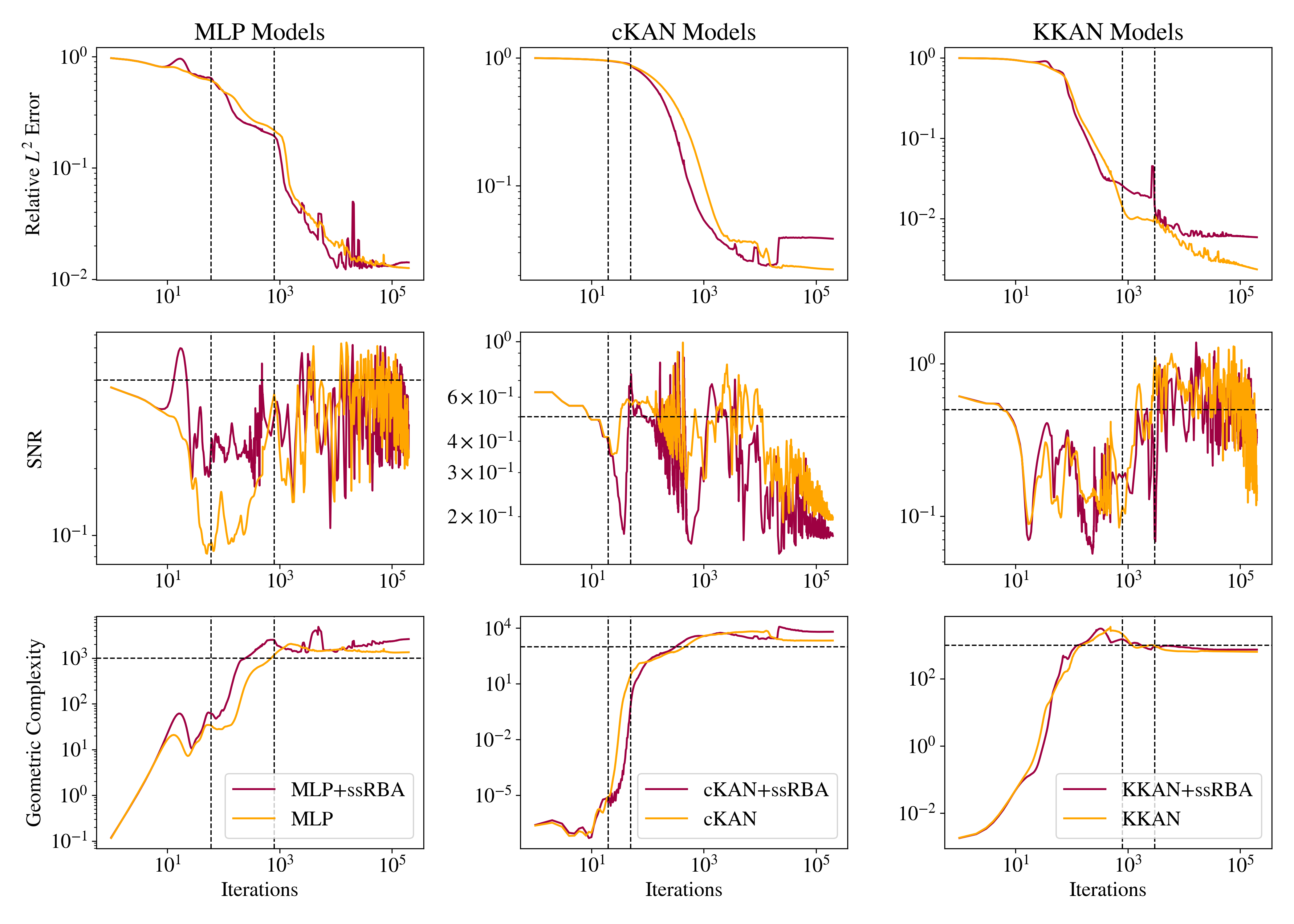}
\caption{Relative $L^2$ error (first row), SNR (second row), and geometric complexity (third row) convergence for discontinuous function approximation using MLP (first column), cKAN (second column), and KKAN models (third column). The three stages of learning are observed across all representation models and are marked, for the worst performing case, by vertical dashed lines.  During the fitting stage, the SNR decreases without significant improvement in the generalization error (i.e., relative $L^2$ error) and with an increase in geometric complexity. In the transition stage, the model explores the optimal direction, resulting in further increases in geometric complexity. In the diffusion stage, the model achieves the optimal complexity and finds the optimal direction, leading to an increase in SNR and significant convergence. However, it can also be observed that, over time, the SNR decreases again. For extreme cases such as cKAN, this decrease in SNR leads to higher geometric complexity, causing the model to overfit and harming the generalization performance.}
    \label{disc_SNR}
\end{figure}

The learning dynamics of a model can be characterized by analyzing the signal-to-noise ratio (SNR) of backpropagated gradients (i.e., \eqref{SNR_Eq}). The SNR provides insight into deterministic and stochastic regimes during training, where high SNR corresponds to deterministic updates, and low SNR indicates stochastic exploration~\cite{anagnostopoulos2024learning}. By studying the variation in SNR, we can identify distinct stages of learning: fitting, transition, and diffusion. These stages have been observed in PIML tasks for MLP and cKAN models, and in this study, we extend the framework to KKANs for function approximation and operator learning tasks.

\begin{figure}[H]
    \centering
    \includegraphics[width=1\linewidth]{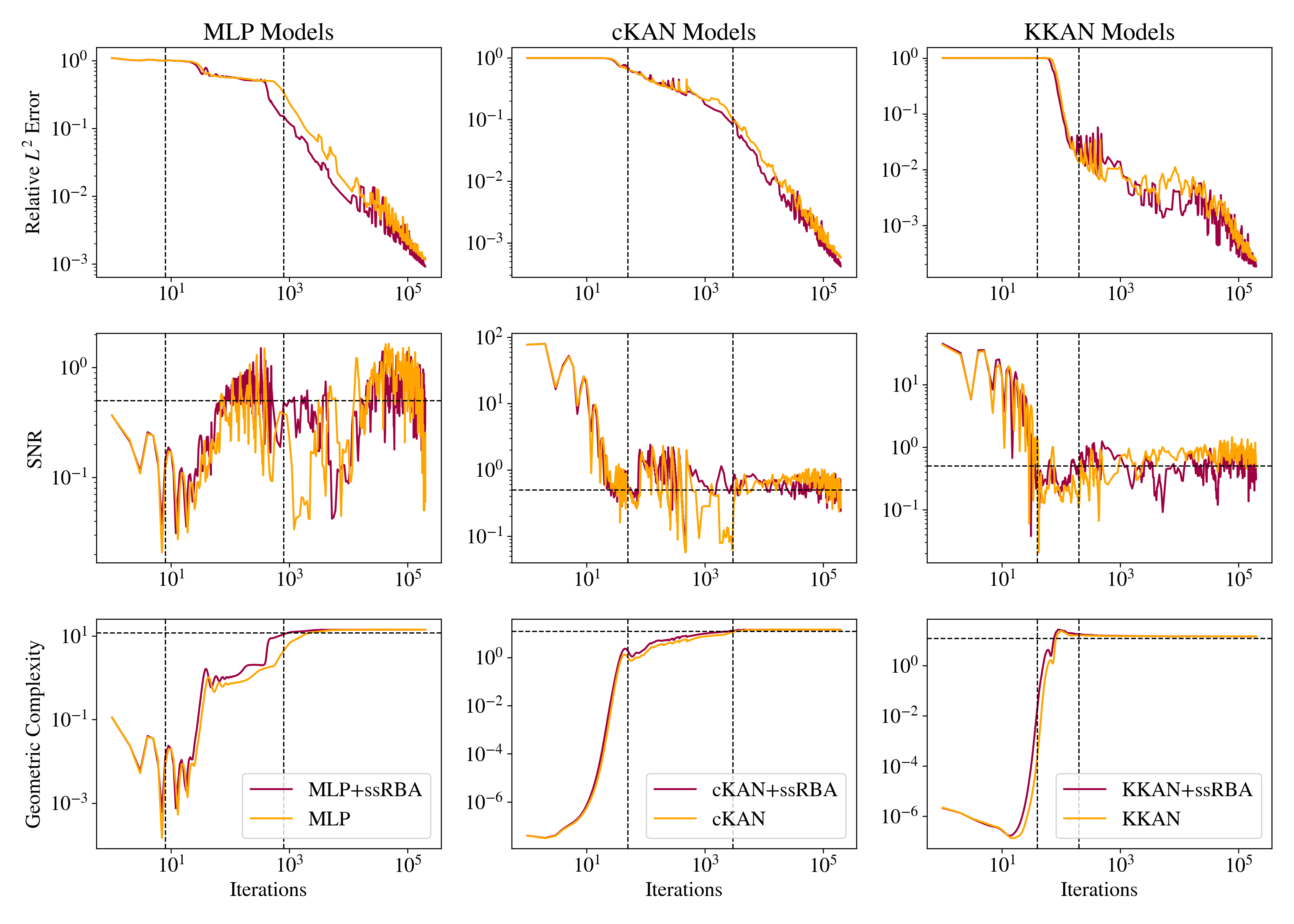}
\caption{Relative $L^2$ error (first row), SNR (second row), and geometric complexity (third row) convergence for smooth function approximation with MLP (first column), cKAN (second column), and KKAN models (third column). The three stages of learning are evident across all models and are indicated, for the worst-performing case, by vertical dashed lines. During the fitting stage, the SNR decreases, accompanied by a slight improvement in the generalization error (i.e., relative $L^2$ error) and increased geometric complexity. Notably, for MLPs, the geometric complexity is initially high and decreases before further increasing again. In the transition stage, the model explores the optimal direction for learning, which results in an increase in geometric complexity. Finally, in the diffusion stage, the model achieves the optimal complexity and direction, causing the SNR to rise and driving significant convergence. For smooth functions, maintaining a high SNR is easier, which supports continuous learning and improved performance.}
    \label{fig:SNR_HO}
\end{figure}

\begin{figure}[H]
    \centering
    \includegraphics[width=1\linewidth]{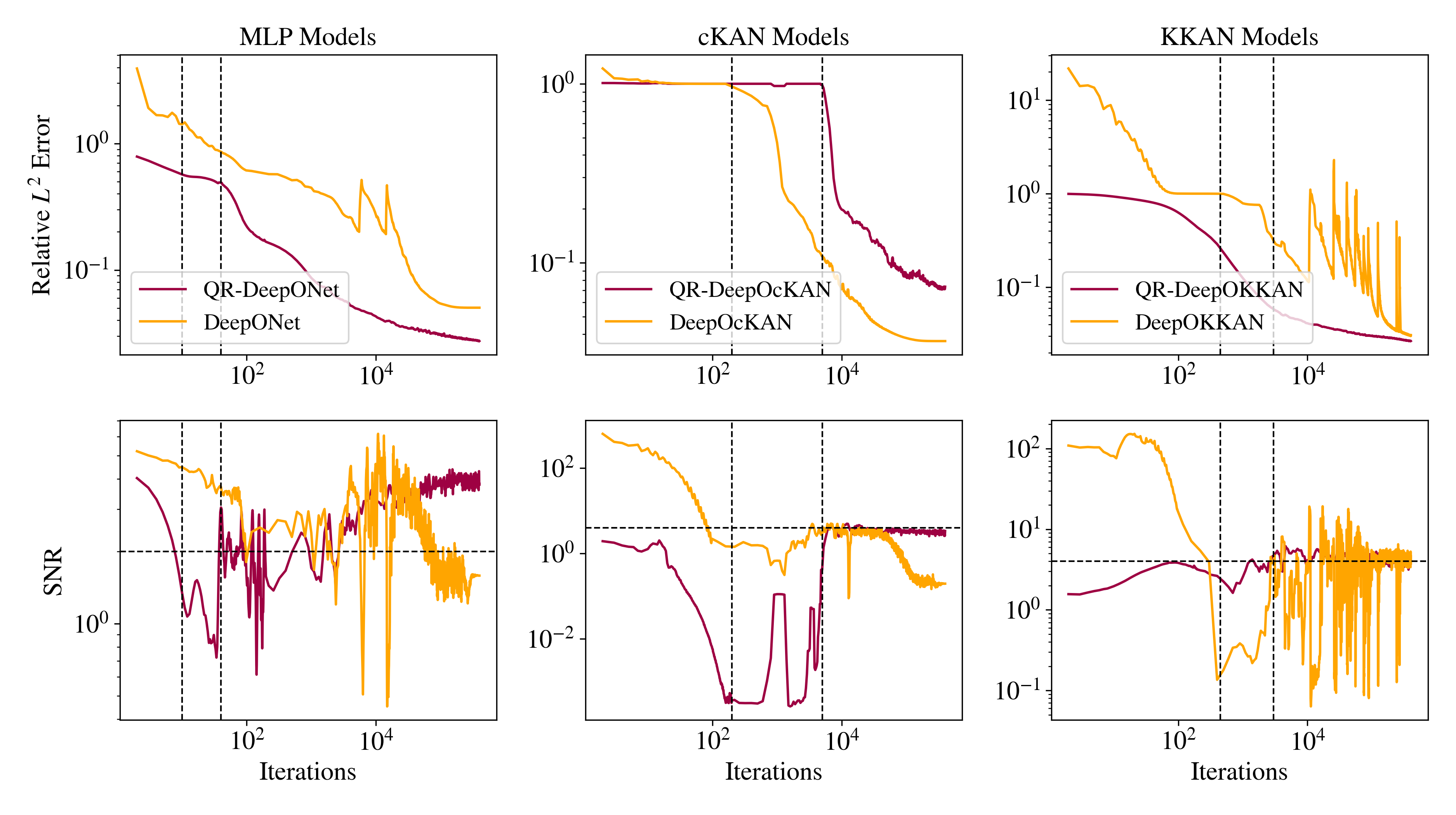}
\caption{Relative $L^2$ error (first row) and SNR (second row) for operator learning tasks using MLP (first column), cKAN (second column), and KKAN models (third column). Due to the high dimensionality of the branch net inputs (100), computing the geometric complexity is computationally challenging. Therefore, only the relative error and SNR results are presented. The three stages of learning are evident across all models, with vertical dashed lines indicating the stages for the worst-performing case. During the fitting stage, the SNR decreases, accompanied by minor improvements in the generalization error (i.e., relative $L^2$ error). In the transition stage, the model explores the optimal direction, with no significant improvement in generalization error. Finally, in the diffusion stage, the model identifies the optimal direction, leading to an increase in SNR and driving significant convergence.}
    \label{SNR_DN}
\end{figure}

In this study, we extend the Information Bottleneck (IB) framework to KKANs as well as function approximation and operator learning tasks. Notably, we experimentally observed that the three stages of learning are present across all tasks, including function approximation (Figures~\ref{disc_SNR} and \ref{fig:SNR_HO}), PIML (Figure~\ref{AC_SNR}), and operator learning (Figure~\ref{SNR_DN}), for all representation models: MLP (first column), cKAN (second column), and KKAN (third column). The SNR is closely linked to the generalization error shown in the first row of all our figures. Additionally, we observed an intriguing connection between the SNR and the geometric complexity, as depicted in the third row of all the figures.

In particular, the three stages of training are described as follows: 

\begin{figure}[H]
    \centering
    \includegraphics[width=1\linewidth]{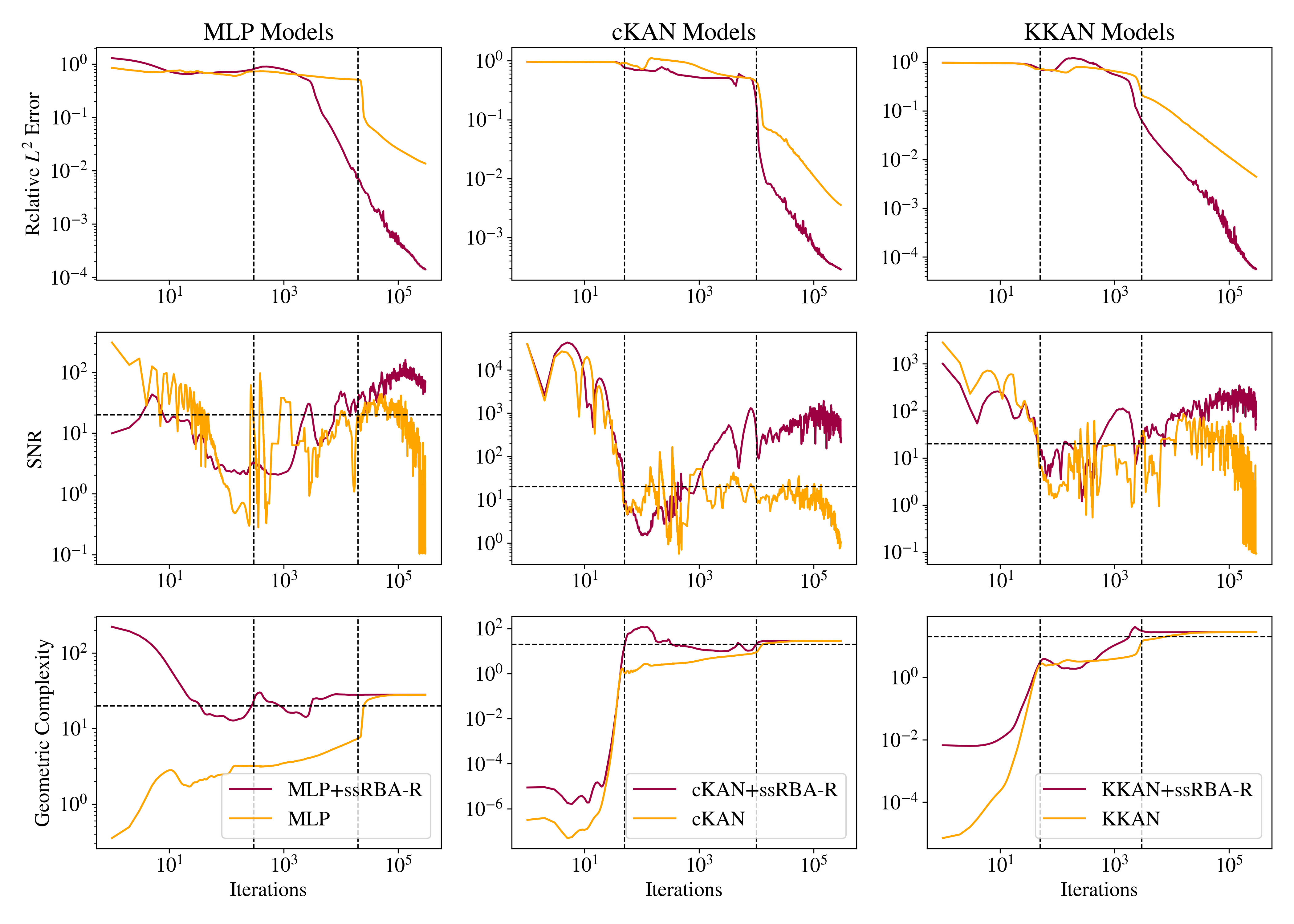}
\caption{Relative $L^2$ error (first row), SNR (second row), and geometric complexity (third row) convergence for solving the Allen-Cahn Equation using MLP (first column), cKAN (second column), and KKAN models (third column). The three stages of learning are evident across all models and are marked, for the worst-performing case, by vertical dashed lines. During the fitting stage, the SNR decreases, accompanied by minor improvements in the generalization error (i.e., relative $L^2$ error) and an increase in geometric complexity. As discussed in the results section, the mMLP increases the geometric complexity of the models. Notably, MLP+ssRBA-R exhibits a significantly high geometric complexity, where the model first simplifies the representation, characterized by fluctuations in the SNR. In the transition stage, the model explores the optimal direction for learning, resulting in a further increase in geometric complexity. However, during this stage, the generalization error does not improve. Finally, in the diffusion stage, the model achieves the optimal complexity and direction, leading the SNR to converge to an optimal value and driving significant improvements in generalization error. It is noteworthy that the proposed ssRBA-R successfully maintains a consistently high SNR, resulting in optimal performance across all representation models.}
    \label{AC_SNR}
\end{figure}

\paragraph{Fitting} The initial phase of training, where gradients are large and agreement between subdomains is high, resulting in a high SNR. During this deterministic phase, the model focuses on reducing the training error across subdomains. However, as the model learns the ``general" trend, the disagreement between subdomains increases, which leads to a low SNR. Therefore, this stage can be identified by a transition from high to low SNR. 

The fitting stage can be clearly identified in all our examples, as shown in the second row of Figures~\ref{disc_SNR},~\ref{fig:SNR_HO},~\ref{AC_SNR}, and~\ref{SNR_DN}. During this stage, the training errors are reduced; however, there is minimal improvement in the relative error on the testing dataset (see the first row of Figures~\ref{disc_SNR},~\ref{fig:SNR_HO},~\ref{AC_SNR}, and~\ref{SNR_DN}). Additionally, during this stage, as the model fits new information, it becomes more complex, leading to an increase in geometric complexity. Notice that if the geometric complexity is high at the beginning of training (as seen in MLP in Figure~\ref{fig:SNR_HO} and ~\ref{AC_SNR}), the model first needs to simplify these representations; this may justify why better initialization schemes are characterized by lower initial geometric complexities~\cite{dherin2022neural}.
  
\paragraph{Transition:} After the model fits the data, it enters an exploration stage where it attempts to minimize the error across all subdomains. During this exploratory stage, there is disagreement on the optimal direction for weight updates, resulting in a low SNR. This phase is also characterized by a minor or null decrease in the generalization error (see Figures~\ref{disc_SNR},~\ref{fig:SNR_HO},~\ref{AC_SNR}, and~\ref{SNR_DN} (first row)). Additionally, as the model attempts to minimize the error along all subdomains, the geometric complexity increases (see Figures~\ref{disc_SNR},~\ref{fig:SNR_HO},~\ref{AC_SNR}, and~\ref{SNR_DN} (third row)).
  
\paragraph{Diffusion:} After the exploration phase, the model becomes sufficiently complex (i.e., after the geometric complexity increases) to reach an agreement on the optimal update direction, leading to a sudden increase in the SNR. Once this optimal direction (i.e., high SNR) is achieved, the generalization error improves significantly, and the geometric complexity decreases as the model becomes more efficient and starts simplifying internal representations. However, as the loss decreases further, the gradients become smaller (i.e., lower signal), which leads to a subsequent decrease in the SNR—eventually, the SNR drops, learning stops, and the generalization error plateaus. This pattern is clearly observed in our operator learning tasks (see Figure~\ref{SNR_DN}). For complex problems such as discontinuous function approximation (see Figure~\ref{disc_SNR}), this decrease in SNR induces an increase in the geometric complexity, which eventually causes the model to overfit. On the other hand, for smooth function approximation tasks, the SNR remains higher, enabling a continuous decrease in the generalization error without overfitting (see Figure~\ref{fig:SNR_HO}). Notice that the ss-RBA introduced for PIML tasks (Section~\ref{ssRBA_section}) successfully maintains a high SNR during the late diffusion stage, enabling continuous and improved learning (see Figure~\ref{AC_SNR}). 

Interestingly, during the diffusion stage, all models converge to the same geometric complexity. This behavior is expected because the geometric complexity \eqref{geometric_complexity_eq} depends solely on the Jacobian of the function, \( \nabla u_x \). Since \( u \) is unique, its derivatives with respect to the inputs are also identical, leading to the same geometric complexity across models. Notice that for discontinuous function approximation (see Figure~\ref{disc_SNR}), the geometric complexity is higher due to the undefined derivatives at the discontinuities. However, this metric proves useful for identifying unnecessarily complex functions, particularly for spatial data or operator learning tasks, such as function approximation. Similarly, in PIML, where the function is learned through its residuals, higher geometric complexity corresponds to larger \( \nabla u_x \) values when evaluated at the testing points, potentially indicating overfitting~\cite{dherin2022neural}.

\section{Summary}
\label{Sec_summary}

The Kolmogorov-Arnold representation theorem (KART) has historically faced skepticism regarding its practical utility, with criticisms citing its computational complexity and infeasibility for direct implementation. Inspired by K{$\dot {\rm u}$}rkov{\'a}'s reinterpretation of KART, we proposed the K{$\dot {\rm u}$}rkov{'a}-Kolmogorov-Arnold Networks (KKANs), a novel two-block architecture that adheres closely to the original theorem. KKANs combine robust MLP-based inner blocks with interpretable basis functions as outer blocks, preserving both computational efficiency and theoretical rigor. We proved that KKANs are universal approximators regardless of the choice of basis functions and for a general class of functions. Also, we extended KKAN's applicability to Physics-Informed Machine Learning (PIML) and operator learning tasks. We evaluated our models' performance using their generalization error (i.e., relative error on testing data), geometric complexity, and learning dynamics via the Information Bottleneck (IB) theory.

To improve  PIML models' performance, we developed a new optimization method for PIML called self-scaled residual-based attention (ssRBA). This method induces uniform convergence and enhances learning dynamics, enabling prolonged learning. Additionally, we proposed a global weighting scheme that scales loss weights based on their gradients, ensuring consistent update directions that minimize all loss terms.

In function approximation tasks, we showed that KKANs consistently outperformed MLPs and cKANs across both smooth and discontinuous function benchmarks. KKANs exhibited faster convergence and superior generalization, maintaining lower geometric complexity throughout training. In contrast, cKANs showed signs of overfitting, with higher final geometric complexity and deteriorating generalization performance. These results highlight KKANs’ ability to efficiently handle both smooth and discontinuous functions while preserving robustness and adaptability.

For PIML tasks, we demonstrated that KKANs offered significant computational advantages while achieving excellent accuracy, particularly when enhanced with ssRBA. KKANs converged faster than alternative methods while maintaining high generalization capabilities. By maintaining a high SNR during the diffusion stage, ssRBA enabled prolonged training and improved learning efficiency, positioning KKANs as strong competitors to state-of-the-art PIML models.

In operator learning tasks, we extended the DeepONet formulation to KKANs and  the QR-DeepONet formulation to cKANs and KKANs. The QR formulation stabilized training and enhanced KKAN and MLP performance, improving accuracy and robustness. However, we observed that the QR approach negatively impacted cKANs, revealing their sensitivity to optimization techniques. KKANs consistently demonstrated adaptability and superior performance, making them suitable for a wide range of operator learning applications.

We analyzed the learning dynamics of all the analyzed models using the Information Bottleneck (IB) theory, extending its application to function approximation and operator learning tasks. We demonstrated that the three stages of learning—fitting, transition, and diffusion—are universal across models and tasks. Additionally, we identified a strong relationship between geometric complexity and SNR, providing deeper insights into learning dynamics. In summary, we observed that during the fitting stage, the model captures general trends, transitioning from high to low SNR as geometric complexity converges to a structured representation. In the transition stage, the model explores subdomains to minimize errors, but the stochastic nature of this phase keeps SNR low while geometric complexity increases gradually.

In the diffusion stage, we observed that once the model becomes complex enough, it identifies the optimal update direction, resulting in a sharp increase in SNR. This stage is critical for reducing generalization error, as the geometric complexity converges to an optimal value unique to the learned solution. However, as gradients weaken and SNR decreases, overfitting can occur in complex tasks like discontinuous function approximation. To address this issue, we used ssRBA to dynamically scale local multipliers during the diffusion stage, maintaining a high SNR. This approach enabled prolonged learning and ensured robust performance for PIML tasks.

In conclusion, we propose KKANs as a transformative approach to implementing KART-based architectures in scientific machine learning. By closely adhering to the Kolmogorov-Arnold theorem and integrating advancements like ssRBA, KKANs bridge the gap between interpretability and computational efficiency. Our success across function approximation, PIML, and operator learning tasks establishes KKANs as a versatile and powerful tool, paving the way for future research and innovation in scientific machine learning.

\section*{Acknowledgements}
We acknowledge the support of the NIH grant R01AT012312, MURI/AFOSR FA9550-20-1-0358 project, the DOE-MMICS SEA-CROGS DE-SC0023191 award, and the ONR Vannevar Bush Faculty Fellowship (N00014-22-1-2795). The second author would like to thank the Crunch Group at Brown University for hosting his two-month visit commencing in August 2024. 

 \bibliographystyle{elsarticle-num} 
 \bibliography{cas-refs,Ref-WLL}

\newpage
\appendix
\section{Proof of Theorem \ref{UAT-KAN-Gen}}\label{UAT-proof}

 Thanks to the KART/variants, we can represent 
  any $f\in C(E^d)$  as in  \eqref{KART-01}: 
\begin{equation}\label{KA-form}
f(\bs x)=\sum_{q=0}^{m} g_q( \xi),\quad\;  \xi:=\xi_q(\bs x)= \sum_{p=1}^d \psi_{p,q}(x_p),
\end{equation}
where $\psi_{p,q}, g_q$ are  univariate  continuous functions. The continuity of 
 $\psi_{p,q}$ on the closed interval $I_\psi$ implies that  there exists $B>0$ such that
$$
\max_{p, q} \max_{x \in I_\psi}\left|\psi_{p,q}(x)\right| \leq \frac {B} d\quad \Rightarrow \quad  \xi \in [-B,B].
$$
Since all $g_q$ are uniformly continuous on $[-B,B]\subseteq I_g$, there exists $\delta>0$ such that 
\begin{equation}\label{gpqxi}
\max_q\, \max _{\substack{|\xi-\eta|<\delta \\\xi,\eta\in [-B,B]}}\left|g_q(\xi)-g_q(\eta)\right|<\frac{\varepsilon}{2(2 m+1)}.
\end{equation}
Using the assumption that ${\mathcal A}_{M_z}(I_z)$ is dense in 
$C(I_z),$ we know that
there exist $ \Psi_{p,q}\in {\mathcal A}_{M_\psi}(I_\psi)$ for each pair $(p,q)$ such that 
 \begin{equation}\label{proofA}
\max _{p,q} \max _{x\in I_\psi} \big|\psi_{p,q}(x)- \Psi_{p,q}(x)\big|<\frac{\delta} d,  
 \end{equation}
and there exists $G_q \in {\mathcal A}_{M_g}(I_g)$  such that 
\begin{equation}\label{proofB}
\max_{q}\, \max _{\eta\in [-B,B]}
\left|g_q(\eta)-G_q(\eta)\right|<\frac{\varepsilon} {2(2 m+1)}.
\end{equation}
Thus, we derive from \eqref{proofA} that 
$$
|\xi-\eta|:=\Big|\sum_{p=1}^d \psi_{p,q}(x_p)-\sum_{p=1}^d  \Psi_{p,q}(x_p)\Big| \le 
\sum_{p=1}^d |\psi_{p,q}(x_p)-\Psi_{p,q}(x_p)|<\delta.
$$
Then, we can assemble $F\in {\mathbb K}_M^{m,d}$ based on 
$\Psi_{p,q}, G_q$ as in \eqref{KappM}. Consequently, using \eqref{KA-form}, \eqref{gpqxi} and \eqref{proofB} leads to 
\begin{equation*}
 \begin{split}
|f(\bs x)-F(\bs x)|&=
\Bigg|\sum_{q=0}^{m} g_q\Big(\sum_{p=1}^d \psi_{p, q}(x_p)\Big)-\sum_{q=0}^{m}G_q\Big(\sum_{p=1}^d  \Psi_{p,q} (x_p)\Big)\Bigg|\\
&\le \sum_{q=0}^{m}\big| g_q(\xi)-G_q(\eta)\big|\le  \sum_{q=0}^{m}  
\{|g_q (\xi)-g_q(\eta)|+|g_q(\eta)-G_q(\eta)|\}<\varepsilon. 
\end{split}
\end{equation*}
This completes the proof.

\section{Representation Models}
\subsection{Multilayer Perceptron (MLP)}
\label{MLP_arch}

The output \( y \) of a Multilayer Perceptron (MLP) is computed through a nested formulation, where \( \sigma \) is the activation function, and \( W^{(l)} \) and \( b^{(l)} \) are the weights and biases of the \( l \)-th layer:
\[
y(\bm{x}) = \sigma\left(W^{(L)} \sigma\left(W^{(L-1)} \ldots \sigma\left(W^{(1)} \bm{x} + b^{(1)}\right) \ldots + b^{(L-1)}\right) + b^{(L)}\right).
\]
Here, \( \bm{x} = (x_1, x_2, \dots) \) is the input vector, and \( L \) is the number of layers. With sufficiently many neurons and a suitable activation function, MLPs can approximate any continuous function on compact subsets of \( \mathbb{R}^n \), as guaranteed by the Universal Approximation Theorem ~\cite{hornik1989multilayer}.


\subsection{Kolmogorov-Arnold Networks (KANs)}
\label{KANs_Arch}

Kolmogorov-Arnold Networks (KANs) are inspired by the Kolmogorov-Arnold representation theorem, which states that any multivariate continuous function \( f(\bm{x}) \) on a bounded domain can be represented as a finite composition of univariate functions and addition ~\cite{liu2024kan}. The function \( f(\bm{x}) \) can be approximated using KANs as:
\begin{equation}
    f(\bm{x}) \approx \sum_{i_{L-1}=1}^{n_{L-1}} \phi_{L-1,i_L,i_{L-1}} \left( \cdots \phi_{1,i_2,i_1} \left( \sum_{i_0=1}^{n_0} \phi_{0,i_1,i_0}(x_{i_0}) \right) \cdots \right).
\label{KAN_net}
\end{equation}
Here, \( L \) is the number of layers, \( n_j \) is the number of neurons in the \( j \)-th layer, and \( \phi_{i,j,k} \) are univariate activation functions.

In ~\cite{liu2024kan}, \( \phi(x) \) was proposed as a combination of basis functions \( b(x) \) and B-splines:
\begin{equation}
    \phi(x) = w_b b(x) + w_s \text{spline}(x),
\end{equation}
where \( w_b \), \( w_s \), and \( c_n \) are trainable parameters. The spline function is defined as:
\begin{align*}
    b(x) &= \frac{x}{1+e^{-x}}, \quad
    \text{spline}(x) = \sum_n c_n B_n(x),
\end{align*}
with splines \( B_n(x) \) characterized by polynomial order \( k \) and grid size \( g \).  


Recursive Chebyshev KANs (cKANs) are used as a baseline to reduce computational costs and improve stability as introduced in~\cite{shukla2024comprehensive}. In cKANs, the univariate functions are defined as:
\begin{align}
    \phi(x) = \sum_{n=1}^{D} C_n T_n(\tanh(x)),
\end{align}
where, \( T_n(x) \) are Chebyshev polynomials, computed recursively:
\begin{equation}
    T_{n+1}(x) = 2xT_n(x) - T_{n-1}(x).
\label{chebyshev_recursive}
\end{equation}
Embedding \(\tanh(x)\) ensures normalization for these polynomials. 


\subsection{KKANs}
\label{KKAN}
The KKAN framework combines a flexible inner block with customizable basis functions in its outer block, enabling high adaptability and accuracy. This formulation integrates multiple components, including polynomial embeddings and specialized basis functions. The complete model is described below.

\subsubsection{Inner Block (ebMLP)}

The inner block computes the feature space embedding, \( \Psi(x) \), for the input variables. For each input dimension \( x_i \), we proceed as follows:

\begin{enumerate}
    \item \textbf{Expand the Input Dimension:}
    \begin{equation}
    H_i^0 = [C_0, T_0(x_i), \cdots, C_{D_e} T_{D_e}(x_i)],
    \end{equation}
    where \( D_e \) is the polynomial degree, \( T_j \) denotes the Chebyshev polynomials, and \( C_j \) are trainable parameters.

    \item \textbf{Apply an MLP with \( L \) Layers:} Each layer is defined as:
    \begin{equation}
    H_i^l = \sigma(W^{l-1} \cdot H_i^{l-1} + b^{l}),
    \end{equation}
    where \( \theta^l = \{W^l, b^l\} \) are the weights and biases of the $l-$th layer, and \( \sigma \) is the activation function.

    \item \textbf{Apply a Second Polynomial Embedding:} Expand the output of the MLP into an \( m \)-dimensional space:
    \begin{align}
    \Psi_i(x_i) &= [C^L_0, T_0(H_i^{L}), \cdots, C^L_{D_e} T_{D_e}(H_i^{L})], \\
    \Psi_i(x_i) &= [\Psi_{i,0}, \cdots, \Psi_{i,m}],
    \end{align}
    where \( C^L_j \) are trainable parameters.
\end{enumerate}

Next, a \textbf{combination layer} aggregates the outputs along the input-dimension coordinate:
\begin{equation}
\xi_q = \sum_{i=1}^{d} \Psi_{q,i}(x_i),
\end{equation}
where \( d \) is the input dimension.

\subsubsection{Outer Block (Basis Functions)}

The outer block, denoted as \( g(\cdot) \), applies a specialized basis function to the output of the inner block. The basis functions explored in this study are described as follows:
\label{Basis}

\subparagraph{Chebyshev} 
For this case, the formulation follows \eqref{chebyshev_recursive}, where the trainable parameters \( C_n \) are initialized from a normal distribution with mean \( 0 \) and variance \( \frac{1}{I(D+1)} \), as described in ~\cite{ss2024chebyshev}.

\subparagraph{Legendre} 
This case is similar to the Chebyshev basis but with Legendre polynomials \( L_n(x) \), which are computed recursively:
\[
L_{n+1}(x) = \frac{(2n-1)}{n}xL_{n}(x) - \frac{(n-1)}{n}L_{n-1}(x).
\]
The trainable parameters \( C_n \) are also initialized from a normal distribution with mean \( 0 \) and variance \( \frac{1}{I(D+1)} \), as described in ~\cite{ss2024chebyshev}.

\subparagraph{Sin Series} 
This basis, introduced in ~\cite{guilhoto2024deeplearningalternativeskolmogorov}, has shown improved performance compared to the vanilla KAN. The basis functions are defined as:
\[
\phi(x) = \sum_{i=1}^{D} C_i b_i,
\]
where each \( b_i(t) \) is given by:
\[
b_i(t) = \frac{\sin(w_i t + p_i) - \mu(w_i, p_i)}{\sigma(w_i, p_i)}.
\]
Here, the frequencies \( w_i \) are initialized from a standard normal distribution (\( \mathcal{N}(0, 1) \)), and the phases \( p_i \) are initialized as \( 0 \). The mean and standard deviation are defined as:
\[
\mu(w_i, p_i) = e^{-w_i^2/2}\sin(p_i), \quad \sigma(w_i, p_i) = \sqrt{\frac{1}{2} - e^{-w_i^2} \cos(p_i) - \mu(w_i, p_i)^2}.
\]

\subparagraph{Chebyshev Grid} 
Combining previous approaches, this basis introduces a sub-expansion of the input using a linear layer within the normalization step. The basis is defined as:
\[
\phi(x) = \sum_{n=1}^{D} C_n T_n\left(\sum_{i=1}^{c} \tanh(W_i x + b_i)\right),
\]
where \( C_n \) are initialized as described in ~\cite{ss2024chebyshev} with \( \mathcal{N}(0, \frac{1}{I(D+1)}) \). The centers \( b_i \) are initialized on a grid in the range \([-0.1, 0.1]\) with c=5, and \( W_i \) is initialized from \( \mathcal{N}(0, \frac{1}{I c}) \). This approach improves expressiveness by displacing the input across multiple centers.
\subparagraph{Radial Basis Functions (RBF)} 
For this case, we follow~\cite{li2024kolmogorov} and define the basis functions as:
\[
\phi(x) = \sum_{n=1}^{D} C_n e^{-\frac{(x - p_n)^2}{2\sigma^2}},
\]
where \( C_n \) are initialized from \( \mathcal{N}\left(0, \frac{1}{I(D+1)}\right) \), and \( p_n \) are initialized from a uniform grid spanning \((-2.0, 2.0)\) with \( D \) steps. Here, \( D \) represents the number of centers, and \( \sigma \) is a hyperparameter controlling the spread of the basis functions.

\subparagraph{Single Radial Basis Functions (RBF-Single)} 
To further evaluate the versatility of our 2-block representation framework, we consider a simplified case. Here, the KART architecture \eqref{KART_arch} is rewritten using a single outer function as follows:
\[
f(x_1, \ldots, x_d) = \sum_{q=0}^{m} G\left(\sum_{p=1}^d \Psi_{p, q}(x_p)\right).
\]

This formulation demonstrates that our framework can seamlessly adapt to representations with a single outer block. We implement and test this example using Radial Basis Functions (RBF), referring to it as ''RBF-Single" throughout the study.

\subsubsection{Full Model}

Finally, we combine the inner and outer blocks as follows:

\begin{equation}
f(x_1,\cdots,x_d) = \sum_{q=0}^mg_q(\xi_q).
\end{equation}

\section{Additional Enhancements}

\subsection{Weight Normalized Modified multi-layer perceptrons}
\label{WN_MMLP}

For this part, we combine the modified multi-layer perceptron (mMLP) introduced in ~\cite{wang2021understanding} and the weight normalization proposed in~\cite{salimans2016weight}. The mMLP aims to augment the efficacy of PIML by embedding the input variables $x$ into the hidden states of the network. On the other hand, weight normalization is a reparameterization technique that accelerates convergence in PIML ~\cite{raissi2020hidden}.  In particular, the inputs are encoded in a feature space by employing two distinct encoders, $U$ and $V$, given by:

\begin{equation}
U = \sigma(x^{0}W^U + b^U), \quad V = \sigma(x^{0}W^V + b^V)
\end{equation}

 The encoders are then assimilated within each hidden layer of a conventional MLP by point-wise multiplication. Thus, each forward pass becomes:

\begin{equation}
\alpha^{l}(x) = \alpha^{l-1}(x)W^{l} + b^{l}, \quad for \ \ l \in \{1, 2, ..., L\}
\end{equation}

\begin{equation}
\alpha^{l}(x) = \sigma(\alpha^{l}(x))
\end{equation}

\begin{equation}
\alpha^{l}(x) = (1-\alpha^{l}(x)) \odot U + \alpha^{l}(x) \odot V,
\end{equation}

\noindent where $x$ is the input, $\alpha^{l}$ and $W^{l}$ are the neurons and weights of layer $l$, $\sigma$ is the activation function and $\odot$ is the element-wise product.  Finally, we include WN by reparameterizing our weights as: 

\begin{align}
    \alpha&=\sigma(W \cdot x + b)\\
    W&=\frac{g}{\lVert\mathbf{v}\rVert_{2}}\mathbf{v}\text{,}
    \label{eq_wn}
\end{align}

\noindent where $\alpha$ is the neuron output, $\sigma$ is the activation function, $x$ is the input vector, $W$ is a weight vector, and $b$ is the bias. As shown in \eqref{eq_wn}, the weight vector $W$ is redefined in terms of new trainable parameters, $\mathbf{v}$ (direction) and $g$ (length). Notice that $||W||=g$, so this reparameterization allows us to decouple the weight's length and direction, which speeds up the model convergence. Since $g$ is a scalar, this modification induces minimal computational overhead ~\cite{salimans2016weight}.

\subsection{Weight-Normalized Adaptive ResNet (WNadResNet)}
\label{AdResNet}

The Weight-Normalized Adaptive ResNet (WNadResNet) builds on the concept of adaptive residual connections proposed in~\cite{wang2024piratenets}. However, our implementation focuses on reducing computational overhead while retaining flexibility and performance. Unlike the approach in~\cite{wang2024piratenets}, which incorporated modified MLPs, we simplify the architecture to achieve comparable performance with KKANs while significantly lowering computational costs.

For each WNadResNet layer, the forward pass is computed as follows:
\begin{enumerate}
    \item Compute the transformed feature \( F \) from the input \( H \) using a weight-normalized layer and a nonlinear activation:
\begin{equation}
F = \sigma(W \cdot H + b),
\end{equation}
where \( W \) and \( b \) are the weight and bias parameters of the layer, and \( \sigma \) is the hyperbolic tangent activation (\( \tanh \)). 
\item Apply a second transformation \( G \) to \( F \):
\begin{equation}
G = W' \cdot F + b',
\end{equation}
where \( W' \) and \( b' \) are parameters of another weight-normalized layer.
\item  Combine the original input \( H \) with the transformed feature \( G \) through an adaptive residual connection:
\begin{equation}
H' = \sigma(\alpha \cdot G + (1 - \alpha) \cdot H),
\end{equation}
where \( \alpha \) is a trainable scalar that adaptively balances the contributions of \( G \) and \( H \).
\end{enumerate}

The weights \( W \) and \( W' \) are reparameterized using weight normalization ~\cite{salimans2016weight}.

By introducing the adaptive parameter \( \alpha \), WNadResNet allows the network to dynamically adjust the residual contribution of the input, enhancing learning flexibility. This framework is particularly effective for KKANs, enabling high performance with minimal computational overhead.

\section{Deep Operator Network (DeepONet)}
\label{DeepONet_appendix}
In this section, we describe the DeepONet~\cite{lu2019deeponet} and QR-DeepONet~\cite{lee2024training} frameworks as described in~\cite{lee2024training}.
\subsection{Operators}
\smallskip

Let $\Omega_X \subset \mathbb{R}^{d_x}$, $\Omega_Y \subset \mathbb{R}^{d_y}$, and $(\mathcal{X}, d_{\mathcal{X}})$ represent a metric space of functions defined over $\Omega_X$. Additionally, let $(\mathcal{Y}, \|\cdot\|_{\mathcal{Y}})$ be a normed vector space of functions defined over $\Omega_Y$. The operator of interest is denoted as:
\[
\mathcal{G}: \mathcal{X} \ni f \mapsto \mathcal{G}[f] \in \mathcal{Y}.
\]

To approximate $\mathcal{G}$, the function $f(x)$ is discretized using a suitable basis $\{\phi_i\}$:
\[
f(x) = \sum_{i=1}^\infty \hat{f}_i \phi_i(x), \quad \hat{f}_i = \langle f, \phi_i \rangle,
\]
where $\langle \cdot, \cdot \rangle$ is an appropriate inner product. The discrete representation of $f$ is given by:
\[
\mathbf{f} = (\hat{f}_1, \dots, \hat{f}_{m_x}),
\]
where $m_x$ is the number of sensors used for the input function $f$. The goal is to approximate the operator $\mathcal{G}$ using a neural network-based model $\mathcal{G}_{NN}$.

\subsection{DeepONet}
\smallskip

For $L \in \mathbb{N}$ and $\mathbf{n} = (n_0, n_1, \dots, n_L) \in \mathbb{N}^{L+1}$, a representation model $\mathcal{M}$ with $L$ layers maps an input $x \in \mathbb{R}^{n_0}$ to $z^L \in \mathbb{R}^{n_L}$. Here, $\mathbf{n}$ specifies the network architecture.

\subsubsection{DeepONet Structure}

DeepONet architectures consist of two main components: the \textbf{branch network} and the \textbf{trunk network}.

\paragraph{Branch Network} The branch network $\mathbf{c}(\cdot; \theta)$ is a vector-valued representation model with $L_b$ layers:
\[
\mathbf{c}(\cdot; \theta) = (c_0(\cdot; \theta), \dots, c_N(\cdot; \theta))^T,
\]
where its architecture is defined as $\mathbf{n}_b = (m_x, n_1^{(b)}, \dots, N+1)$, and $\theta$ represents the trainable parameters.

\paragraph{Trunk Network} The trunk network $\phi(\cdot; \mu)$ is a vector-valued representation model with $L_t$ layers:
\[
\phi(\cdot; \mu) = (1, \phi_0(\cdot; \mu))^T,
\]
where $\phi_0(\cdot; \mu) = (\phi_1(\cdot; \mu), \dots, \phi_N(\cdot; \mu))$ with an architecture $(d_y, n_1^{(t)}, \dots, n_{L_t-1}^{(t)}, N)$. Here, $\mu$ represents the trainable parameters.

\subsubsection{DeepONet Approximation}
\smallskip

The output of DeepONet is defined as the inner product of the branch and trunk networks:
\[
O_{\text{net}}[\mathbf{f}; \Theta](y) = \phi^T(y; \mu) \mathbf{c}(\mathbf{f}; \theta),
\]
which can be expanded as:
\[
O_{\text{net}}[\mathbf{f}; \Theta](y) = c_0(\mathbf{f}; \theta) + \sum_{j=1}^N c_j(\mathbf{f}; \theta) \phi_j(y; \mu).
\]
Here, $\Theta = \{\mu, \theta\}$ is the set of trainable parameters of the DeepONet.

\subsubsection{Training}
\smallskip

 Let $\{f_k\}_{k=1}^{K}$ be a set of input functions from $\mathcal{X}$ and $u_{k}(\cdot) = \mathcal{G}[f_k](\cdot)$ be the corresponding output functions in $\mathcal{Y}$. Let $||\cdot||_{\mathcal{Y}_{m_y}}$ be a discretized norm. The objective is to optimize the parameters of DeepONet by minimizing the following loss:
\[
\mathcal{L}(\Theta) = \frac{1}{K}\sum_{k=1}^{K} ||O_{\text{net}}[\mathbf{f}_k; \Theta](\cdot) - u_k(\cdot)||_{\mathcal{Y}_{m_y}}^p.
\]

\subsection{QR-DeepONet}

QR-DeepONet extends the standard DeepONet framework by introducing a reparameterization of the trunk network and leveraging a QR decomposition for improved stability during training~\cite{lee2024training}.

The functional space $\mathcal{Y}$ is defined as $L_{\omega}^p(\Omega_y)$, with the norm given by:
\[
||g||_{\mathcal{Y}} = \left(\int_{\Omega_y} |g(y)|^p d\omega\right)^{1/p}, \quad \forall g \in \mathcal{Y},
\]
where $\omega$ is a probability measure satisfying $\int_{\Omega_y} d\omega(y) = 1$. 

For practical applications, the discrete version of this norm, computed via Monte Carlo sampling, is given by:
\[
||g||_{\mathcal{Y}_{m_y}} = \left(\frac{1}{m_y} \sum_{i=1}^{m_y} |g(y_i)|^p \omega(y_i)\right)^{1/p}, \quad \forall g \in \mathcal{Y},
\]
where $\{y_i\}_{i=1}^{m_y}$ are i.i.d. random samples from $\omega$, and $m_y$ is the number of output sensors. 

To discretize the function $g \in \mathcal{Y}$, let:
\[
\mathbf{g} = (g(y_1), \dots, g(y_{m_y}))^T.
\]
The training data is then represented as:
\[
(\mathbf{f}_k, \mathbf{u}_k) = (f_k(x_1), \dots, f_k(x_{m_x}), u_k(y_1), \dots, u_k(y_{m_y})), \quad k = 1, \dots, K.
\]

\subsubsection{Loss Function}
\smallskip

The QR-DeepONet loss function is defined as:
\[
\mathcal{L}(\{\mu, \theta\}) = \frac{1}{K} \sum_{k=1}^K \frac{1}{m_y} \sum_{i=1}^{m_y} \left| \phi^T(y_i; \mu)\mathbf{c}(\mathbf{f}_k; \theta) - u_k(y_i) \right|^p.
\]

This loss function can be reformulated using matrix representations for computational efficiency. Define:
\begin{itemize}
    \item The trunk matrix:
\[
\mathbf{\Phi}(\mu) = 
\begin{bmatrix}
\phi^T(y_1; \mu) \\
\vdots \\
\phi^T(y_{m_y}; \mu)
\end{bmatrix} \in \mathbb{R}^{m_y \times (N+1)}.
\]
\item The branch matrix:
\[
\mathbf{C}(\theta) = [\mathbf{c}(\mathbf{f}_1; \theta), \dots, \mathbf{c}(\mathbf{f}_K; \theta)] \in \mathbb{R}^{(N+1) \times K}.
\]
\item The target output matrix:
\[
\mathbf{U} = [\mathbf{u}_1, \dots, \mathbf{u}_K] \in \mathbb{R}^{m_y \times K}.
\]
\end{itemize}

Using these matrices, the loss function can be compactly expressed as:
\[
\mathcal{L}(\{\mu, \theta\}) = \frac{1}{Km_y} ||\mathbf{\Phi}(\mu) \mathbf{C}(\theta) - \mathbf{U}||_{p,p}^p.
\]

\subsubsection{Reparameterization of the Trunk Network}
\smallskip

Let $T \in \mathbb{R}^{(N+1) \times (N+1)}$ be a trainable square matrix. The trunk network is reparameterized as:
\[
\hat{\phi}(\cdot; \mu, T) = T^T \phi(\cdot; \mu).
\]
The output of the network becomes:
\[
O_{\text{net}}[\mathbf{f}](y) = \hat{\phi}^T(y; \mu, T) \mathbf{c}(\mathbf{f}; \theta) = \phi^T(y; \mu) T \mathbf{c}(\mathbf{f}; \theta).
\]
This reparameterization modifies the original branch network $\mathbf{c}(\mathbf{f}; \theta)$ into $T \mathbf{c}(\mathbf{f}; \theta)$. For standard MLPs, this transformation adjusts the last layer’s weights and biases:
\[
T \mathbf{c}(\mathbf{f}; \theta) = T W^{L_b} \sigma(z^{L_b-1}) + T b^{L_b}.
\]

\subsubsection{Two-Step Training}
To optimize QR-DeepONet, a two-step training procedure is employed:
\begin{enumerate}
    \item \textbf{Optimize the Trunk Network.} First, the trunk network parameters $\mu$ and an auxiliary matrix $A \in \mathbb{R}^{(N+1) \times K}$ are optimized by solving:
    \begin{equation}
    \min_{\mu, A} ||\mathbf{\Phi}(\mu) A - \mathbf{U}||_{p,p}^p.
    \label{train_trunk}
    \end{equation}
    After optimization, perform a QR decomposition of $\mathbf{\Phi}(\mu^*)$:
    \[
    \mathbf{\Phi}(\mu^*) = Q^*R^*,
    \]
    where $Q^*$ is orthogonal, and $R^*$ is upper triangular. Set $T^* = (R^*)^{-1}$.
    \item \textbf{Optimize the Branch Network.} Using the precomputed $R^*$ and $A^*$, optimize the branch network parameters $\theta$ by solving:
    \begin{equation}
    \min_{\theta} ||\mathbf{C}(\theta) - R^* A^*||. 
    \label{train_branch}
    \end{equation}
\end{enumerate}

This two-step process decouples the optimization of the trunk and branch networks, leveraging the QR decomposition to ensure numerical stability and effective training.

\section{self-scaled Residual-based-attention Algorithm}

\vspace{5mm}
\begin{algorithm}[H]
  \caption{self-scaled Residual-based attention}
  \label{ssRBA_alg}
  \textbf{Input:} \\
  Representation model (i.e., MLP, cKAN, or KKAN): $\mathcal{M}$\\
  Training points: $X_D$ and/or $X_B$,$X_E$\\
  Optimizer parameters: $lr$ for MLP or cKAN and $lr_{\Psi},lr_{G}$ for  KKAN.\\
  ssRBA parameters: $\eta$, $\lambda_{max_0}$,$\nu$, $c$,$\lambda_{cap}$, $\alpha$, $m_E$, $\gamma_{g}$\\
  Number of iterations per stage: $N_{stage}$\\
  Total number of training of iterations: $N_{train}$\\
 \textbf{Output:} \\
 Optimized network parameters $\theta$
  \begin{algorithmic}[1]
  \STATE Initialize the network parameters: $\theta$ for MLP or $\theta=\{\theta_\psi,\theta_g\}$ for KKAN
    \STATE Initialize RBA: $\lambda^0_{\alpha,i}=0.1\lambda_{max0}$  $\forall \alpha,i$  with $\alpha,i=\{B,D,E\}$\\
  \FOR{$k<N_{train}$}
  \STATE Update maximum RBA upper bound: $\lambda_{max}=\max(\lambda_{max0}+k//N_{stage},\lambda_{cap})$
    \STATE Update decay rate: $\gamma^k=1-\eta/\lambda_{max}$
  \FOR{each $\alpha\in\{B,D,E\}$}
  \STATE Compute the sampling p.d.f.s: $   p_{\alpha}^{(k)}\leftarrow(\bm{\lambda}_{\alpha}^{(k)})^{\nu}/{\mathbb{E}[(\bm{\lambda}_{\alpha}^{(k)})^{\nu}]}+c$
  \STATE Sample $bs$ points from $X_{\alpha}$: $X^k_{\alpha}\sim p_{\alpha}^{(k)}$
  \STATE Compute network prediction: $u_{\alpha,i}\leftarrow\mathcal{M}(\theta, x^k_{\alpha,i})$ for $\forall x_{\alpha,i}\in X^k_{\alpha} $
  \STATE Compute residuals: $r_{\alpha,i}$ using $u_{\alpha,i}$ and equations~\ref{bcs_res}, ~\ref{data_res} or ~\ref{PDE_res}.
      \STATE Update RBA: $\lambda_{\alpha,i}^{k+1} \leftarrow \gamma^k\lambda_{\alpha,i}^{k}+\eta\|r_{\alpha,i}|/{\max_j |r_{\alpha,j}|}$
      \STATE Compute loss term: $\mathcal{L}^k_{\alpha}=\langle(\lambda_{\alpha,i}^{k}r_{\alpha,i}^{k})^2\rangle$
      \STATE Compute gradient:$\nabla_{\theta}\mathcal{L}^k_{\alpha}$
      \STATE Compute the average gradient magnitude :$\|\nabla_{\theta}\bar{\mathcal{L}}^k_{\alpha}\|=\gamma_g\|\nabla_{\theta}\bar{\mathcal{L}}^{k-1}_{\alpha}\|+(1-\gamma_g)\|\nabla_{\theta}\mathcal{L}^{k-1}_{\alpha}\|$
    \ENDFOR
    \STATE Update data global weight: $m_D^{k}=\alpha m_D^{k-1}+(1-\alpha)m_E\|\nabla_{\theta}\bar{\mathcal{L}_E}^k\|/{\|\nabla_{\theta}\bar{\mathcal{L}_D}^k\|}$
    \STATE Update boundary global weight: $m_B^{k}=\alpha m_B^{k-1}+(1-\alpha)m_E\|\nabla_{\theta}\bar{\mathcal{L}_E}^k\|/{\|\nabla_{\theta}\bar{\mathcal{L}_B}^k\|}$
    \STATE Define total update direction: $     p^k\leftarrow-m_E\nabla_{\theta}\mathcal{L}_E^{k}-m_B^k\nabla_{\theta}\mathcal{L}_B^{k}-m_D^k\nabla_{\theta}\mathcal{L}_D^{k}$
    \IF{$\mathcal{M} = \text{KKAN}$}
    \STATE Update inner-block parameters: $\theta_\psi^{k+1}\leftarrow\theta^{k}_\psi-lr_\psi^kp^k_\psi$\\
    \STATE Update outer-block parameters:     $\theta_g^{k+1}\leftarrow\theta^{k}_g-lr_g^kp^k_g$
    \ELSE
    \STATE Update parameters: $\theta^{k+1}\leftarrow\theta^{k}-lr^kp^k$
    \ENDIF
  \ENDFOR
  \end{algorithmic}
\end{algorithm}

\begin{figure}[t]
    \centering
    \includegraphics[width=1\linewidth]{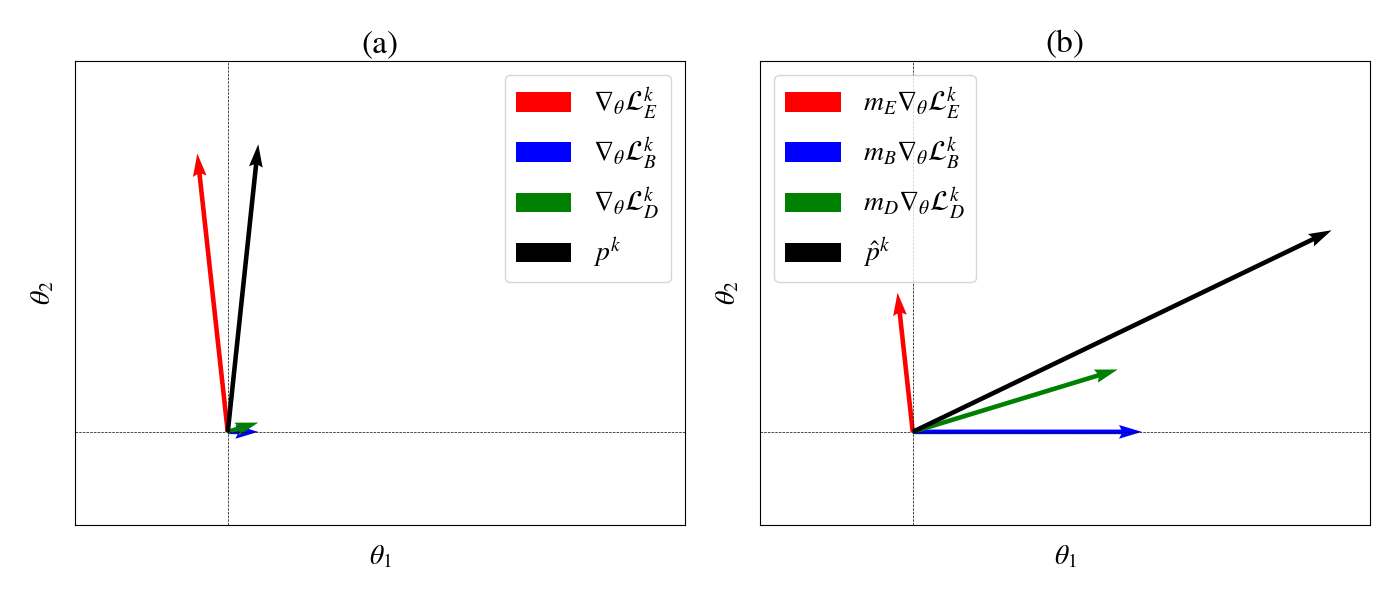}
\caption{Visualization of loss gradients as directional vectors for an idealized models model parameterized by $\theta=\{\theta_1,\theta_2\}$. (a) When the magnitude of one component dominates, the update direction $p^k$ becomes imbalanced, leading to poor convergence. (b) By applying appropriate global weights, the update direction is scaled, resulting in a balanced update $\hat{p}^k$ that accommodates all directions.}

    \label{Grads_dirs}
\end{figure}
\section{Implementation Details}
\label{Imp_Details}

All models were trained using the ADAM optimizer~\cite{kingma2014adam} with an exponential learning rate decay. The training was conducted using JAX with a single precision on an NVIDIA GeForce RTX 3090 GPU.

For KKAN, the inner blocks were constructed using enhanced basis MLPs (ebMLPs), and the outer blocks utilized suitable basis functions. The activation function for both MLPs and KKANs was the hyperbolic tangent, $\tanh(\cdot)$. Weight normalization~\cite{salimans2016weight} was applied to all models to stabilize training and improve convergence. For cKAN models, we follow ~\cite{ss2024chebyshev} and initialize our parameters using a truncated normal distribution with a standard deviation given by $\sigma = \frac{1}{I(D+1)}$ where $I$ is the input dimension and $D$ is the polynomial degree. For KKAN models, we initialize our basis functions as described in ~\ref{Basis}.

Further specific implementation details for each experiment are described in the following sections.
\subsection{Function Approximation}
\subsubsection{Discontinuous}
\begin{table}[H]
\centering
\begin{tabular}{|l|c|c|c|}\hline
Hyperparameter&MLP&cKAN&KKAN\\\hline
$N_{train}$&2e5&2e5&2e5\\
Number of hidden layers $N$&5&4&4\\
Hidden layer dimension $H$&100&40&32\\
Number of KKAN features $m$&&&32\\
Polynomial degree $D$&&7&7\\
ebMLP Polynomial degree $D_e$&&&7\\
Initialization&Glorot~\cite{glorot2010understanding}&$\mathcal{N}\left(0,\frac{1}{I(D+1)}\right)$ ~\cite{ss2024chebyshev} &Glorot~\cite{glorot2010understanding}\\
Learning rate $lr$ & 1e-3 &2e-4&\\
KKAN: Learning rate inner $lr_{\Psi}$ & & &1e-3 \\
KKAN: Learning rate outer $lr_{G}$& & & 2e-4\\
$lr-$Decay rate & 0.9& 0.9& 0.9\\
$lr-$Decay step & 5000& 5000& 5000\\

ssRBA:$\gamma$&0.999&0.999&0.999\\
ssRBA:$\eta$&0.01&0.01&0.01\\
ssRBA:$\lambda_{max0}$&10&10&10\\
ssRBA:$\lambda_{cap}$&20&20&20\\
ssRBA:$N_{stage}$&50000&50000&50000\\
ssRBA:$m_E$&1.0&1.0&1.0\\
\hline
\end{tabular}
\caption{Implementation details for discontinuous function approximation. $I$ denotes the input dimension. For KKAN, there are two polynomial degrees: \( D \) for the outer blocks and \( D_e \) for the ebMLP used in the inner blocks. The initialization details for the outer blocks are described in the preceding section.}
\label{Disc_imp_dets}
\end{table}
\subsubsection{Smooth}
\begin{table}[H]
\centering
\begin{tabular}{|l|c|c|c|}\hline
Hyperparameter&MLP&cKAN&KKAN\\\hline
$N_{train}$&2e5&2e5&2e5\\
Number of hidden layers $N$&5&4&4\\
Hidden layer dimension $H$&100&40&32\\
Number of KKAN features $m$&&&32\\
Polynomial degree $D$&&5&15\\
ebMLP Polynomial degree $D_e$&&&15\\
Initialization&Glorot~\cite{glorot2010understanding}&$\mathcal{N}\left(0,\frac{1}{I(D+1)}\right)$ ~\cite{ss2024chebyshev} &Glorot~\cite{glorot2010understanding}\\
Learning rate $lr$ & 1e-3 &2e-4&\\
KKAN: Learning rate inner $lr_{\Psi}$ & & &1e-3 \\
KKAN: Learning rate outer $lr_{G}$& & & 2e-4\\
$lr-$Decay rate & 0.9& 0.9& 0.9\\
$lr-$Decay step & 5000& 5000& 5000\\

ssRBA:$\gamma$&0.999&0.999&0.999\\
ssRBA:$\eta$&0.01&0.01&0.01\\
ssRBA:$\lambda_{max0}$&10&10&10\\
ssRBA:$\lambda_{cap}$&20&20&20\\
ssRBA:$N_{stage}$&50000&50000&50000\\
ssRBA:$m_E$&1.0&1.0&1.0\\
\hline
\end{tabular}
\caption{Implementation details for smooth function approximation. $I$ denotes the input dimension. For KKAN, there are two polynomial degrees: \( D \) for the outer blocks and \( D_e \) for the ebMLP used in the inner blocks. The initialization details for the outer blocks are described in the preceding section.}
\label{Smooth_imp_dets}
\end{table}
\subsection{Physics-informed Machine Learning}
\subsubsection{Allen-Cahn Equation}

\begin{table}[H]
\centering
\begin{tabular}{|l|c|c|c|}\hline
Hyperparameter&MLP&cKAN&KKAN\\\hline
$N_{train}$&3e5&3e5&3e5\\
Number of hidden layers $N$&6&4&4\\
Hidden layer dimension $H$&64&32&32\\
Number of KKAN features $m$&&&64\\
Polynomial degree $D$&&5&9\\
ebMLP Polynomial degree $D_e$&&&2\\
Initialization&Glorot~\cite{glorot2010understanding}&$\mathcal{N}\left(0,\frac{1}{I(D+1)}\right)$ ~\cite{ss2024chebyshev} &$U\left(-\sqrt{\frac{3}{I}},\sqrt{\frac{3}{I}}\right)$\\
Learning rate $lr$ & 1e-3 &2e-4&\\
KKAN: Learning rate inner $lr_{\Psi}$ & & &1e-3 \\
KKAN: Learning rate outer $lr_{G}$& & & 2e-4\\
$lr-$Decay rate & 0.9& 0.9& 0.9\\
$lr-$Decay step & 5000& 5000& 5000\\
Batch size&1e4&1e4&1e4\\

\hline
\end{tabular}
\caption{Implementation details for solving the Allen-Cahn Equation part (a). KKAN models include features of size \( m = 64 \), with polynomial degrees \( D \) for the outer blocks and \( D_e \) for the ebMLP inner blocks. Initialization strategies are Glorot~\cite{glorot2010understanding} for MLPs, Gaussian-based for cKAN, and uniform for KKAN as detailed in the table. Learning rates (\( lr \)) and decay parameters are specified for each architecture.}
\label{ACa_dets}
\end{table}

\begin{table}[H]
\centering
\begin{tabular}{|l|c|c|c|}\hline
Hyperparameter&MLP&cKAN&KKAN\\\hline
$N_{train}$&3e5&3e5&3e5\\
Number of hidden layers $N$&6&4&4\\
Hidden layer dimension $H$&64&32&32\\
Fourier Feature embedding~\cite{wang2021eigenvector} degree&10&10&10\\
Architecture enhancement& WNmMLP& & WNmMLP\\
Number of KKAN features $m$&&&64\\
Polynomial degree $D$&&5&9\\
ebMLP Polynomial degree $D_e$&&&2\\
Initialization&Glorot~\cite{glorot2010understanding}&$\mathcal{N}\left(0,\frac{1}{I(D+1)}\right)$ ~\cite{ss2024chebyshev} &$U\left(-\sqrt{\frac{3}{I}},\sqrt{\frac{3}{I}}\right)$\\
Learning rate $lr$ & 1e-3 &2e-4&\\
KKAN: Learning rate inner $lr_{\Psi}$ & & &1e-3 \\
KKAN: Learning rate outer $lr_{G}$& & & 2e-4\\
$lr-$Decay rate & 0.9& 0.9& 0.9\\
$lr-$Decay step & 5000& 5000& 5000\\
Batch size&1e4&1e4&1e4\\
ssRBA-R:$\gamma$&0.999&0.999&0.999\\
ssRBA-R:$\eta$&0.01&0.01&0.01\\
ssRBA-R:$\lambda_{max0}$&10&10&10\\
ssRBA-R:$\lambda_{cap}$&20&20&20\\
ssRBA-R:$\gamma_g$&0.99&0.99&0.99\\
ssRBA-R:$\alpha$&0.99975&0.99975&0.99975\\
ssRBA-R:$\nu$&2.0&2.0&2.0\\
ssRBA-R:$c$&0.5&0.5&0.5\\
ssRBA-R:$N_{stage}$&50000&50000&50000\\
ssRBA-R:$m_E$&1.0&1.0&1.0\\
\hline
\end{tabular}
\caption{Implementation details for solving the Allen-Cahn Equation part (b). WNmMLP refers to the weight-normalized modified MLP architecture~\cite{salimans2016weight, wang2021understanding}. Fourier feature embeddings are used with a degree of 10 across all models. For KKAN, the architecture includes features of size \( m = 64 \), with polynomial degrees \( D \) and \( D_e \) for the outer blocks and ebMLP inner blocks, respectively. Initialization strategies are Glorot~\cite{glorot2010understanding} for MLPs, Gaussian-based for cKAN, and uniform for KKAN, as described in the table.}
\label{ACb_imp_dets}
\end{table}

\begin{table}[H]
\centering
\begin{tabular}{|l|c|c|c|}\hline
Hyperparameter&MLP&cKAN&KKAN\\\hline
$N_{train}$&3e5&3e5&3e5\\
Number of hidden layers $N$&6&5&4\\
Hidden layer dimension $H$&128&64&64\\
Fourier Feature embedding~\cite{wang2021eigenvector} degree&10&10&10\\
Architecture enhancement& WNmMLP& & WNadResNet\\
Number of KKAN features $m$&&&64\\
Polynomial degree $D$&&5&5\\
ebMLP Polynomial degree $D_e$&&&7\\
Initialization&Glorot~\cite{glorot2010understanding}&$\mathcal{N}\left(0,\frac{1}{I(D+1)}\right)$ ~\cite{ss2024chebyshev} &Glorot~\cite{glorot2010understanding}\\
Learning rate $lr$ & 1e-3 &2e-4&\\
KKAN: Learning rate inner $lr_{\Psi}$ & & &1e-3 \\
KKAN: Learning rate outer $lr_{G}$& & & 2e-4\\
$lr-$Decay rate & 0.9& 0.9& 0.9\\
$lr-$Decay step & 5000& 5000& 5000\\

ssRBA:$\gamma$&0.999&0.999&0.999\\
ssRBA:$\eta$&0.01&0.01&0.01\\
ssRBA:$\lambda_{max0}$&10&10&10\\
ssRBA:$\lambda_{cap}$&20&20&20\\
ssRBA:$\gamma_g$&0.99&0.99&0.99\\
ssRBA:$\alpha$&0.99975&0.99975&0.99975\\
ssRBA:$N_{stage}$&50000&50000&50000\\
ssRBA:$m_E$&1.0&1.0&1.0\\
\hline
\end{tabular}
\caption{Implementation details for solving the Allen-Cahn Equation part (c). WNmMLP refers to the weight-normalized modified MLP architecture~\cite{salimans2016weight, wang2021understanding}, while WNadResNet refers to the weight-normalized adaptive residual network (see Section~\ref{AdResNet}). Fourier feature embeddings with a degree of 10 are used across all models. For KKAN, the architecture includes features of size \( m = 64 \), with polynomial degrees \( D \) for the outer blocks and \( D_e \) for the ebMLP inner blocks. Initialization strategies are Glorot~\cite{glorot2010understanding} for MLPs and KKANs, and Gaussian-based for cKANs. Learning rates (\( lr \)), decay rates, and ssRBA parameters are listed in the table.}
\label{ACc_imp_dets}
\end{table}
\subsection{Operator Learning}
\subsubsection{Burgers Equation}

\begin{table}[H]
\centering
\begin{tabular}{|l|c|c|c|}\hline
Hyperparameter&MLP&cKAN&KKAN\\\hline
$N_{train}$&4e5&4e5&4e5\\
Number of hidden layers $N$&6&5&5\\
Hidden layer dimension $H$&100&32&32\\
Embedding dimension &100&100&100\\
Number of KKAN features $m$&&&32\\
Polynomial degree $D$&&5&5\\
ebMLP Polynomial degree $D_e$&&&5\\
Architecture enhancement& & & WNadResNet\\
Initialization&Glorot~\cite{glorot2010understanding}&$\mathcal{N}\left(0,\frac{1}{I(D+1)}\right)$ ~\cite{ss2024chebyshev} &~\cite{glorot2010understanding}\\
Learning rate $lr$ & 1e-3 &3e-4&\\
KKAN: Learning rate inner $lr_{\Psi}$ & & &1e-3 \\
KKAN: Learning rate outer $lr_{G}$& & & 1e-3\\
$lr-$Decay rate & 0.9& 0.9& 0.99\\
$lr-$Decay step & 2500& 2500& 5000\\
\hline
\end{tabular}
\caption{Implementation details for the Burgers equation using the DeepONet framework. The embedding dimension represents the number of neurons in the last layer of the branch and trunk networks. KKAN models include features of size \( m = 32 \), with polynomial degrees \( D \) for the outer blocks and \( D_e \) for the ebMLP inner blocks. Initialization strategies are Glorot~\cite{glorot2010understanding} for MLPs and KKANs, and Gaussian-based for cKANs. Learning rates (\( lr \)), decay rates, and other hyperparameters are detailed in the table. WNadResNet refers to the weight-normalized adaptive residual network used in KKANs for enhanced learning performance.}
\label{DONet_dets}
\end{table}

\begin{table}[H]
\centering
\begin{tabular}{|l|c|c|c|}\hline
Hyperparameter&MLP&cKAN&KKAN\\\hline
$N_{train}$ trunk&2e5&2e5&2e5\\
$N_{train}$ branch&4e5&4e5&4e5\\
Number of hidden layers $N$&6&5&5\\
Hidden layer dimension $H$&100&32&32\\
Embedding dimension &100&100&100\\
Number of KKAN features $m$&&&32\\
Polynomial degree $D$&&5&5\\
ebMLP Polynomial degree $D_e$&&&5\\
Architecture enhancement& & & WNadResNet\\
Initialization&Glorot~\cite{glorot2010understanding}&$\mathcal{N}\left(0,\frac{1}{I(D+1)}\right)$ ~\cite{ss2024chebyshev} &~\cite{glorot2010understanding}\\
Learning rate $lr$ & 1e-3 &3e-4&\\
KKAN: Learning rate inner $lr_{\Psi}$ & & &1e-3 \\
KKAN: Learning rate outer $lr_{G}$& & & 1e-3\\
$lr-$Decay rate & 0.9& 0.9& 0.99\\
$lr-$Decay step & 2500& 2500& 5000\\
\hline
\end{tabular}
\caption{Implementation details for the Burgers equation using the QR-DeepONet framework. The embedding dimension refers to the number of neurons in the last layer of the branch and trunk networks. The number of parameters is increased due to the trainable matrix \( A \), which, in this case, has \( 3500 \times 3500 \) parameters. KKAN models include features of size \( m = 32 \), with polynomial degrees \( D \) for the outer blocks and \( D_e \) for the ebMLP inner blocks. Initialization strategies are Glorot~\cite{glorot2010understanding} for MLPs and KKANs, and Gaussian-based for cKANs. Learning rates (\( lr \)) and decay parameters are detailed in the table. WNadResNet refers to the weight-normalized adaptive residual network used in KKANs.}
\label{QR-DONet_dets}
\end{table}

\section{Additional Results}
\subsection{Function Approximation}
\subsubsection{Discontinuous}
\begin{figure}[H]
    \centering
    \includegraphics[width=1\linewidth]{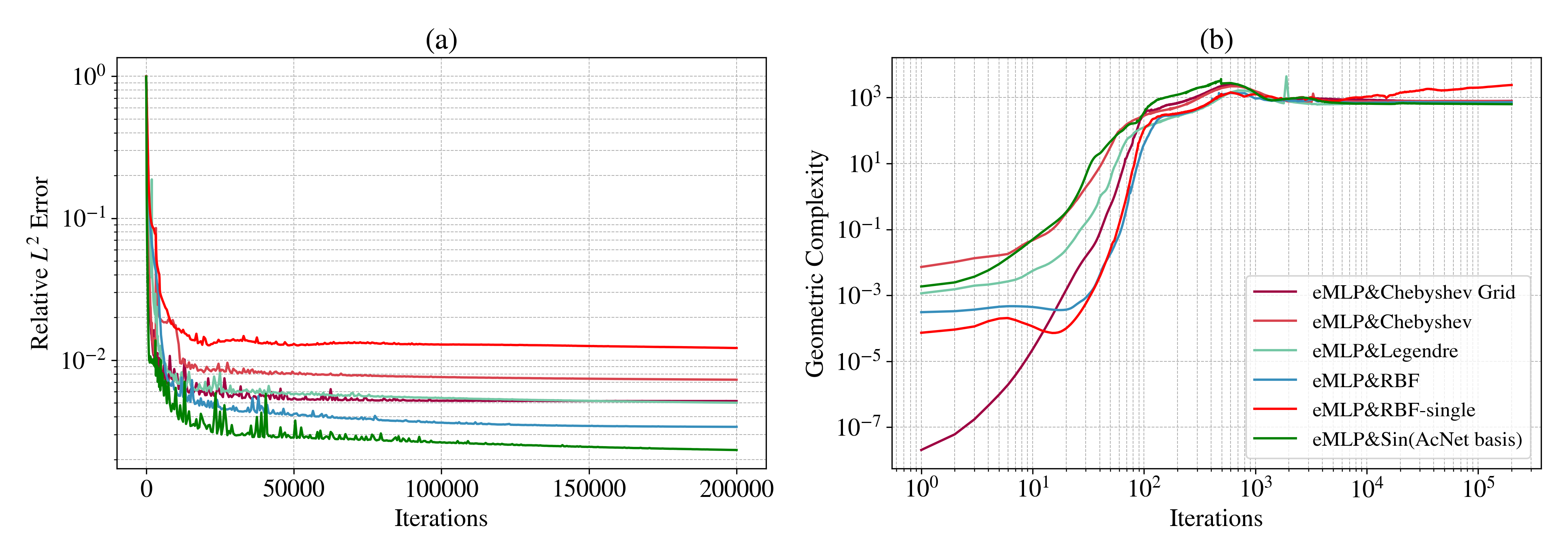}
    \caption{Results for the highly discontinuous function for different basis functions. (a) Relative $L^2$ error convergence on the testing dataset, evaluated on a uniform $256\times256$ mesh. The best-performing model is obtained using the sin-series basis introduced in~\cite{guilhoto2024deeplearningalternativeskolmogorov}. (b) Geometric complexity evolution during training. All models converge to the same complexity except for RBF-Single, which is higher, indicating that it is possibly over-fitting.}
    \label{fig:Disc_Other_basis}
\end{figure}

\subsubsection{Smooth}
\begin{figure}[H]
    \centering
    \includegraphics[width=1\linewidth]{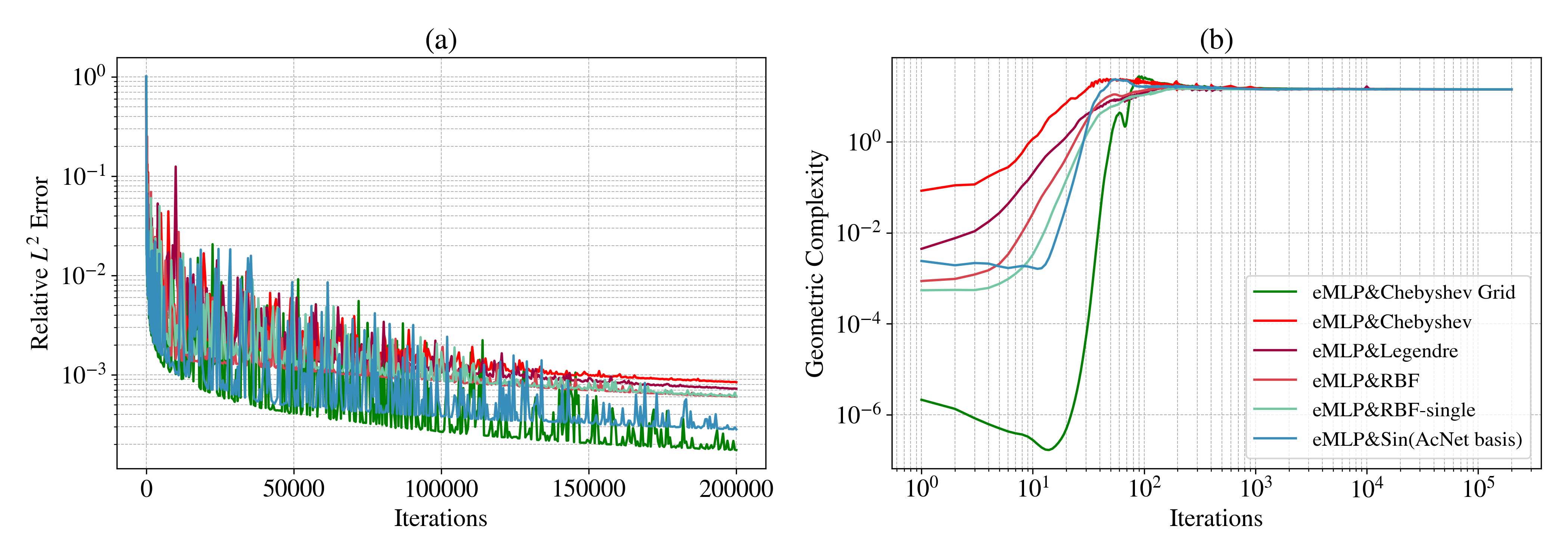}
    \caption{Results for the highly discontinuous function for different basis functions. (a) Relative $L^2$ error convergence on the testing dataset, evaluated on a uniform $256\times256$ mesh. The best-performing model is obtained using the sin-series basis introduced in~\cite{guilhoto2024deeplearningalternativeskolmogorov}. (b) Geometric complexity evolution during training.}
    \label{fig:Smooth_Other_basis}
\end{figure}
\subsection{Physics-Informed Machine Learning}
\subsubsection{Allen Cahn}

\begin{figure}[H]
    \centering
    \includegraphics[width=1\linewidth]{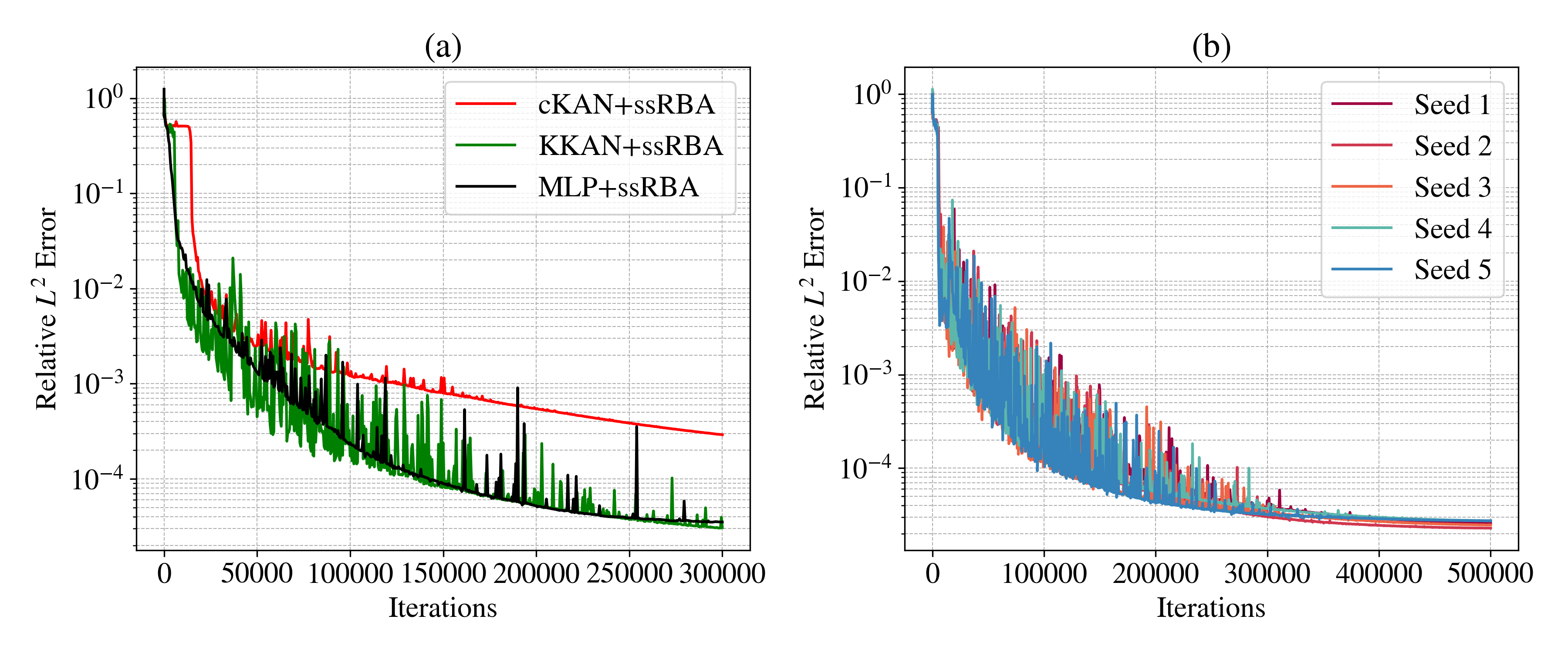}
\caption{Results for solving the Allen-Cahn Equation. (a) Relative $L^2$ error convergence for models trained with ssRBA and full-batch training over 300,000 ADAM iterations. (b) Relative $L^2$ error convergence for KKAN+ssRBA models initialized with five different seeds and trained for 500,000 iterations, demonstrating robustness to initialization.}
    \label{fig:AC-other}
\end{figure}

\begin{figure}[H]
    \centering
    \includegraphics[width=1\linewidth]{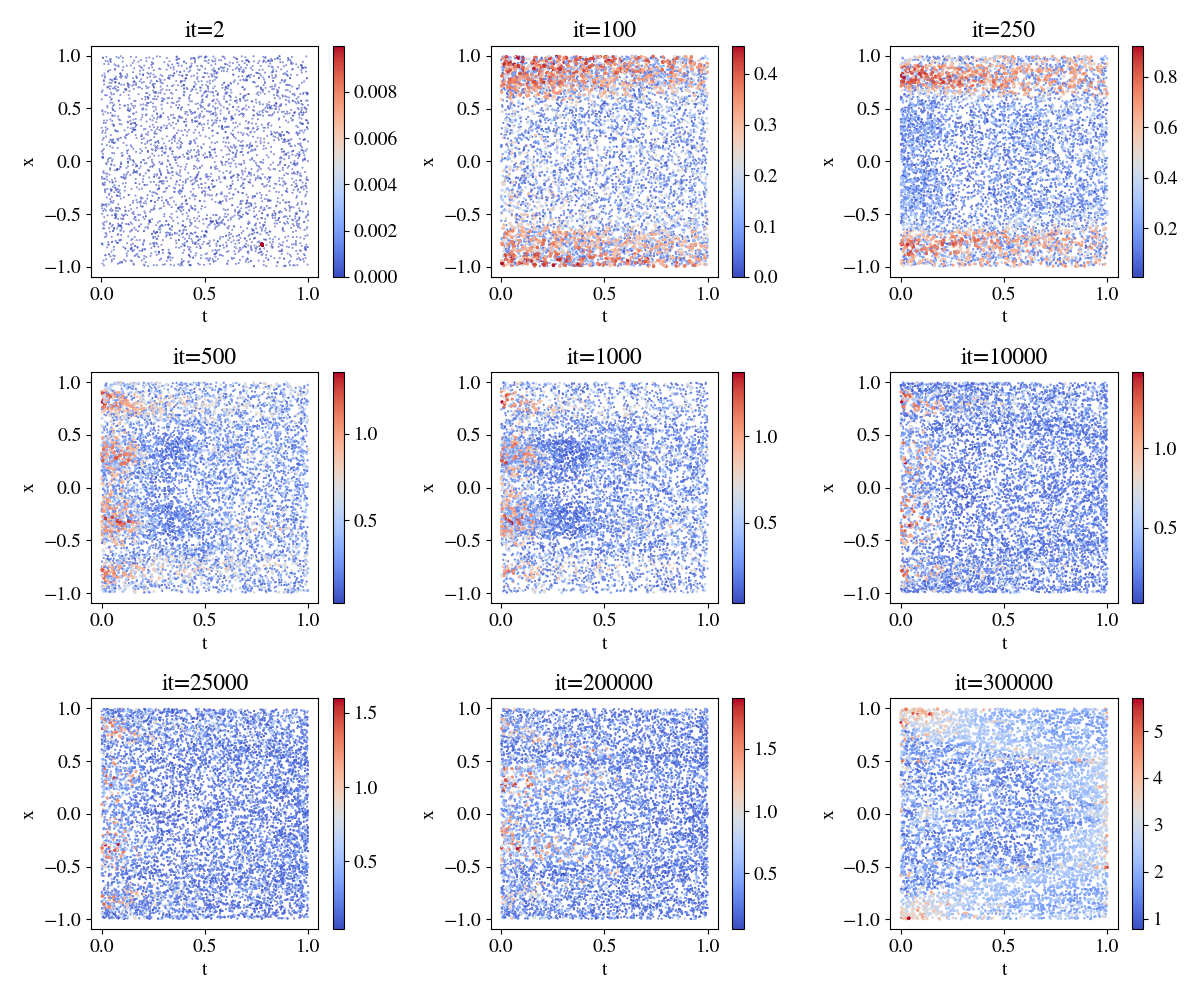}
    \caption{ssRBA-Rweight Evolution for the best performing KKAN model for Allen-Cahn Equation}
    \label{fig:RBAR-KART}
\end{figure}

\begin{figure}[H]
    \centering
    \includegraphics[width=1\linewidth]{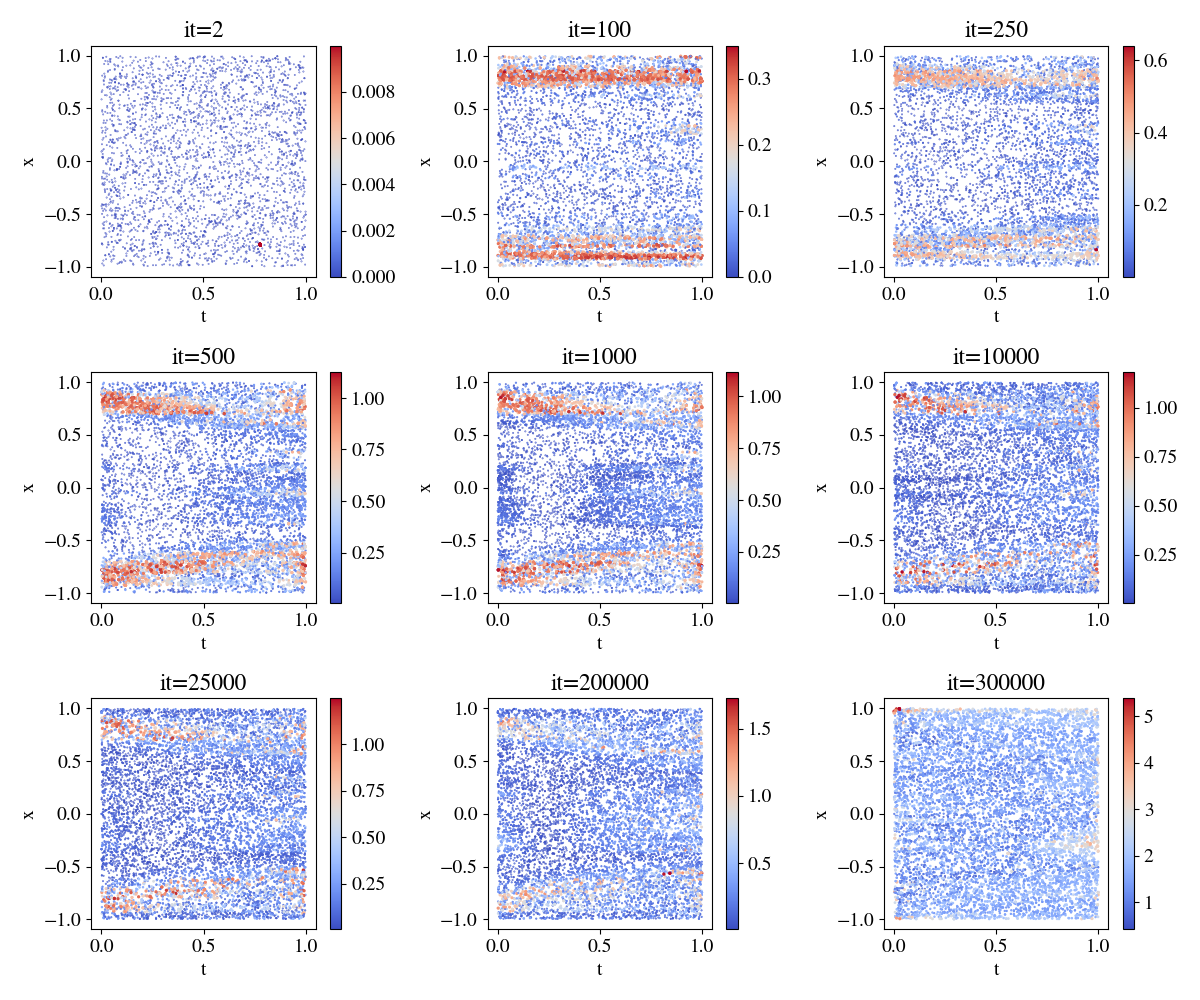}
    \caption{ssRBA-R weight Evolution for the best performing MLP for Allen-Cahn Equation}
    \label{fig:RBAR-MLP}
\end{figure}

\begin{figure}[H]
    \centering
    \includegraphics[width=1\linewidth]{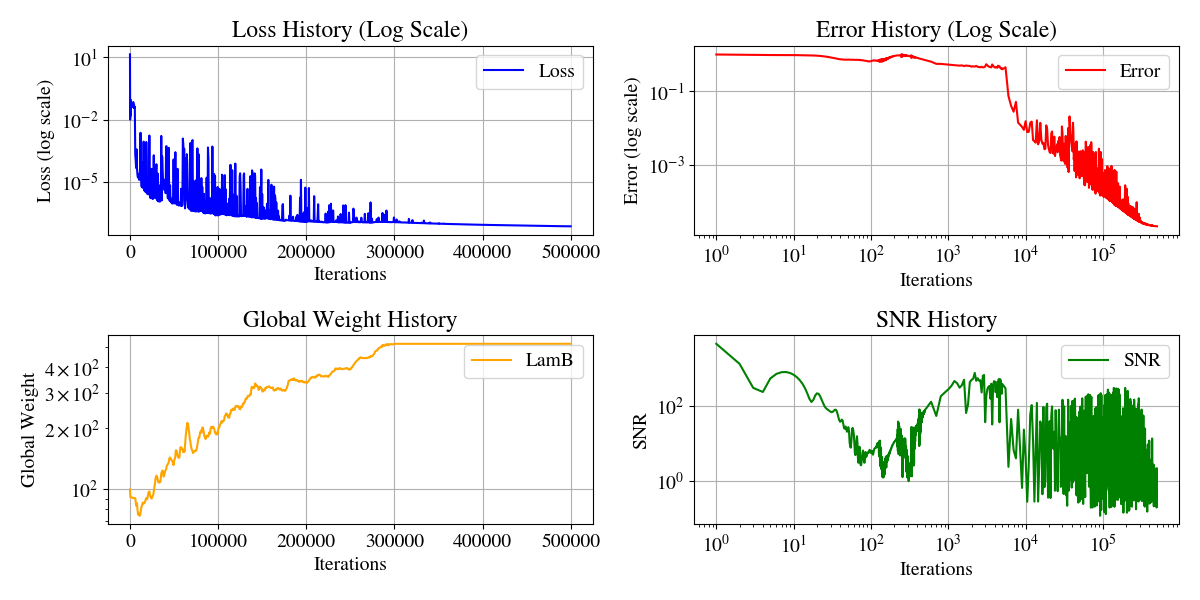}
    \caption{Loss, global weight, relative $L^2$ and SNR convergence history for the best performing KKAN+ssRBA model.}
    \label{fig:ssRBA-KKAN}
\end{figure}

\end{document}